%% file: main.tex
\documentclass[10pt,journal]{IEEEtran}
\usepackage{amsmath,amsfonts}
\usepackage{algorithm}
\usepackage{array}
\usepackage[caption=false,font=normalsize,labelfont=sf,textfont=sf]{subfig}
\usepackage{textcomp}
\usepackage{stfloats}
\usepackage{url}
\usepackage{verbatim}
\usepackage{graphicx}
\usepackage[numbers, compress, sort]{natbib}
\usepackage[noend]{algpseudocode}
\usepackage{algpseudocode}
\usepackage[export]{adjustbox}

\usepackage[colorlinks=true,linkcolor=blue,citecolor=blue,allcolors=blue]{hyperref}
\usepackage{multicol}
\usepackage{amsthm}
\usepackage[dvipsnames]{xcolor}
\usepackage{cancel}
\usepackage{url}
\usepackage{stackrel, amssymb}
\usepackage{mathtools}
\usepackage{amsthm}
\usepackage{bm}
\usepackage{bbm}
\usepackage{enumitem}
\usepackage{todonotes}
\usepackage{tikz}
\usepackage{xcolor}
\usepackage[arrow, matrix, curve]{xy}
\usepackage{extarrows}
\usepackage{tkz-graph}
\usepackage{cleveref}
\usepackage{mathrsfs}
\usepackage{array}
 \usepackage{multirow, makecell}
 \usepackage{tabularx,adjustbox,booktabs}

\usepackage{pdflscape}

\makeatletter
\newcommand*{\indep}{%
  \mathbin{%
    \mathpalette{\@indep}{}%
  }%
}
\newcommand*{\nindep}{%
  \mathbin{
    \mathpalette{\@indep}{\not}
  }%
}
\newcommand*{\@indep}[2]{%
  \sbox0{$#1\perp\m@th$}
  \sbox2{$#1=$}
  \sbox4{$#1\vcenter{}$}
  \rlap{\copy0}
  \dimen@=\dimexpr\ht2-\ht4-.2pt\relax
  \kern\dimen@
  {#2}%
  \kern\dimen@
  \copy0 
}
\makeatother

\def\*#1{\mathbf{#1}}
\usepackage{cleveref}

\usepackage{pifont}
\newcommand{\cmark}{\ding{51}}%
\newcommand{\xmark}{\ding{55}}%

\newcolumntype{x}[1]{>{\centering\let\newline\\\arraybackslash\hspace{0pt}}p{#1}}

\usepackage[dvipsnames]{xcolor}

\usepackage{tcolorbox} 

\tcbuselibrary{skins}

\newtcolorbox[auto counter, number within=section]{titlepanel}[1][]{%
    colback=gray!20, 
    colframe=black, 
    fonttitle=\bfseries,
    coltitle=black,
    center title,
    sharp corners,
    boxrule=0.2pt, 
    width=\linewidth,
    halign=center,
    left=0pt,
    right=0pt,
    top=0pt,
    bottom=0pt,
    enhanced,
    #1
}

\newtheorem{definition}{Definition}
\newtheorem{theorem}{Theorem}
\newtheorem{proposition}{Proposition}
\newtheorem{corollary}{Corollary}
\newtheorem{remark}{Remark}

\theoremstyle{definition}
\newtheorem{exmp}{Example}

\begin{document}

\title{dcFCI: Robust Causal Discovery Under Latent Confounding,
Unfaithfulness, and Mixed Data}

\author{Ad\`ele H. Ribeiro$^{1}$ and Dominik Heider\,$^{1}$
\thanks{
$^{1}$University of Münster, Institute of Medical Informatics, Münster, Germany
}
}
\markboth{Journal of \LaTeX\ Class Files,~Vol.~14, No.~8, August~2021}%
{Ribeiro \& Heider: dcFCI: Robust Causal Discovery Under Latent Confounding,
Unfaithfulness, and Mixed Data}

\maketitle

\begin{abstract}
Causal discovery is central to inferring causal relationships from observational data.
In the presence of latent confounding, algorithms
such as Fast Causal Inference (FCI) learn a Partial Ancestral Graph (PAG)
representing the true model's Markov Equivalence Class. However, their correctness
critically depends on \textit{empirical faithfulness}, the assumption that observed
(in)dependencies perfectly reflect those of the underlying causal model, which
often fails in practice due to limited sample sizes.
To address this, we introduce the first nonparametric score to assess a PAG's
compatibility with observed data, even with mixed variable types. This score is
both \textit{necessary and sufficient} to characterize structural uncertainty and distinguish between distinct PAGs.
We then propose data-compatible FCI (dcFCI), the first hybrid causal discovery
algorithm to jointly
address latent confounding, empirical unfaithfulness,
and mixed data types. dcFCI integrates our score into an (Anytime)FCI-guided
search that systematically explores, ranks, and validates candidate PAGs.
Experiments on synthetic and real-world scenarios demonstrate that dcFCI significantly
outperforms state-of-the-art methods, often recovering the true PAG even in small
and heterogeneous datasets. Examining top-ranked PAGs further
provides valuable insights into structural uncertainty, supporting more robust and informed causal reasoning and decision-making.
\end{abstract}


\section{Introduction}
\IEEEPARstart{P}{earl}'s causality framework \cite{pearl2000causality},
has profoundly transformed the field of causal inference. Based on the principles of
Structural Causal Models (SCMs), it provided rigoroups and intuitive tools for
accurately modeling and inferring causal relationships from observational studies.
This framework not only deepens our understanding of the true underlying causal
mechanisms -- thereby offering explanations for real-world phenomena --
but also enables the prediction of effects of \emph{unrealized} interventions
and the evaluation of counterfactual scenarios. These capabilities
are instrumental for decision-making and policy analysis in complex systems and
can inform the design of follow-up interventional studies.

Causal discovery is central to causal inference, particularly when background
knowledge alone cannot fully describe the underlying causal structure.
It aims to recover the directed acyclic graphical model (or causal diagram) $G$,
implied by the underlying SCM \( \mathcal{M} \).
Since some variables in \( \mathcal{M} \) may be unobserved, the goal is
to reconstruct \( G \) using only the observed variables \( \*V \) and
a dataset $\mathcal{D}$ sampled from the joint distribution \( P(\*V) \).
If \( \*V \) excludes certain confounder variables,
a situation known as \emph{causal insufficiency}, and no functional or
distributional assumptions are imposed on \( \mathcal{M} \), multiple graphical
models can equally represent \( P(\*V) \), rendering \( G \)
unidentifiable in general. As a result, algorithms aim to recover the Markov
Equivalence Class (MEC) of $G$, which consists of all models that entail exactly
the same set of conditional (in)dependencies and are asymptotically
indistinguishable based on goodness-of-fit scores.

Typically, a MEC is graphically represented by a Partial Ancestral Graph (PAG),
which encodes both the set of conditional independencies -- accessible via
\emph{m-separation}, an extension of d-separation for PAGs
\citep{jaber2022causal} -- and the ancestral relationships shared by all models
in the class. Specifically, tails and arrowheads in a PAG
indicate, respectively, ancestral (causal) and non-ancestral (non-causal)
relationships common to all models within the MEC.
A circle (\(\circ\)) edge mark indicates a non-invariant relationship,
meaning that within the MEC, there is at least one model where the edge mark
is a tail and another where it is an arrowhead.
Once a PAG is learned, it can be utilized in downstream causal analysis,
including effect identification and estimation
\citep{maathuis2015generalizedbd, perkovi2018complete, jaber2022causal},
enabling a complete and fully data-driven approach to causal inference.

The Fast Causal Inference (FCI) algorithm \citep{spirtes2001causation, zhang2008fci}
is a seminal approach to causal discovery, renowned for its theoretical
rigor, ability to handle latent confounding and selection bias,
and adaptability to various data types by accepting any type of conditional
independence test.
However, its soundness heavily relies on both \textit{distributional}
and \textit{empirical faithfulness}. Distributional faithfulness ensures that
\(P(\*V)\) precisely reflects all and only
the conditional (in)dependencies implied by d-separation in the true causal
diagram.
Some studies have explored detecting and relaxing this assumption,
though they are limited to causally sufficient settings
\citep{zhang2008detection, uhler2013geometry, zhalama2017weakening}.
However, empirical faithfulness, which states that \(\mathcal{D}\)
perfectly reflect these (in)dependencies, has received far less attention.
In practice, this translates to
assuming that inferences from conditional independence tests are \textit{perfect},
allowing FCI to operate without
accounting for any uncertainty in these decisions.
This is critical, as errors in these inferences -- often due to limited sample size --
can propagate through the learning process, potentially leading to
multiple incorrect causal orientations.

To address FCI's limitations, several extensions and alternative approaches
have been developed. Conservative FCI (cFCI)
\citep{colombo2012learning, colombo2014order} improves robustness by leaving
edges unoriented when their orientation relies on potentially unfaithful
conditional independencies. Other methods integrate scoring techniques, such as
Bayesian Constraint-Based Causal Discovery (BCCD) \citep{claassen2012bccd}, which
combines a logical adaptation of cFCI with Bayesian reliability scores.
Recent search-based algorithms focus on identifying the optimal Maximal
Ancestral Graph (MAG) \citep{richardson2002ancestral} -- a subclass of
Markov-equivalent models that share all ancestral relationships -- using
the Bayesian Information Criterion (BIC) for linear Gaussian models.
Notable methods include Differentiable Causal Discovery (DCD)
\citep{bhattacharya2021differentiable}, which employs gradient-based optimization,
and MAG Structure Learning (MAGSL) \cite{rantanen2021maximal},
which guarantees the globally optimal MAG through exact search.
Moreover, Greedy PAG Search (GPS) \citep{claassen2022greedy}
performs a greedy search directly in the space of PAGs, also
leveraging the Gaussian BIC. \Cref{sec:related_work} provides
a detailed discussion of the related literature.

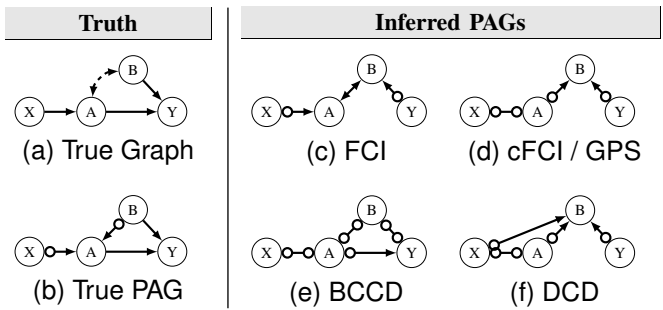
\begin{figure}[ht]
    \centering
    \begin{minipage}{\linewidth}

    \begin{minipage}{0.31\linewidth}
        \begin{titlepanel}[width=\linewidth]
                {\small \bf Truth}
        \end{titlepanel}

        \centering
        \subfloat[True Graph]{
        \resizebox{0.86\linewidth}{!}{%
          \begin{tikzpicture}[node distance=0.75cm, auto]
          \tikzset{vertex/.style = {shape=circle, draw, minimum size=1.5em}}
            \node (X) [circle, draw] {X};
            \node (A) [circle, draw, right=of X] {A};
            \node (Y) [circle, draw, right=1.2cm of A] {Y};
            \node (B) [circle, draw, above right=0.7cm of A] {B};
            \draw[-latex, line width=0.5mm] (A) -- (Y);
            \draw[-latex, line width=0.5mm] (B) -- (Y);
            \draw[latex-latex, dashed, line width=0.5mm, bend left=40] (A) to (B);
            \draw[-latex, line width=0.5mm] (X) -- (A);
          \end{tikzpicture}
          }
          \label{fig:ex1_ADMG}
        }

        \subfloat[True PAG]{
        \resizebox{0.86\linewidth}{!}{%
             \begin{tikzpicture}[node distance=0.75cm, auto]
              \node (X) [circle, draw] {X};
              \node (A) [circle, draw, right=of X] {A};
              \node (Y) [circle, draw, right=1.2cm of A] {Y};
              \node (B) [circle, draw, above right=0.7cm of A] {B};
              \draw[-latex, line width=0.5mm] (A) -- (Y);
              \draw[-latex, line width=0.5mm] (B) -- (Y);
              \draw[o-latex, line width=0.5mm] (B) to (A);
              \draw[o-latex, line width=0.5mm] (X) -- (A);
            \end{tikzpicture}
            }
            \label{fig:ex1_true_PAG}
        }
    \end{minipage}%
    \begin{minipage}{0.03\linewidth}
      \centering
      \hspace{0.05cm}
    \rule{0.5pt}{4cm}
    \end{minipage}
    \begin{minipage}{0.63\linewidth} 
       \begin{titlepanel}[width=\linewidth]
      {\small \bf Inferred PAGs}
      \end{titlepanel}

        \vspace{-1em}
        \centering
        \subfloat[FCI]{
         \resizebox{0.43\linewidth}{!}{%
           \begin{tikzpicture}[node distance=0.75cm, auto]
          \node (X) [circle, draw] {X};
          \node (A) [circle, draw, right=of X] {A};
          \node (Y) [circle, draw, right=1.2cm of A] {Y};
          \node (B) [circle, draw, above right=0.7cm of A] {B};
          \draw[o-latex, line width=0.5mm] (X) -- (A);
          \draw[o-latex, line width=0.5mm] (Y) -- (B);
          \draw[latex-latex, line width=0.5mm] (A) to (B);
        \end{tikzpicture}
        }
        \label{fig:ex1_FCI_PAG}
        }
        \subfloat[cFCI / GPS]{
        \resizebox{0.43\linewidth}{!}{%
         \begin{tikzpicture}[node distance=0.75cm, auto]
          \node (X) [circle, draw] {X};
          \node (A) [circle, draw, right=of X] {A};
          \node (Y) [circle, draw, right=1.2cm of A] {Y};
          \node (B) [circle, draw, above right=0.7cm of A] {B};
          \draw[o-o, line width=0.5mm] (X) -- (A);
          \draw[o-latex, line width=0.5mm] (Y) -- (B);
          \draw[o-latex, line width=0.5mm] (A) to (B);
        \end{tikzpicture}
        }
        \label{fig:ex1_cFCI_PAG}
        }

        \subfloat[BCCD]{
        \resizebox{0.43\linewidth}{!}{%
         \begin{tikzpicture}[node distance=0.75cm, auto]
          \node (X) [circle, draw] {X};
          \node (A) [circle, draw, right=of X] {A};
          \node (Y) [circle, draw, right=1.2cm of A] {Y};
          \node (B) [circle, draw, above right=0.7cm of A] {B};
          \draw[o-latex, line width=0.5mm] (A) -- (Y);
          \draw[o-o, line width=0.5mm] (B) -- (Y);
          \draw[o-o, line width=0.5mm] (B) to (A);
          \draw[o-o, line width=0.5mm] (X) -- (A);
        \end{tikzpicture}
        }
        \label{fig:ex1_BCCD_PAG}
        }
        \subfloat[DCD]{
        \resizebox{0.43\linewidth}{!}{%
         \begin{tikzpicture}[node distance=0.75cm, auto]
          \node (X) [circle, draw] {X};
          \node (A) [circle, draw, right=of X] {A};
          \node (Y) [circle, draw, right=1.2cm of A] {Y};
          \node (B) [circle, draw, above right=0.7cm of A] {B};
          \draw[o-latex, line width=0.5mm] (X) to (B);
          \draw[o-latex, line width=0.5mm] (Y) -- (B);
          \draw[o-latex, line width=0.5mm] (A) to (B);
          \draw[o-o, line width=0.5mm] (X) -- (A);
        \end{tikzpicture}
        }
        \label{fig:ex1_DCD_PAG}
        }
     \end{minipage}
    \end{minipage}

    \caption{True causal diagram (a) and PAG (b) compared with PAGs inferred
    by SOTA algorithms: FCI (c), cFCI and GPS (d), BCCD (e), and DCD (f).}
    \label{fig:ex_1}
\end{figure}

Despite these advances, significant challenges remain due to empirical unfaithfulness.
Conditional independence tests often yield inaccurate results in low-data scenarios
and can still be unreliable with relatively large sample sizes and a
handful of variables.
Additionally, their effectiveness depends on the significance level,
which is often arbitrarily set and may not reflect the varying levels of
uncertainty in the data. As a result, most existing algorithms tend to produce
a PAG that is \emph{fundamentally flawed}, failing not only to accurately
represent the true model's MEC but also to ensure its correct characterization
and encode best data-supported conditional (in)dependencies
(see \Cref{fig:ex_1} and \Cref{sec:description_fig1} for details).

The fact that state-of-the-art (SOTA) algorithms
produce fundamentally different outputs from the same data
raises serious concerns about their robustness in practice.
Constraint-based approaches have a strong theoretical foundation,
but are highly vulnerable in unfaithful scenarios due to their sensitivity to
the order of independence tests and error propagation.
In contrast, scoring procedures offer greater robustness by assessing
overall compatibility with the data. However, their effectiveness depends on
conducting a well-structured search and scores are often constrained by specific
parametric and distributional assumptions.
This underscores the urgent need for a hybrid approach that
combines the strengths of both worlds. Such an approach should integrate a
flexible and reliable global measure of data-PAG compatibility, adaptable to
various data types, ensuring optimal alignment between
observed (in)dependences and m-separations, while effectively
addressing inconsistencies in MEC characterization.

In light of this, this work makes the following contributions:

\begin{enumerate}
    \item A sufficient and necessary nonparametric score is introduced
    to assess data-PAG alignment, the first to support mixed data types,
    by accounting for posterior probabilities of essential (in)dependencies,
    directly derived from p-values of likelihood-ratio conditional independence tests.
    \item A novel hybrid causal discovery algorithm, called
    data-compatible FCI (dcFCI), is presented, enhancing the robustness and reliability
    of inferred PAGs by accurately encoding best supported
    (in)dependencies,
    while adhering to the MEC characterization and quantifying uncertainty.
    \item Simulation studies demonstrate that dcFCI outperforms SOTA algorithms in
    low-data, empirically unfaithful regimes, with its top-ranked inferred PAGs
    often including the true PAG, even in datasets with mixed data types.
    \item In a real-world application, dcFCI demonstrated greater robustness
    than SOTA algorithms, consistently identifying causal relations across
    overlapping variable subsets.
    \item The \texttt{dcFCI} R package, which implements the algorithm with
    support for PAG scoring from conditional independence tests across
    continuous, binary, and multinomial variables, is available at
    \url{https://github.com/adele/dcFCI}.
\end{enumerate}

\section{Illustrative Example -- \Cref{fig:ex_1}}
\label{sec:description_fig1}
To illustrate the unreliability of SOTA algorithms under unfaithfulness,
consider the causal diagram in \Cref{fig:ex1_ADMG}.
The independencies implied by d-separation are exactly
\( \{(B \indep X),\,(X \indep Y \mid A, B)\} \).
From a finite dataset \(\mathcal{D}\) of 10,000 samples generated
by a Gaussian linear SCM adhering to the true causal diagram,
and using appropriate conditional independence tests, one observes,
at a significance level of 0.01 (a typical choice in practice),
the results in Table \ref{tab:ex1_citests}, including
not only all the
expected true independencies but also unfaithful ones:
\( \{(B \indep X \mid A),\,(B \indep X \mid Y),\,(B \indep X \mid A, Y),
\,(A \indep Y \mid X),\,(X \indep Y \mid A)\} \).

Based on these results, FCI produces the PAG in
\Cref{fig:ex1_FCI_PAG}.\footnote{This is achieved by removing the edges between
the variable pairs identified as conditionally independent and then applying
Rule 0, which results in the unshielded colliders
$X \ast\!\!\rightarrow A \leftarrow\!\!\ast B$ and
$A \ast\!\!\rightarrow B \leftarrow\!\!\ast Y$.} It
not only significantly differs from the true PAG in
\Cref{fig:ex1_true_PAG}, suggesting incorrect non-ancestral relationships
(\(B \leftarrow\!\!\circ Y\)), but also encodes (in)dependence constraints that
are not well-supported by the data. For example, it erroneously implies
(through m-separation) that $(A \indep Y)$, $(X \indep Y)$, and  $(X \indep Y | B)$.
Furthermore, it encodes false, not well-supported dependencies, such as
$(B \nindep X \mid A)$ and  $(X \nindep Y \mid A, B)$.

In contrast, cFCI detects ambiguity in the triple \(\langle B, A, X \rangle\),
leaving it unoriented in the resulting PAG (\Cref{fig:ex1_cFCI_PAG}).
However, this disrupts the MEC characterization, making it incompatible with
probabilistic or causal inference tools designed for PAGs.
For instance, applying m-separation considering this ambiguous triple falsely implies
that \((B \nindep X)\). A similar issue arises with BCCD: while it correctly
recovers the true PAG's skeleton, the resulting graph (\Cref{fig:ex1_BCCD_PAG})
fails to represent a unique MEC, rendering it unsuitable for downstream analyses.

Turning to score-based algorithms, leveraging the Gaussian BIC
(well-suited for this dataset), GPS learns the same PAG as cFCI.
DCD and MAGSL, by first searching for a MAG before deriving the corresponding PAG,
produce valid PAGs. However, while MAGSL accurately recovers the true PAG,
DCD infers the structure in \Cref{fig:ex1_DCD_PAG}, which poorly represents the actual causal
relationships and observed (in)dependencies.

\section{Background}

\subsection{Notation and Terminology}

We use boldface capital letters for variable sets and boldface lowercase
letters for their value assignments.

We denote \( P(\*V) \) as the joint distribution over \( \*V \).
For any disjoint subsets \(\*X\), \(\*Y\), and \(\*Z\) of
\( \*V \), the conditional independence relation between \(\*X\) and \(\*Y\)
given \(\*Z\), where \(\*Z\) may be empty, is denoted by
\((\*X, \*Y \mid \*Z)_P\). This relation must either indicate conditional
independence, written as \((\*X \indep \*Y \mid \*Z)_P\), or conditional
dependence, represented by \((\*X \nindep \*Y \mid \*Z)_P\).

To characterize conditional independence in graphical models, we consider the
m-separation criterion for MAGs and PAGs as defined in \cite{zhang2008causal} and
\cite{jaber2022causal}, respectively. Let \( M \) be a MAG and \( \mathcal{P} \)
be a PAG, both on the vertex set \( \*V \). For any disjoint subsets
\( \*X \), \( \*Y \), and \( \*Z \subset \*V \), m-separation of \( \*X \) and
\( \*Y \) by \( \*Z \) in \( M \) (resp. \( \mathcal{P} \)) is denoted by
\((\*X \indep_m \*Y \mid \*Z)_{M}\) (resp.
\((\*X \indep_m \*Y \mid \*Z)_{\mathcal{P}}\)). Conversely, m-connectedness is
represented by \((\*X \nindep_m \*Y \mid \*Z)_{M}\) (resp.
\((\*X \nindep_m \*Y \mid \*Z)_{\mathcal{P}}\)). When the context is unambiguous,
we may omit the subscript indicating the associated probability distribution or
model for clarity.

Further foundational concepts are provided in \Cref{ape:further_definitions}.

\subsection{Characterization of the Markov Equivalence Class
in the Presence of Latent Confounding} 
\label{sec:mec}

As noted in the introduction, MAGs are
commonly used in causal discovery when latent confounders are present.
Their advantage lies in encoding all (non-)ancestral and
conditional (in)dependence relations implied by a causal diagram,
without explicitly modeling latent variables \citep{richardson2002ancestral}.
If \( M \) is the MAG of a causal diagram \( G \), then the class of MAGs that
encode (via m-separation) the same set of conditional (in)dependence relations
as \( M \) constitutes the MEC of \( M \) and, consequently, of \( G \), and is
represented by a PAG \( \mathcal{P} \) \citep{zhang2007characterization}.

The most straightforward way to characterize a MEC is through the set of
conditional (in)dependence relations it entails. Two MAGs, \( M_1 \) and \( M_2 \),
over a set of variables \( \*V \), are Markov-equivalent if and only if,
for any three disjoint subsets \( \*X, \*Y, \*Z \subset \*V \), it holds that
\( (\*X \indep_m \*Y \mid \*Z)_{M_1} \) if and only if
\( (\*X \indep_m \*Y \mid \*Z)_{M_2} \). In other words, MECs that differ in any
of these implied (in)dependencies are distinct.

A clear limitation of this characterization is that the number of conditional
independence relations involved can quickly become impractical.
To reduce the complexity, one can leverage 
that m-separations and m-connections implied by a MAG define an independence
model that forms a graphoid \cite{lauritzen2014markov, lauritzen2018markov},
thereby enabling
the use of the composition property. Specifically, for any disjoint sets
of variables \( \*X_1 \), \( \*X_2 \), \( \*Y \), and \( \*Z \), the conditions
\( (\*X_1 \indep_m \*Y \mid \*Z) \) and \( (\*X_2 \indep_m \*Y \mid \*Z) \)
imply \( (\*X_1 \cup \*X_2 \indep_m \*Y \mid \*Z) \). In words,
pairwise conditional (in)dependence relations are sufficient to determine
higher-order (in)dependencies. However, even with this simplification,
the number of relations still grows rapidly as the number of variables increases,
making it impractical for large systems.

Specifically, for a set $\*V$ with $|\*V| = p$ variables, the number of pairwise
conditional independence relations, when considering all possible conditioning
sets of up to $r_{\text{max}} = p-2$ variables, grows combinatorially
with $p$ and is given by:
\begin{equation}
\label{eq:pairwise_ci}
T = \binom{p}{2} \times \sum_{r=0}^{r_{\text{max}}} \binom{p-2}{r}.
\end{equation}
For $p =$ 4, 5, 10, and 20, 
this results in $T=$ 24, 80, 11,520,
and 49,807,360 conditional independence relations, respectively.

Despite these challenges, MECs of MAGs can be uniquely characterized
by their skeleton (adjacencies) and collider triples, either unshielded of identified
via discriminating paths, also known as colliders with order
\cite{zhang2007characterization, ali2009mec}.
Recently, \cite{claassen2022greedy} proposed a simpler recursive definition
of triples with order that eliminates the need to explicitly identify
discriminating paths. They also introduced the MAG-to-MEC algorithm,
which efficiently computes all such triples in polynomial time.
These results show that the skeleton and colliders with order
encapsulate all conditional independencies needed to define the
complete PAG, with all other orientations derived purely from graphical rules.
Any change to these components necessarily alter the MEC.
For further details, see \Cref{ape:markov_equivalence}.

\subsection{Likelihood-Ratio Tests for Conditional Independence Among Variables of Diverse Types}
\label{sec:lr-tests}

Constraint-based causal discovery algorithms such as the FCI and its variants
aim to learn the true model's MEC
by leveraging observed conditional (in)dependencies,
typically determined via conditional independence tests.

The most commonly used conditional independence tests are likelihood-ratio tests
or their approximations \cite{tsagris2018constraint}.
Examples include the partial correlation test for continuous
multivariate Gaussian variables, as well as the G-test, and the asymptotically
equivalent \(\chi^2\)-test for categorical variables.
As shown by \cite{tsagris2018constraint},
likelihood-ratio tests can be straightforwardly constructed to evaluate
conditional independence among variables of diverse types.
See further details in \Cref{ape:likelihood_ratio_tests}.
These tests can be efficiently implemented using regression models,
such as linear, logistic, multinomial, or ordinal regression, for continuous,
binary, nominal, and ordinal variables, respectively. Non-linear, more
complex models may also be employed, depending on the data and specific requirements.

\subsection{Bayes Factor for Standard Test Statistics}
\label{sec:bayes_factors}

The performance of constraint-based causal discovery algorithms heavily
relies on accurately inferring both (conditional) dependencies and independencies
among observed variables. These inferences are typically made using statistical
tests for the null hypothesis of (conditional) independence, against the alternative
hypothesis of (conditional) dependence. This approach has two major drawbacks:
1) it depends on arbitrarily chosen p-value thresholds (e.g., 0.01 or 0.05) to define
statistical significance, and 2) p-values do not quantify the evidence for either
the null or alternative hypotheses.
Note that it is incorrect to interpret
the absence of evidence to reject the null hypothesis as support for or acceptance
of the null hypothesis. Failing to reject the null hypothesis simply means
that the data do not provide sufficient evidence to contradict it.

In contrast, in a Bayesian framework, Bayes factors quantify the relative
evidence provided by the data \(\mathcal{D}\) in favor of one hypothesis over
another. Let \(H_0\) be the null hypothesis (e.g., conditional independence)
and \(H_1\) the alternative (e.g., conditional dependence).
The Bayes factor in favor of \(H_1\) is:
\[
BF_{10} = \frac{P(\mathcal{D} \mid H_1)}{P(\mathcal{D} \mid H_0)},
\]
in which \(P(\mathcal{D} \mid H_1)\) and \(P(\mathcal{D} \mid H_0)\) denote
the marginal likelihoods of the data (\(\mathcal{D}\)) under \(H_1\) and \(H_0\),
respectively. A \(BF_{10} > 1\) 
favors \(H_1\), while \(BF_{10} < 1\) favors 
\(H_0\).

By applying Bayes' theorem, we can further compute the posterior probability of
each hypothesis. This incorporates both prior beliefs, \(P(H_0)\) and \(P(H_1)\),
which can be non-informative (equal to 0.5), and the Bayes factor, thus
considering both data and any prior knowledge.

Despite their potential, the practical use of Bayes factors has been limited by
the need to explicitly define alternative hypotheses. However, recently,
\cite{johnson2023bayes} provided closed-form expressions for Bayes factors derived
from standard test statistics, including \(\chi^2\) from likelihood-ratio tests,
which are widely used for testing conditional independencies in causal discovery,
as discussed in \Cref{sec:lr-tests}.
The authors have also made an R package called BFF
publicly available, offering tools to calculate the Bayess factor for various
standard test statistics \citep{bff2023rpackage}.
See further details in \Cref{ape:bff}.

\section{A Nonparametric PAG Score}

The set of conditional (in)dependence relations in the true distribution
\( P(\*V) \) defines an independence model, \(\mathscr{I}(P)\).
Since \( P(\*V) \) is typically unknown in practice, it must be estimated
from an observational dataset \(\mathcal{D}\) for \(\*V\), which defines
\(\mathscr{I}(\mathcal{D})\), the independence model of \(\mathcal{D}\).
Importantly, the equality \(\mathscr{I}(\mathcal{D}) = \mathscr{I}(P)\) holds
only under empirical faithfulness, which is rarely the case in limited datasets
due to sampling variability or noise \citep{uhler2013geometry}.
Similarly, the set of all m-separations and m-connections in a PAG \(\mathcal{P}\)
defines an independence model, \(\mathscr{I}_{m}(\mathcal{P})\).
Causal discovery aims to identify a PAG \(\mathcal{P}\) such that
\(\mathscr{I}_m(\mathcal{P})\) best matches \(\mathscr{I}(\mathcal{D})\).
In empirically faithful settings,
\(\mathscr{I}(\mathcal{D})\) accurately reflects \(\mathscr{I}(P)\),
ensuring \(\mathscr{I}_m(\mathcal{P})\) aligns with
\(\mathscr{I}(P)\).
Consequently, under distributional faithfulness, inferences drawn
from \(\mathcal{P}\), including ancestral (causal) relationships,
are guaranteed to be valid in the true model and all Markov equivalent models.

We formalize this compatibility between the independence models of a
PAG and a dataset in the following definition:

\begin{definition}[Data-PAG Compatibility]
It refers to the degree of alignment between the independence
model induced by a given PAG \( \mathcal{P} \), denoted \( \mathscr{I}_m(\mathcal{P}) \),
and the independence model derived from the observed dataset \( \mathcal{D} \),
denoted \( \mathscr{I}(\mathcal{D}) \).
\end{definition}

Here, we aim to develop a compatibility score for a PAG $\mathcal{P}$ with respect
to a dataset $\mathcal{D}$, denoted $S_{\mathcal{P},\mathcal{D}}$. This score
is designed to quantify the alignment between the (in)dependence relations in
$\mathscr{I}_m(\mathcal{P})$ and those in $\mathscr{I}(\mathcal{D})$, thereby
providing a measure of how well the observed data $\mathcal{D}$ supports
$\mathcal{P}$.

Due to space constraints, all proofs are in \Cref{ape:proofs}.

\subsection{A Straightforward Score for Data-PAG Compatibility}
\label{sec:straightforward_score}

Building on the most straightforward characterization of MECs and leveraging the
compositional property of independence models for mixed graphs, we propose a
straightforward compatibility score $S_{\mathcal{P},\mathcal{D}}$ that reflects
how well the observed data $\mathcal{D}$ supports \textit{all}
pairwise conditional (in)dependencies implied by m-separations
and m-connections in a PAG $\mathcal{P}$.

Let $ H_{(X \indep Y | \*Z)_{P}}$ be the hypothesis that
$(X \indep Y | \*Z)_{P}$ and
$H_{(X \nindep Y | \*Z)_{P}}$ be the hypothesis that
$(X \nindep Y | \*Z)_{P}$.
The sets of (conditional) (in)dependence
hypothesesimplied by
$\mathcal{P}$ are:
\begin{equation}
\label{eq:straightforward_hypotheses}
\begin{aligned}
\*H_{\indep(\mathcal{P},P)} = &\{
H_{(X \indep Y \mid \*Z)_{P}}
\mid  X, Y \in \*V, \, \*Z \subseteq \*V \!\setminus\! \{X, Y\}, \\
&\text{ and } (X \indep_m Y | \*Z)_{\mathcal{P}} \}\\
\*H_{\nindep(\mathcal{P},P)} = & \{
H_{(X \nindep Y \mid \*Z)_{P}}
\mid  X, Y \in \*V, \, \*Z \subseteq \*V \!\setminus\! \{X, Y\}, \\
&\text{ and } (X \nindep_m Y | \*Z)_{\mathcal{P}} \}.
\end{aligned}
\end{equation}

As outlined in Section \ref{sec:bayes_factors},
the posterior probability for a single conditional independence hypothesis or
its alternative, the conditional dependence hypothesis, can be derived
from any \(\chi^2\) statistic or corresponding p-value of likelihood-ratio tests.
However, calculating the joint probability of multiple hypotheses is challenging,
as they are typically not mutually independent. To address this, we propose
estimating the bounds on the probability of the conjunction of these hypotheses
using the Fréchet inequality \citep{frechet1935generalisation}.

\begin{definition}
\label{def:straightf_compat_score}
Let $\*H_{\indep(\mathcal{P},P)}$ and $\*H_{\nindep(\mathcal{P},P)}$ be as in
\Cref{eq:straightforward_hypotheses} and define
$\*H_{\mathcal{P},P} = \*H_{\indep(\mathcal{P},P)} \cup \*H_{\nindep(\mathcal{P},P)}$.
Further, let $H_i \in \*H_{\mathcal{P},P}$, with  $i = 1, \dots, T$.
The straightforward data-PAG compatibility score is the posterior probability:
$$
S_{\mathcal{P},\mathcal{D}} =  P \left( \*H_{\mathcal{P},P} | \mathcal{D} \right)
= P\left(\bigcap_{i=1}^{T} H_i \mid \mathcal{D}\right),
$$
with Fréchet bounds :
$$
\begin{aligned}
&\max\!\left(\!0, \sum_{i=1}^{T} P(H_i \mid \mathcal{D}) \!-\! (T-1)\!\right)
\!\leq\! S_{\mathcal{P},\mathcal{D}} \!\leq\! \min_{i} P(H_i \mid \mathcal{D}).
\end{aligned}
$$
\end{definition}

\begin{remark}
Since \( S_{\mathcal{P},\mathcal{D}} \) in \Cref{def:straightf_compat_score}
captures all (in)dependence relations implied by \(\mathcal{P}\), it is evident
that it fully characterizes its MEC and data compatibility.
\end{remark}

Refer to \Cref{ex:straigthf_score_baselines} for scoring of the PAGs
in Figure \ref{fig:ex_1} using the straightforward data-PAG compatibility score.

\vspace{0.5em}
\textbf{Limitations.}
The straightforward score offers a key advantage: it functions as a standalone
measure, relying solely on the data and the PAG, making it well-suited for
evaluating and comparing the compatibility of any PAG with the observed dataset.
However, it has several notable limitations.
First, it relies on the joint distribution of
all pairwise conditional independence relations. As the number of relations
increases combinatorially with the number of variables, the score quickly
approaches a Fréchet lower bound of zero,
complicating statistically meaningful comparisons among PAGs,
as shown in \Cref{ex:straigthf_score_baselines}.
Second, the estimation of the score's bounds treats
all hypotheses equally, ignoring their relevance to the model's
structure and interdependencies.

To address these limitations, we propose a targeted scoring approach
in the following section, specifically designed for selecting the
highest-scoring PAGs from a list of candidates.

\subsection{An Incremental, MEC-Targeted PAG Score}

As discussed in the previous section, the straightforward data-PAG
compatibility score can be overly conservative in practice. This is because
it not only typically considers a large number of conditional independence
hypotheses, but also overlooks their structural relevance and interconnectness.

We propose refining the scoring approach
by focusing exclusively on the conditional (in)dependence hypotheses
essential for identifying the
skeleton and triples with order, which, as described in \Cref{sec:mec},
fully characterize the MEC.

We also propose leveraging the iterative PAG construction from
\textit{Anytime FCI} \citep{spirtes01anytime}.
For \( p = |\*V| \) variables, the algorithm iterates over conditioning set
sizes \( r = 0, \dots, p-2 \).  At each iteration \( r \), it constructs an
intermediate PAG, called the \( r \)-PAG,
denoted as \( \mathcal{P}^{(r)} = (\*V, \*E^{(r)}) \),
by incorporating independencies conditioned
on sets of size \( \leq r \). The equivalence class of
\( \mathcal{P}^{(r)} \) is referred to as the \( r \)-MEC.
Notably, every \( m \)-separation, \( m \)-connection relative to a set
of size \( \leq r \),  and definite (non-)ancestral relation in
\( \mathcal{P}^{(r)} \) remains valid in \( \mathcal{P} \)
and, consequently, in all models within its MEC.
Over iterations, this $r$-PAG is refined with larger conditioning sets,
integrating new information while ensuring consistency with earlier stages.
This incremental approach allows stopping the algorithm's
outer loop at any iteration, yielding a valid but potentially less informative $r$-PAG.
Notably, $r$-PAGs constructed with smaller \(r\) values
are generally more reliable, as larger conditioning sets
often reduce statistical power,
increasing the risk of false conclusions and a cascade of orientation errors.

Under faithfulness, identifying a minimal separator of at most \( r \)
variables for each (conditionally) independent pair is sufficient to
accurately recover the skeleton and colliders with order,
fully determining the \( r \)-PAG \( \mathcal{P}^{(r)} \).
This follows directly from the soundness and completeness of the
\textit{Anytime FCI}.

Under unfaithfulness, observed (in)dependencies may be false or non-minimal,
leading to conflicts and ambiguities
in collider detection. To address this,
we propose quantifying uncertainty in the $r$-MEC's key
components by assessing the posterior probabilities of relevant
(in)dependence hypotheses.

Let $\mathfrak{S}(\mathcal{P}^{(r)})$ denote the skeleton of $\mathcal{P}^{(r)}$.
The set of (in)dependence hypotheses on $P(\*V)$
relevant to $\mathfrak{S}(\mathcal{P}^{(r)})$,
denoted $\*H_{\mathfrak{S}(\mathcal{P}^{(r)}), P}$, includes
those ensuring that adjacent nodes remain dependent given any set of size of at most
$r$, while non-adjacent nodes are independent given their minimal separators but
dependent when conditioning on any subset of the minimal separator that omits
at least one node \citep{tian1998finding}.

\begin{definition}
\label{def:skel_hypotheses}
Let \( \operatorname{MinSep}_{\mathcal{P}^{(r)}}(X, Y) \)
be the set of minimal separators for a pair of non-adjacent nodes
\( \{X, Y\} \) implied by \( \mathcal{P}^{(r)} \).
The set of in(dependence) hypotheses relevant
to $\mathfrak{S}(\mathcal{P}^{(r)})$, denoted by
\( \*H_{\mathfrak{S}(\mathcal{P}^{(r)}), P} \), is given by the union of
\begin{enumerate}[label=(\alph*)]
\item \label{itm:skel_a}
$\{ H_{( X \nindep Y | \*Z )_P} \mid \{X, Y\} \in \*E^{(r)}$ and
$\*Z \subseteq \*V \setminus \{X, Y\}$, with $|\*Z | \leq r \}$
\item \label{itm:skel_b}
$\{ H_{(X \indep Y | \*Z)_P}| \{X, Y\}~\not\in~\*E^{(r)},
\*Z~\in~\operatorname{MinSep}_{\mathcal{P}^{(r)}}(X,Y)\}$
\item \label{itm:skel_c}
$\{ H_{( X \nindep Y | \*Z' )_P} \mid \{X, Y\} \not\in \*E^{(r)}$ and $|\*Z'| = |\*Z| - 1$,
where $\*Z \neq \emptyset$ and $\*Z \in \operatorname{MinSep}_{\mathcal{P}^{(r)}}(X,Y)\}$.
\end{enumerate}
\end{definition}

Before definining the the in(dependence) hypotheses relevant to the
colliders with order $\mathfrak{C}({\mathcal{P}^{(r)}})$ in $\mathcal{P}^{(r)}$,
consider:

\begin{definition}
\label{def:triple_corresponds}
    A triple with order \(\langle A, B, C \rangle\) corresponds to a
    (not necessarily unique) pair of non-adjacent
    nodes \(\{X, Y\}\) in a PAG \(\mathcal{P}^{(r)}\) if one of the following holds:
    \begin{itemize}
        \item \( X = A \), \( Y = C \), and \(\langle A, B, C \rangle\) is
        an unshielded triple; 
        \item
        \(\langle X, \ldots, A, B, Y \rangle\) (with \( Y = C \))
         or  %
         \(\langle Y, \ldots, A, B, X \rangle\) (with \( X = C \))
        forms a discriminating path for $B$.
    \end{itemize}
\end{definition}

The next two results establish the relevance of
$\operatorname{MinSep}_{\mathcal{P}^{(r)}}$ for
both non-collider and collider triples with order:
\begin{proposition}
\label{prop:minsep_noncollider}
Let $\*Z \in \operatorname{MinSep}_{\mathcal{P}^{(r)}}(X,Y)$.
For every $Z_i \in \*Z$, there exist nodes $(X',Y')$, possibly equal to $(X,Y)$,
and $\*Z' \in \operatorname{MinSep}_{\mathcal{P}^{(r)}}(X',Y')$
such that $Z_i \in \*Z'$ and $\langle X', Z_i, Y' \rangle$
is a non-collider triple with order corresponding to $(X',Y')$.
\end{proposition}

\begin{proposition}
\label{prop:minimality_triples_order}
Any triple with order $\langle A, B, C \rangle$ in $\mathcal{P}^{(r)}$
\emph{corresponds to} a pair of non-adjacent nodes $(X, Y)$ such that:
$\langle A, B, C \rangle$ is a non-collider triple iff 
$B \in \operatorname{MinSep}_{\mathcal{P}^{(r)}}(X, Y)$,  and
$\langle A, B, C \rangle$ is a collider triple iff 
$B \not \in \operatorname{MinSep}_{\mathcal{P}^{(r)}}(X, Y)$.
Thus, the sets $\operatorname{MinSep}_{\mathcal{P}^{(r)}}$
fully characterize all such triples.
\end{proposition}

We now define $\*H_{\mathfrak{C}({\mathcal{P}^{(r)}}), P}$,
the set of dependence hypotheses relevant to
$\mathfrak{C}({\mathcal{P}^{(r)}})$, the colliders with order in $\mathcal{P}^{(r)}$:
\begin{definition}
\label{def:col_hypotheses}
The set of dependence hypotheses relevant to
$\mathfrak{C}({\mathcal{P}^{(r)}})$, the colliders with order in $\mathcal{P}^{(r)}$, is given as follows:
$$
\begin{aligned}
& \*H_{\mathfrak{C}({\mathcal{P}^{(r)}}), P} =  \{ H_{( X \nindep Y | \*Z )_P} \mid  \exists \langle A, B, C \rangle \in \mathfrak{C}({\mathcal{P}^{(r)}}) \\
& \quad  \text{ s.t.: } \langle A, B, C \rangle \text{ corresponds to } \{X, Y\} \text{ and }  \\
& \quad \*Z = \*S \cup \{B\}, \text{ where } \*S \in \operatorname{MinSep}_{\mathcal{P}^{(r)}}(X, Y) \}.
\end{aligned}
$$
\end{definition}

\begin{theorem}[Score Completeness]
\label{thm:complete_hypotheses_uncertainty}
The set of (in)dependence hypotheses
\(\*H_{\mathcal{P}^{(r)}, P} = \*H_{\mathfrak{S}(\mathcal{P}^{(r)}), P}
\cup \*H_{\mathfrak{C}(\mathcal{P}^{(r)}), P}\),
as defined in Definitions
\ref{def:skel_hypotheses} and \ref{def:col_hypotheses},
is both sufficient and necessary to
fully characterize the uncertainty involved in both identifying \( \mathcal{P}^{(r)} \)
and discriminating it from any distinct \( r \)-PAG.
\end{theorem}

Additionally, the next Corollary follows from
\Cref{thm:complete_hypotheses_uncertainty}:
\begin{corollary}[Score Equivalence]
\label{cor:score_equivalent}
The score \( S_{\mathcal{P}^{(r)}, P} = P(\*H_{\mathcal{P}^{(r)}, P} | \mathcal{D}) \),
where \( \*H_{\mathcal{P}^{(r)}, P} \) is defined in
\Cref{thm:complete_hypotheses_uncertainty}, is score-equivalent.
Asymptotically, MAGs within \( \mathcal{P}^{(r)} \)
receive identical scores, while distinct \( r \)-PAGs
receive different scores.
\end{corollary}

Although \(\*H_{\mathcal{P}^{(r)}, P}\) fully characterizes the uncertainty
in $\mathcal{P}^{(r)}$, if often omits
relations, making some PAGs potentially incomparable.
To enable effective comparison,
their scores should account for the union of all relevant hypotheses and
focus exclusively on the hypotheses where the PAGs differ.

Let \( \mathcal{L}^{(r)} = \{ \mathcal{P}_i^{(r)} \}_{i=1}^m \) be a list of
\( m \) candidate PAGs, each associated with relevant conditional (in)dependencies
\( \*H_{\mathcal{P}^{(r)}_i, P} \). Define \( \*R_{\mathcal{L}^{(r)}} \)
as the union of these relations. Specifically:
\begin{equation}
\begin{aligned}[t]
& \*R_{\mathcal{L}^{(r)}}  = \{ (X,Y | \*Z) \mid  H_{(X \indep Y | \*Z)_P}
\in \bigcup_{i=1}^{m} \*H_{\mathcal{P}^{(r)}_i, P} \text{ or }\\
& \quad H_{(X \nindep Y | \*Z)_P} \in \bigcup_{i=1}^{m} \*H_{\mathcal{P}^{(r)}_i, P}\}.
\end{aligned}
\end{equation}
The hypotheses for each \( \mathcal{P}_i \in \mathcal{L}^{(r)} \) are
augmented as follows:
\begin{equation}
\label{eq:comp_mec_hypotheses}
\begin{aligned}
& \*H^{+}_{\mathcal{P}_i^{(r)}, P} = \{ H_{( X \indep Y | \*Z )_P }  \mid
(X, Y | \*Z) \in \*R_{\mathcal{L}^{(r)}} \text{ and } \\
& \quad (X \indep_m Y | \*Z)_{\mathcal{P}_i} \} \, \, \cup
\{ H_{( X \nindep Y | \*Z )_P}  \mid \\
& \quad (X, Y | \*Z) \in \*R_{\mathcal{L}^{(r)}} \text{ and } (X \nindep_m Y | \*Z)_{\mathcal{P}_i}  \}
\end{aligned}
\end{equation}

Define the set of
hypotheses common to all PAGs in \( \mathcal{L}^{(r)} \):
\begin{equation}
\label{eq:common_mec_hypotheses}
\*H_{\cap_{\mathcal{L}^{(r)}}, P} = \bigcap_{\mathcal{P}^{(r)}_i \in
\mathcal{L}^{(r)}} \*H^{+}_{\mathcal{P}^{(r)}_i, P}.
\end{equation}

Then, we can define a score for any \( \mathcal{P}_i^{(r)} \in \mathcal{L}^{(r)} \),
conditional on the common hypotheses in \( \mathcal{L}^{(r)} \), as follows:
\begin{equation}
\label{eq:comparable_mec_score}
S_{\mathcal{P}^{(r)}_i,\mathcal{D}, \mathcal{L}^{(r)}} =
P \left( \*H^{+}_{\mathcal{P}^{(r)}_i, P} \! \setminus \! \*H_{\cap_{\mathcal{L}^{(r)}}, P}
\mid \*H_{\cap_{\mathcal{L}^{(r)}}, P}, \mathcal{D} \right).
\end{equation}

Note that, if \( \mathcal{L}^{(r)} \) encompass all plausible $r$-PAGs consistent with
the (in)dependencies observed in \( \mathcal{D} \), conditioned on
sets of size up to \( r \), then the shared hypotheses can be considered certain
(i.e., \( P \left( \*H_{\cap_{\mathcal{L}^{(r)}}, P} \mid \mathcal{D} \right) = 1 \)).
In this case, the highest-scoring $r$-PAGs in \( \mathcal{L}^{(r)} \) can be identified
by comparing only the posterior probability of the conflicting hypotheses.

Next theorem follows directly from the preceding reasoning and
\Cref{thm:complete_hypotheses_uncertainty}, establishing a sufficient
and necessary MEC-targeted PAG score, conditional on a list of candidate \(r\)-PAGs.

\begin{theorem}[Score Completeness for Candidate Comparison]
\label{thm:score_complete_comparison}
If \( \mathcal{L}^{(r)} \) includes all $r$-PAGs consistent with
\( \mathcal{D} \), then the score of any
\( r \)-PAG \( \mathcal{P}^{(r)}_i \in \mathcal{L}^{(r)} \) simplifies to
$$
S_{\mathcal{P}^{(r)}_i,\mathcal{D}, \mathcal{L}^{(r)}} =
P \left( \*H^{+}_{\mathcal{P}^{(r)}_i, P}
\setminus \*H_{\cap_{\mathcal{L}^{(r)}}, P} \mid \mathcal{D} \right),
$$
with Fréchet bounds as in \Cref{def:straightf_compat_score},
and is both sufficient and necessary to discriminate \( \mathcal{P}^{(r)}_i \)
from any \( \mathcal{P}^{(r)}_j \in \mathcal{L}^{(r)} \).
\end{theorem}

This score not only disregards the conditional
independence hypotheses shared by
all candidate $r$-PAGs in the given list but also those irrelevant for
characterizing the $r$-MECs of candidate $r$-PAGs.
We next demonstrate how this score can guide decisions
in a novel hybrid causal discovery algorithm.

\section{Data-Compatible Fast Causal Inference (dcFCI)}

In this section, we introduce a novel hybrid causal discovery algorithm,
data-compatible Fast Causal Inference (dcFCI). It combines the
strengths of constraint-based methods, such as FCI and its variants, with our
nonparametric, MEC-targeted scoring approach. The approach is highly adaptable,
as it accommodates various data types and distributions by accepting any
likelihood-ratio-based conditional independence test.

The dcFCI algorithm follows a structure akin to the FCI algorithm and its
variants, with a main loop iterating over increasing sizes $r$ of conditioning
sets. However, a key distinction is that, rather than incrementally refining a
single PAG, it performs a greedy search over plausible PAGs, systematically
refining and selecting valid, top-scoring candidates as it progresses.
In doing so, dcFCI effectively overcomes the limitations of SOTA techniques,
particularly when dealing with limited, empirically unfaithful datasets.
Its hybrid design allows it to systematically account for uncertainties
during the learning process while ensuring
the validity of the MEC characterization and greater alignment with observed
dependencies and independencies. By refining the search for plausible causal
structures, dcFCI provides a powerful framework for causal discovery in complex,
real-world settings, offering both accuracy and robustness with enhanced flexibility.

\begin{algorithm}
\caption{dcFCI Algorithm}
\label{alg:dcfci}
\begin{algorithmic}[1]
\Require Data set $\mathcal{D}$, maximum size for the separating sets $r_{\text{max}}$, significance level $\alpha$, number $k$ of selected PAGs at each iteration.

\State \textbf{Step 1: Initialize list of candidate PAGs}

\State $ p \gets $ number of variables in $\mathcal{D}$, i.e., $|\mathbf{V}|$.
\State $\mathcal{L}^{(-1)} \gets \{\mathcal{P}^{(-1)}\}$ \Comment{Initialize with the complete graph}
\State $r \gets 0$ \Comment{Initialize separator size}

\While{$r \leq \operatorname{min}\{r_{\text{max}}, p-2\}$}
    \State \textbf{Step 2: Construct candidate $(r)$-PAGs}
    \State $\mathcal{L}^{(r)} \gets \mathcal{L}^{(r-1)}$ \Comment{Carry over previous PAGs}

    \For{each $\mathcal{P}_{i}^{(r)}$ in $\mathcal{L}^{(r)}$}
        \State  $\operatorname{PotMinSeps}_i^{(r)} \gets $ empty list

        \For{each pair of adjacent nodes $(X, Y)$ in $\mathcal{P}_{i}^{(r)}$}
            \For{each \textit{plausible} separating set $\mathbf{S} \subseteq \mathbf{V} \setminus \{X, Y\}$ with $|\mathbf{S}| = r$}
                \If{$(X \indep Y | \mathbf{S})_{\mathcal{D}}$ with p-value $> \alpha$ or $P(H_{(X \indep Y | \mathbf{S})_{P}}) > 0.5$}
                    \State Add $\mathbf{S}$ to $\operatorname{PotMinSeps}_i^{(r)}[\{X, Y\}]$
                \EndIf
            \EndFor
        \EndFor

        \State $\operatorname{SepSetsList}_i^{(r)} \gets \operatorname{PowerSet(\operatorname{PotMinSeps}_i^{(r)})}$
        \For{each $\operatorname{SepSet}_{ij}$ in $\operatorname{SepSetsList}_i^{(r)}$}
            \State Build $\mathcal{P}^{(r)}_{ij}$ from $\mathcal{P}^{(r)}_i$
            by cutting edges between separated nodes in $\operatorname{SepSet}_{ij}$ and applying Zhang's rules.
           \If{$\mathcal{P}^{(r)}_{ij}$ is a valid PAG and captures exactly the selected minimal m-separations}
                \State Add $\mathcal{P}^{(r)}_{ij}$ to $\mathcal{L}^{(r)}$
            \EndIf
        \EndFor
    \EndFor

    \State \textbf{Step 3: Score All Valid $(r)$-PAG Candidates}
    \For{each $\mathcal{P}^{(r)}_i$ in $\mathcal{L}^{(r)}$}
        \State Estimate Fréchet bounds for score
        $S_{\mathcal{P}_{i}^{(r)},\mathcal{D}, \mathcal{L}^{(r)}}$
    \EndFor

    \State \textbf{Step 4: Select Best $(r)$-PAG Candidates}
    \State Sort $\mathcal{L}^{(r)}$ by the upper bound of each PAG's score
    \State $\mathcal{L}^{(r)} \gets$ top $k$ PAGs in $\mathcal{L}^{(r)}$
    \State $r \gets r + 1$
\EndWhile

\State \Return $\langle \mathcal{L}^{(r)},
S_{\mathcal{P}^{(r)},\mathcal{D},\mathcal{L}^{(r)}} \rangle$
\end{algorithmic}
\end{algorithm}

The pseudo-algorithm of dcFCI is shown in Algorithm \ref{alg:dcfci}. It starts
with a list \( \mathcal{L}^{(-1)} \) containing the complete PAG over $p$
variables. In each iteration \(r = 0, \ldots, r_{\text{max}} \),
where \( r_{\text{max}} \leq p-2 \),
dcFCI initializes a list \( \mathcal{L}^{(r)} \) of candidate \( r \)-PAGs
by carrying over the PAGs from the previous interation's list
\( \mathcal{L}^{(r-1)} \). This ensures a
consistent and incremental refinement of the candidate PAGs.
For each PAG \( \mathcal{P}_i^{(r)} \) in \( \mathcal{L}^{(r)} \),
the algorithm evaluates all pairs of adjacent nodes \( \{X, Y\} \) and
identifies a set of potential minimal separating sets
\( \text{PotMinSeps}^{(r)}_i[\{X, Y\}] \).
These sets, \( \mathbf{S} \), with \( |\mathbf{S}| = r \), are selected based
on their plausibility as separators in the oriented
\( \mathcal{P}_i^{(r)} \) and the satisfaction of
\( (X \indep Y \mid \mathbf{S}) \),
either with the p-value above the specified significance level $\alpha$, or
with \( P(H_{(X \indep Y \mid \mathbf{S})_{P}}) > 0.5 \).
Note that \(\alpha\) is used solely to select potential conditional
independencies, with lower \(\alpha\) values admitting more candidate separators.
Thus, dcFCI relies minimally on \(\alpha\), using it primarily to define the
search space.

Rather than updating the PAG skeleton
with all identified minimal separators, as in FCI,
dcFCI constructs all possible $r$-PAGs from
subsets of \( \text{PotMinSeps}^{(r)}_i \), refining the skeleton and
applying orientation rules. It then
retains only valid PAGs that
accurately encode via m-separation the used minimal conditional independencies,
preventing conflicts due to
conditioning set order
\citep{colombo2014order}.
Each PAG in \( \mathcal{L}^{(r)} \) is then scored
(\Cref{thm:score_complete_comparison}), and the top \( k \)
are selected for output or the next iteration.

\begin{theorem}[dcFCI Soundness]
\label{thm:dcfci_soundness}
Given a dataset \( \mathcal{D} \) for \( \*V \) with \( |\*V| = p \) and a
sufficiently small significance level \( \alpha \), if, for each \( r = 0, \dots,
r_{\text{max}} \), where \( r_{\text{max}} \leq p-2 \), the true \( \mathcal{P}^{(r)} \)
is among the $k$ most data-compatible \( r \)-PAGs,
(according to $S_{\mathcal{P}^{(r)}, \mathcal{D}, \mathcal{L}^{(r)}}$)
then dcFCI's output includes
the true \( \mathcal{P}^{(r_{\text{max}})} \).
\end{theorem}

The following corollary establishes a weaker notion of empirical faithfulness
that is sufficient (but not necessary) to guarantee that dcFCI outputs the
true PAG with \( k=1 \).

\begin{corollary}
\label{cor:weak_faithfulness}
If, for each \( r = 0, \dots, r_{\text{max}} \),
$\forall H_i \in
\*H_{\mathcal{P}^{(r)}, P} = \*H_{\mathfrak{S}(\mathcal{P}^{(r)}), P} \cup
\*H_{\mathfrak{C}(\mathcal{P}^{(r)}), P}$
(Definitions \ref{def:skel_hypotheses} and \ref{def:col_hypotheses})
satisfies \( P(H_i \mid \mathcal{D}) \geq 0.5 \), then dcFCI outputs the true
\( \mathcal{P}^{(r_{\text{max}})} \) with \( k=1 \).
\end{corollary}

Besides setting \( k \) to a small value (e.g., \( k=1 \) or \( 2 \)),
potential PAGs are constructed in parallel to manage computational complexity.
Also, dcFCI can also be halted earlier by setting \( r_{\max} < p-2 \),
reducing reliance on conditional independence tests with larger sets,
which often lack statistical power. This enables the inference of more reliable,
though potentially less informative, PAGs. The complexity analysis of dcFCI
and optimization strategies are detailed in \Cref{ape:dcfci_complexity}.

Importantly, dcFCI outputs the highest-scoring PAG(s)
and provides access to all evaluated PAGs, enabling
uncertainty quantification by allowing users to
compare scores across different causal hypotheses.
This facilitates more informed decisions based on the most robust
orientations.

In summary, dcFCI stands out as a powerful and versatile causal discovery
algorithm for real-world applications by:
1) effectively addressing latent confounding, enabling causal learning given data
from any subset of variables;
2) accommodating a wide range of datasets, including those with mixed data types
and distributions, ensuring broad applicability;
3) providing greater robustness to empirical unfaithfulness by evaluating data
compatibility and ensuring the validity of MEC characterization of the output PAG,
thus supporting downstream causal analyses designed for PAGs; and
4) enabling uncertainty quantification, allowing users to explore and assess the
reliability of a range of plausible causal hypotheses.
Table \ref{tab:algorithm_features} compares dcFCI's features with other
SOTA algorithms.
The parentheses in BCCD's handling of latent confounding reflect its scoring
limitation, which assumes causal sufficiency.

\begin{table}[h]
\centering
\caption{Comparison of dcFCI's features with SOTA algorithms.}
\setlength{\tabcolsep}{1pt}
\begin{tabular}{l|x{1.7cm}|x{1.5cm}|x{1.5cm}|x{1.7cm}}
  \hline
  \multirow{2}{*}{Algorithm} & Latent Confounding & Mixed Data Type & MEC Validity & Uncertainty Quantification \\
  \hline
  \hline
  FCI \citep{zhang2008fci} &  \cmark & \cmark & \xmark & \xmark  \\
  cFCI \citep{colombo2012learning} &  \cmark & \cmark & \xmark & \xmark \\
  BCCD \citep{claassen2012bccd} &  (\cmark) & \xmark & \xmark & \xmark \\
  DCD \citep{bhattacharya2021differentiable} & \cmark & \xmark & \cmark & \xmark \\
  MAGSL  \citep{rantanen2021maximal} & \cmark & \xmark & \cmark & \xmark \\
  GPS  \citep{claassen2022greedy} & \cmark & \xmark & \xmark & \xmark \\
  dcFCI &  \cmark & \cmark & \cmark & \cmark  \\
  \hline
\end{tabular}
\label{tab:algorithm_features}
\end{table}

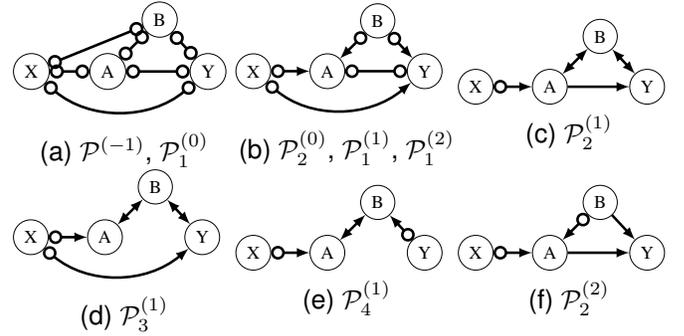
\begin{figure}[htbp]
    \centering
    \begin{minipage}{\linewidth}

    \begin{minipage}{0.32\linewidth}
    \subfloat[$\mathcal{P}^{(-1)}$, $\mathcal{P}_1^{(0)}$]{
        \centering
        \resizebox{\linewidth}{!}{%
        \begin{tikzpicture}[node distance=0.75cm, auto]
        \tikzset{vertex/.style = {shape=circle, draw, minimum size=1.5em}}
          \node (X) [circle, draw] {X};
          \node (A) [circle, draw, right=of X] {A};
          \node (Y) [circle, draw, right=1.2cm of A] {Y};
          \node (B) [circle, draw, above right=0.7cm of A] {B};
          \draw[o-o, line width=0.5mm] (A) -- (B);
          \draw[o-o, line width=0.5mm] (X) -- (A);
          \draw[o-o, line width=0.5mm] (X) -- (B);
          \draw[o-o, line width=0.5mm, bend right=40] (X) to (Y);
          \draw[o-o, line width=0.5mm] (B) -- (Y);
          \draw[o-o, line width=0.5mm] (A) -- (Y);
        \end{tikzpicture}}
        \label{fig:dcfci_ord0_1}
        }
    \end{minipage}
    \begin{minipage}{0.32\linewidth}
    \subfloat[$\mathcal{P}_2^{(0)}$, $\mathcal{P}_1^{(1)}$, $\mathcal{P}_1^{(2)}$]{
        \centering
        \resizebox{\linewidth}{!}{%
        \begin{tikzpicture}[node distance=0.75cm, auto]
          \node (X) [circle, draw] {X};
          \node (A) [circle, draw, right=of X] {A};
          \node (Y) [circle, draw, right=1.2cm of A] {Y};
          \node (B) [circle, draw, above right=0.7cm of A] {B};
          \draw[o-o, line width=0.5mm] (A) -- (Y);
          \draw[o-latex, line width=0.5mm] (B) -- (Y);
          \draw[o-latex, line width=0.5mm] (B) to (A);
          \draw[o-latex, line width=0.5mm] (X) -- (A);
          \draw[o-latex, line width=0.5mm, bend right=40] (X) to (Y);
        \end{tikzpicture}
        }
        \label{fig:dcfci_ord0_2}
    }
    \end{minipage}
    \begin{minipage}{0.32\linewidth}
    \subfloat[$\mathcal{P}_2^{(1)}$]{
        \centering
        \resizebox{\linewidth}{!}{%
        \begin{tikzpicture}[node distance=0.75cm, auto]
          \node (X) [circle, draw] {X};
          \node (A) [circle, draw, right=of X] {A};
          \node (Y) [circle, draw, right=1.2cm of A] {Y};
          \node (B) [circle, draw, above right=0.7cm of A] {B};
          \draw[-latex, line width=0.5mm] (A) -- (Y);
          \draw[latex-latex, line width=0.5mm] (B) -- (Y);
          \draw[latex-latex, line width=0.5mm] (B) to (A);
          \draw[o-latex, line width=0.5mm] (X) -- (A);
        \end{tikzpicture}
        }
        \label{fig:dcfci_ord1_2}
    }
    \end{minipage}
    \end{minipage}
    \begin{minipage}{\linewidth}
    \begin{minipage}{0.32\linewidth}
    \subfloat[$\mathcal{P}_3^{(1)}$]{
        \centering
        \resizebox{\linewidth}{!}{%
        \begin{tikzpicture}[node distance=0.75cm, auto]
          \node (X) [circle, draw] {X};
          \node (A) [circle, draw, right=of X] {A};
          \node (Y) [circle, draw, right=1.2cm of A] {Y};
          \node (B) [circle, draw, above right=0.7cm of A] {B};
          \draw[latex-latex, line width=0.5mm] (B) -- (Y);
          \draw[latex-latex, line width=0.5mm] (B) to (A);
          \draw[o-latex, line width=0.5mm] (X) -- (A);
          \draw[o-latex, line width=0.5mm, bend right=40] (X) to (Y);
        \end{tikzpicture}
        }
        \label{fig:dcfci_ord1_3}
    }
    \end{minipage}
    \begin{minipage}{0.32\linewidth}
    \subfloat[$\mathcal{P}_4^{(1)}$]{
        \centering
        \resizebox{\linewidth}{!}{%
        \begin{tikzpicture}[node distance=0.75cm, auto]
          \node (X) [circle, draw] {X};
          \node (A) [circle, draw, right=of X] {A};
          \node (Y) [circle, draw, right=1.2cm of A] {Y};
          \node (B) [circle, draw, above right=0.7cm of A] {B};
          \draw[latex-o, line width=0.5mm] (B) -- (Y);
          \draw[latex-latex, line width=0.5mm] (B) to (A);
          \draw[o-latex, line width=0.5mm] (X) -- (A);
        \end{tikzpicture}
        }
        \label{fig:dcfci_ord1_4}
    }
    \end{minipage}
    \begin{minipage}{0.32\linewidth}
    \subfloat[$\mathcal{P}_2^{(2)}$]{
        \centering
        \resizebox{\linewidth}{!}{%
        \begin{tikzpicture}[node distance=0.75cm, auto]
          \node (X) [circle, draw] {X};
          \node (A) [circle, draw, right=of X] {A};
          \node (Y) [circle, draw, right=1.2cm of A] {Y};
          \node (B) [circle, draw, above right=0.7cm of A] {B};
          \draw[-latex, line width=0.5mm] (A) -- (Y);
          \draw[-latex, line width=0.5mm] (B) -- (Y);
          \draw[o-latex, line width=0.5mm] (B) to (A);
          \draw[o-latex, line width=0.5mm] (X) -- (A);
        \end{tikzpicture}
        }
        \label{fig:dcfci_ord2}
    }
    \end{minipage}
    \end{minipage}
    \caption{All candidate $r$-PAGs generated by dcFCI in Example 1 for
    $r = 0, 1, 2$. (a) initial PAG $\mathcal{P}^{(-1)}$ and 0-PAG $\mathcal{P}_1^{(0)}$,
    implying the set of (conditional) independencies
    $\mathscr{I}^{(-1)} = \mathscr{I}_1^{(0)} = \emptyset$;
    (b) $\mathcal{P}_2^{(0)}$, $\mathcal{P}_1^{(1)}$, and $\mathcal{P}_1^{(2)}$,
    implying $\mathscr{I}_2^{(0)} = \mathscr{I}_1^{(1)} = \mathscr{I}_1^{(2)} = {(B \indep X)}$;
    (c)-(e) 1-PAGs $\mathcal{P}_2^{(1)}$, $\mathcal{P}_3^{(1)}$, and $\mathcal{P}_4^{(1)}$,
    derived from $\mathcal{P}_2^{(0)}$, implying, respectively,
    $\mathscr{I}_2^{(1)} = \mathscr{I}_2^{(0)} \cup {(X \indep Y | A)}$,
    $\mathscr{I}_3^{(1)} = \mathscr{I}_2^{(0)} \cup {(A \indep Y | X)}$, and
    $\mathscr{I}_4^{(1)} = \mathscr{I}_2^{(0)} \cup {(X \indep Y | A), (A \indep Y | X)}$;
    (f) 2-PAG $\mathcal{P}_2^{(2)}$, derived from $\mathcal{P}_1^{(1)}$, with
    $\mathscr{I}_2^{(2)} = \mathscr{I}_1^{(1)} \cup {(X \indep Y | A, B)}$,
    representing dcFCI’s optimal and true PAG.}
    \label{fig:dcfci_steps}
\end{figure}

\begin{exmp} Consider the example from Figure \ref{fig:ex_1},
with conditional independence test results in
Table \ref{tab:ex1_citests}.
As noted in the Introduction,
MAGSL, which performs an exact search,
successfully recovers the true PAG.
In contrast, all other SOTA algorithms
(FCI, cFCI, BCCD, DCD, and GPS) failed to do so.
As shown below, dcFCI also recovers the true PAG.

The algorithm starts with a complete graph, with all nodes connected by
circle-circle edges (see line 3 of the pseudo-algorithm).
At iteration \( r =  0 \), all 0-PAGs are constructed.
Consider $\alpha = 0.01$.
The only observed marginal independence is \( (B \indep X) \),
with a p-value of 0.496, \( P_{H_0} = 0.671 \),
and \( P_{H_1} = 0.328 \). Thus, two candidate 0-PAGs are identified:

\begin{itemize}
    \item PAG \( \mathcal{P}_1^{(0)} \): No independence applied, yielding
    the PAG in \Cref{fig:dcfci_ord0_1}, with score
    \( S_{\mathcal{P}_{1}^{(0)},\mathcal{D}} = [0.328, 0.328] \).
    \item PAG \( \mathcal{P}_2^{(0)} \): Independence \( (B \indep X) \)
    applied, producing the PAG in \Cref{fig:dcfci_ord0_2}, with score
    \( S_{\mathcal{P}_{2}^{(0)},\mathcal{D}} = [0.671, 0.671] \).
\end{itemize}

Only the top \( k \) scoring 0-PAGs proceed to iteration \( r =  1 \).
With \( k = 1 \), dcFCI retains only \( \mathcal{P}_2^{(0)} \) and
evaluates conditional independencies given sets of size \( r =  1 \).
It identifies \( (X \indep Y \mid A) \), with a p-value of 0.105,
\( P_{H_0} = 0.388 \), and \( P_{H_1} = 0.612 \), as well as \( (A \indep Y \mid X) \),
with a p-value of 0.0270, \( P_{H_0} = 0.198 \), and \( P_{H_1} = 0.802 \).
These yield four candidate 1-PAGs, differing only
in the inclusion of these two independencies:
\begin{itemize}
    \item PAG \( \mathcal{P}_1^{(1)} \): None is applied to $\mathcal{P}_2^{(0)}$,
    yielding the PAG in \Cref{fig:dcfci_ord0_2}, with score
    \( S_{\mathcal{P}_{1}^{(1)},\mathcal{D}} = [0.413, 0.611] \),
    reflecting the probabilities of $(X \nindep Y | A)$ and $(A \nindep Y | X)$.
    \item PAG \( \mathcal{P}_2^{(1)} \): $(X \indep Y | A)$
    is applied to $\mathcal{P}_2^{(0)}$, yielding the PAG
    in \Cref{fig:dcfci_ord1_2}, with score
    \( S_{\mathcal{P}_{2}^{(1)},\mathcal{D}} = [0.190, 0.388]\),
    reflecting the probabilities of $(X \indep Y | A)$ and $(A \nindep Y | X)$.
    \item PAG \( \mathcal{P}_3^{(1)} \): $(A \indep Y | X)$
    is applied to $\mathcal{P}_2^{(0)}$, yielding the PAG in
    \Cref{fig:dcfci_ord1_2}, with score
    \( S_{\mathcal{P}_{3}^{(1)},\mathcal{D}} = [0.000, 0.198] \), reflecting
    the probabilities of $(X \nindep Y | A)$ and $(A \indep Y | X)$.
    \item PAG \( \mathcal{P}_4^{(1)} \): Both
    \( (X \indep Y | A) \) and \( (A \indep Y | X) \) are applied to $\mathcal{P}_2^{(0)}$.
    However, the resulting PAG introduces unintended
    m-separations, implying \( ( A \indep Y ) \) and \( ( X \indep Y ) \), and
    therefore, contradicting the minimality of the applied independencies.
    Thus, this candidate is discarded.
\end{itemize}

As before, only the top \( k \) scoring 1-PAGs proceed to the next iteration.
At this stage, the algorithm retains only \( \mathcal{P}_1^{(1)} \) and evaluates
conditional independencies given sets of size \( r =  2 \). The only
observed independence is \( (X \indep Y \mid A, B) \), with a p-value of
0.536, \( P_{H_0} = 0.683 \), and \( P_{H_1} = 0.317 \).
Based on this, two candidate 2-PAGs are identified:

\begin{itemize}
    \item PAG \( \mathcal{P}_1^{(2)} \): No independence is applied to $\mathcal{P}_1^{(1)}$,
    yielding the PAG in \Cref{fig:dcfci_ord0_1}, with score
    \( S_{\mathcal{P}_{1}^{(2)},\mathcal{D}} = [0.317, 0.317] \),
    reflecting the probability of $(X \nindep Y | A, B)$.
    \item PAG \( \mathcal{P}_2^{(2)} \): $(X \indep Y | A, B)$
    is applied to $\mathcal{P}_1^{(1)}$, yielding the PAG in \Cref{fig:dcfci_ord2},
    with score \( S_{\mathcal{P}_{2}^{(0)},\mathcal{D}} = [0.683, 0.683] \),
    reflecting the probability of this independence.
\end{itemize}

At this point, the while loop in line 5 terminates without proceeding to \( r =  3 \).
The highest-scoring PAG in \( \mathcal{L}^{(2)} \), \( \mathcal{P}_2^{(2)} \),
is selected and returned as the sole estimated PAG by the dcFCI algorithm.
Notably, this PAG corresponds to the true PAG.
\end{exmp}

\section{Simulation Studies}
\label{sec:simulations}

We evaluate the performance of the dcFCI against
SOTA algorithms -- FCI, cFCI, BCCD, DCD, and GPS. To assess the
accuracy of the inferred PAGs, we use
the Structural Hamming Distance (SHD), along with the
False Discovery Rate (FDR) and False Omission Rate (FOR).
We also compare the algorithms' ability to
recover the true PAG, ensure the output validity as a MEC,
and recover structures compatible with the data.
For the latter, we use the straightforward data-PAG compatibility score,
as defined in \Cref{def:straightf_compat_score},
which measures how well the inferred PAG aligns with the data,
independent of a candidate PAG list. Details on these metrics and the execution of
each algorithm are provided in \Cref{sec:simulation_details}.

We consider two sets of simulations: one in \Cref{sec:gaussian_sims},
focused on causal discovery with Gaussian variables, as existing score-based
approaches are implemented exclusively for this setting, and another in
\Cref{sec:mixed_sims}, focused on datasets with mixed
(continuous, binary and multinomial) data types.

For conditional independence testing, we use likelihood-ratio tests
for diverse variable types, implemented in our dcFCI R package, with a
significance threshold of $\alpha = 0.05$.
These tests rely on generalized linear regression models, following
the approach by \citep{tsagris2018constraint}, detailed in
Sections \ref{sec:lr-tests} and \ref{ape:likelihood_ratio_tests}.

For simulations with Gaussian datasets, we further assessed the
algorithms' performance using both the BIC
distance, which measures how closely the BIC of the inferred model aligns with
that of the true model, and the BIC of the inferred model itself, a widely used
goodness-of-fit metric for linear Gaussian SEMs.
These additional metrics offer deeper insights into not only the accuracy and
data compatibility of the inferred PAGs, but also the broader implications of
leveraging the BIC score in causal discovery.

Finally, while this work focuses on improving robustness under empirical
unfaithfulness rather than scalability we also compare
dcFCI's time efficiency for completeness.

\subsection{Using Gaussian Datasets}
\label{sec:gaussian_sims}

Our first set of simulations focuses on causal discovery of Gaussian
linear Structural Equation Models (SEMs), as score-based
approaches -- BCCD, DCD, MAGSL, and GPS -- are implemented exclusively
under this assumption.

For this study, we randomly generated 10 unique PAGs, each comprising
5 nodes and featuring a diverse combination of directed and bidirected edges.
Then, for each PAG and sample sizes of \( N = \{1000, 5000, 10000, 50000\} \),
30 distinct datasets were generated based on Gaussian linear SEMs adhering to
the structure of a valid causal diagram within the the PAG.
This was achieved using the \texttt{simulateSEM} function from the
\texttt{dagitty} R package \citep{textor2016robust}.
The model coefficients were randomly assigned, but their absolute values were
constrained to exceed 0.2 to prevent very weak associations.
As a result, this process yielded a random sample of 300 potentially
unfaithful Gaussian datasets for each considered sample size.

Despite its greedy approach, dcFCI captures uncertainty in the causal orientations,
potentially outputting multiple PAGs equally compatible with the data.
As the sample size increases and uncertainty decreases, the algorithm tends to
return a single PAG. Details on the distribution of PAGs are in \Cref{sec:app_sim_gaussian}.

We begin the comparison of dcFCI with SOTA algorithms by presenting in Table
\ref{tab:gaussian_valid_PAGs} their ability to infer valid PAGs.
A PAG is valid if it represents the expected MEC of a valid MAG or causal diagram
(see further details in \Cref{sec:simulation_details}).

\begin{table}[h]
\centering
\caption{Percentage of valid inferred PAGs from Gaussian datasets.}
\begin{tabular}{c|c|c|c|c|c|c|c}
  \hline
N & 
FCI & cFCI & BCCD & DCD & MAGSL & GPS & dcFCI \\
  \hline
  \hline
  \multirow{1}{*}{1,000} & 97.7  &  46.7  &  68.7  &  \bf 100.0  &  \bf 100.0  &  96.7  &  \bf 100.0  \\
  \multirow{1}{*}{5,000}  & 96.3  &  49.3  &  74.7  &  \bf 100.0  &  \bf 100.0  &  91.3  &  \bf 100.0  \\
  \multirow{1}{*}{10,000}  & 97.0  &  67.0  &  77.7  &  \bf 100.0  &  \bf 100.0  &  88.0  &  \bf 100.0  \\
  \multirow{1}{*}{50,000}  & 99.7  &  94.3  &  87.0  &  \bf 100.0  &  \bf 100.0  &  86.7  &  \bf 100.0  \\
  \hline
\end{tabular}
\label{tab:gaussian_valid_PAGs}
\end{table}

Notably, FCI and GPS occasionally generate invalid PAGs
when applied to unfaithful data.
Since both cFCI and BCCD determine edge orientations without accounting for
the overall MEC characterization and the implications of m-separation,
they often produce invalid PAGs, even with large datasets. In contrast, since
DCD and MAGSL are specifically designed to first learn a MAG and
then convert it into a PAG, the resulting PAG is always valid.
For dcFCI, which may output a list of compatible PAGs rather than a single one,
we considered the output valid only if all PAGs in the list were valid.
Remarkably, dcFCI exclusively infers
valid PAGs, as its design inherently enforces this restriction.

Table \ref{tab:gaussian_true_pag_rec_rate} shows the recovery rate (percentage) of
the true PAG across increasing sample sizes.
For dcFCI, we provide percentages for three cases: when the output list contains
only the true PAG, when it includes the true PAG with others having the same
upper score bound, and when it includes the true PAG with others having
overlapping score bounds (i.e., statistically equally plausible PAGs).
A hyphen is used to avoid repetition when percentages do not change across cases.

\begin{table}[ht]
\centering
\setlength{\tabcolsep}{1.5pt}
\caption{True PAG recovery rate in simulations with Gaussian datasets.}
\begin{tabular}{c|c|c|c|c|c|c|x{1.2cm}|x{1.8cm}}
  \hline
\multirow{2}{*}{N} & \multirow{2}{*}{FCI} & \multirow{2}{*}{cFCI} & \multirow{2}{*}{BCCD} &
\multirow{2}{*}{DCD} &\multirow{2}{*}{MAGSL} & \multirow{2}{*}{GPS} & dcFCI ($k=1$) & dcFCI \newline ($k=2$) \\
  \hline
 \hline
  \multirow{1}{*}{1,000} & 3.3 & 0.3 & 0 & 2 & \bf 10.3 & 5.3 & 8.3/8.7/- & \bf 8.3/-/13.3 \\
  \multirow{1}{*}{5,000} & 16.3 & 5 & 0.3 & 0.3 & 30.3 & 12 & \bf 32.7/33/- & \bf 32.7/35.7/51.7 \\
  \multirow{1}{*}{10,000} & 56.3 & 33.7 & 0 & 2.7 & \bf 78.3 & 28 & 74.7/-/- & \bf 80.3/81.7/85.3 \\
  \multirow{1}{*}{50,000} & 93.3 & 86.3 & 0 & 1.3 & \bf 98.3 & 33 & \bf  98.3/-/- & \bf 98.67/-/- \\
   \hline
\end{tabular}
\label{tab:gaussian_true_pag_rec_rate}
\end{table}

Interestingly, dcFCI and MAGSL achieve comparable results with
recovery rates substantially higher than those of other algorithms,
even in low-sample settings.
Notably, dcFCI outperforms MAGSL (which performs an exact search)
in some cases with \( k=1 \) and consistently with \( k=2 \), demonstrating
its effectiveness in identifying the true PAG with its greedy approach,
while still accounting for uncertainty.
Notably, for five-variable graphs, 1,000 samples, often the practical limit,
is usually insufficient to recover the true PAG. With 50,000 samples,
MAGSL and dcFCI succeed nearly always, while FCI and cFCI perform slightly worse,
and other algorithms struggle even with larger datasets.

\Cref{tab:gaussian_faith_pag_rec_rate} shows true PAG recovery rates for varying degrees
of empirical faithfulness. Under full faithfulness based on p-values
(all and only dependencies with p-value $\leq \alpha$),
FCI, cFCI, MAGSL, and dcFCI always recover the true PAG. However,
achieving this level of faithfulness is very difficult, even with large samples.
Full faithfulness based on posterior probabilities
(all and only dependencies with \( P(H_1 | \mathcal{D}) \geq 0.5) \)) is more
attainable but still challenging.
Many more datasets, however, satisfy \( r \)-MEC faithfulness,
where all key (in)dependencies for the skeleton and colliders with order hold.
In these cases, as stated in \Cref{cor:weak_faithfulness},
dcFCI consistently recovers the true PAG, unlike the SOTA methods.
Importantly, \( r \)-MEC faithfulness is a sufficient, but not necessary
condition. As shown in the last column of \Cref{tab:gaussian_faith_pag_rec_rate},
dcFCI can still recover the true PAG even when some of these essential
(in)dependencies are weakly supported by the data.

\Cref{tab:gaussian_all_sims_shd} compares dcFCI with SOTA algorithms in terms of
SHD, while \Cref{tab:gaussian_all_sims_fdr} and \Cref{tab:gaussian_all_sims_for}
present FDR and FOR, respectively. For each sample size, we
report the minimum (Min), first quartile (Q1), median (Med), mean, third quartile (Q3),
and maximum (Max) values across 300 simulations, with lower values indicating
better accuracy. The second-to-last column shows the p-value from a sign test
comparing each SOTA algorithm with dcFCI ($k=1$). Significant improvements
(i.e., significantly lower values for dcFCI) are marked by a green downward arrow in the last column.

\begin{table}[h!]
\centering
\setlength{\tabcolsep}{3pt}
\caption{Comparison in terms of Structural Hamming Distance (SHD) in simulations with Gaussian variables.
Lower SHD indicates more accurate edges.
Green downward arrows indicate significant improvement of dcFCI over SOTA algorithms.}
\begin{tabular}{l|c|c|c|c|c|c|c|c|c}
  \hline
  \multirow{2}{*}{Algorithm} & \multirow{2}{*}{N} & \multicolumn{6}{c|}{SHD}  & \multicolumn{2}{c}{Comparison}  \\
 & & Min & Q1 & Med & Mean & Q3 & Max & p-value & Diff\\
  \hline
 \hline
  FCI & \multirow{7}{*}{$1,000$} &   0 &   4 &   5 & 5.3 &   7 &  12 & 0.097 & -- \\
  cFCI &  &   0 & 5.8 &   7 & 7.2 &   9 &  12 & 1.9e-11 & $\color{PineGreen}{\boldsymbol\downarrow}$ \\
  BCCD &  &   2 &   5 &   7 & 6.8 &   9 &  15 & 6.1e-08 & $\color{PineGreen}{\boldsymbol\downarrow}$ \\
  DCD &  &   0 &   6 &   8 &   8 &  10 &  16 & 4.9e-19 & $\color{PineGreen}{\boldsymbol\downarrow}$ \\
  MAGSL &  &   0 &   3 &   5 &   5 &   7 &  14 & 0.125 &  -- \\
  GPS &  &   0 &   5 &   7 & 6.5 &   9 &  15 & 4.3e-06 & $\color{PineGreen}{\boldsymbol\downarrow}$ \\
  dcFCI &  &   0 &   3 &   5 & 5.6 &   8 &  15 &   &  \\
  \hline
  FCI & \multirow{7}{*}{$5,000$} &   0 &   2 &   4 & 3.9 &   6 &  13 & 0.021 & $\color{PineGreen}{\boldsymbol\downarrow}$ \\
  cFCI &  &   0 &   4 &   6 & 5.9 &   8 &  13 & 6.9e-24 & $\color{PineGreen}{\boldsymbol\downarrow}$ \\
  BCCD &  &   0 &   4 & 5.5 & 5.8 &   8 &  14 & 2.7e-24 & $\color{PineGreen}{\boldsymbol\downarrow}$ \\
  DCD &  &   0 &   7 &   9 & 8.4 &  10 &  14 & 4.5e-47 & $\color{PineGreen}{\boldsymbol\downarrow}$ \\
  MAGSL &  &   0 &   0 &   3 & 3.4 &   5 &  14 & 0.45 & -- \\
  GPS &  &   0 &   3 &   6 & 5.6 &   8 &  17 & 1.3e-19 & $\color{PineGreen}{\boldsymbol\downarrow}$ \\
  dcFCI &  &   0 &   0 &   3 & 3.5 &   5 &  16 &   &  \\
  \hline
  FCI & \multirow{7}{*}{$10,000$} &   0 &   0 &   0 & 1.7 &   3 &  13 & 3.3e-07 & $\color{PineGreen}{\boldsymbol\downarrow}$ \\
  cFCI &  &   0 &   0 &   2 & 2.9 &   5 &  13 & 3.3e-26 & $\color{PineGreen}{\boldsymbol\downarrow}$ \\
  BCCD &  &   2 &   2 &   5 & 5.1 &   7 &  16 & 2.4e-56 & $\color{PineGreen}{\boldsymbol\downarrow}$ \\
  DCD &  &   0 &   6 &   8 & 7.8 &  10 &  18 & 1.3e-76 & $\color{PineGreen}{\boldsymbol\downarrow}$ \\
  MAGSL &  &   0 &   0 &   0 & 0.82 &   0 &  15 & 0.45 & -- \\
  GPS &  &   0 &   0 &   3 & 3.7 & 6.2 &  14 & 5.9e-30 & $\color{PineGreen}{\boldsymbol\downarrow}$ \\
  dcFCI &  &   0 &   0 &   0 &   1 &   2 &  15 &   &  \\
  \hline
  FCI & \multirow{7}{*}{$50,000$} &   0 &   0 &   0 & 0.27 &   0 &   9 & 6.1e-05 & $\color{PineGreen}{\boldsymbol\downarrow}$ \\
  cFCI &  &   0 &   0 &   0 & 0.42 &   0 &   9 & 1.5e-11 & $\color{PineGreen}{\boldsymbol\downarrow}$ \\
  BCCD &  &   2 &   4 &   6 & 6.7 &  10 &  13 & 9.8e-91 & $\color{PineGreen}{\boldsymbol\downarrow}$ \\
  DCD &  &   0 &   6 &   9 & 8.1 &  10 &  15 & 1.6e-89 & $\color{PineGreen}{\boldsymbol\downarrow}$ \\
  MAGSL &  &   0 &   0 &   0 & 0.08 &   0 &   8 & 1 & -- \\
  GPS &  &   0 &   0 &   2 & 2.9 &   4 &  14 & 5.9e-53 & $\color{PineGreen}{\boldsymbol\downarrow}$ \\
  dcFCI &  &   0 &   0 &   0 & 0.05 &   0 &   5 &   &  \\
  \hline
\end{tabular}
\label{tab:gaussian_all_sims_shd}
\end{table}

The results for SHD, FDR, and FOR further confirm that dcFCI performs on par
with MAGSL, while significantly outperforming all other SOTA  algorithms across
all scenarios. Notably, dcFCI consistently exhibits a superior performance as
the sample size increases. With $1,000$ and $5,000$ samples, it achieves
reasonable accuracy, while with $10,000$ and $50,000$ samples, the accuracy
improves substantially, with error rates dropping to zero at the median.
 FCI, cFCI, and GPS also show improvement with increasing sample sizes,
 although GPS's performance improvement is less pronounced. In contrast,
 BCCD and DCD remain among the worst-performing algorithms throughout all
 considered scenarios.

For comparisons based on the data-PAG compatibility score, which requires
identifying conditional independencies implied by m-separation, as well as
BIC distance and actual BIC, which involve selecting a valid MAG within the
given PAG, the output PAG must be valid.
Tables \ref{tab:gaussian_valid_sims_pag_score}, \ref{tab:gaussian_valid_sims_bic_distance},
and \ref{tab:gaussian_valid_sims_bic} summarize these comparisons.
The third column in each table reports the number of simulations in which
the SOTA algorithm produced a valid PAG.

Across nearly all scenarios, from small to large sample sizes -- except for
comparisons with BCCD and MAGSL at 1,000 samples -- dcFCI significantly
outperforms the SOTA algorithms, inferring PAGs with substantially higher
compatibility with the data. Additionally, dcFCI consistently infers PAGs with
BIC values closer to those of the true PAGs, although these values are not
always the smallest. Notably, all score-based algorithms, except MAGSL,
infer PAGs with lower BIC values. This somewhat surprising result, previously
observed in the GPS work, raises concerns about the instability of BIC score
computation and its use in causal discovery \citep{claassen2022greedy}.

For completeness, Tables \ref{tab:gaussian_valid_sims_shd},
\ref{tab:gaussian_valid_sims_fdr}, and \ref{tab:gaussian_valid_sims_for} present
comparisons in terms of SHD, FDR, and FOR, focusing exclusively on simulations
where the SOTA algorithms produced valid PAGs. The results are consistent
with those obtained using all simulations, further demonstrating
dcFCI's robustness, which extends beyond ensuring PAG validity.

\Cref{tab:gaussian_all_sims_time_taken} provides the runtime
comparison showing that dcFCI is significantly faster
than both MAGSL and DCD.

\subsection{Using Datasets with Mixed Data Types}
\label{sec:mixed_sims}

A key advantage of dcFCI lies in its versatility, as the proposed
MEC-targeted data-PAG compatibility score is nonparametric and, therefore, not
constrained by specific distributional assumptions. This flexibility allows dcFCI
to be applied to any dataset, provided there is a method to compute the posterior
probabilities of conditional (in)dependence hypotheses. As discussed in Section
\ref{sec:lr-tests}, such probabilities can be easily derived from standard statistics,
such as the \(\chi^2\) statistic from likelihood-ratio tests -- a widely used approach
for assessing conditional independence among diverse types of variables.

In this section, we evaluate and compare dcFCI's performance in causal
discovery on datasets containing a mixture of 3 continuous, 1 binary, and 1 multinomial
variables. Unlike dcFCI, existing hybrid and score-based algorithms,
such as BCCD, DCD, MAGSL, and GPS, are limited to causal discovery in linear
Gaussian models, making them unsuitable for this comparison. Therefore, the
results presented here focus on a direct comparison of dcFCI with the FCI and cFCI algorithms.

We used the same 10 PAGs from the previous study.  We generated 30 distinct datasets
for each PAG and sample sizes of \( N = \{1000, 5000, 10000, 50000\} \),
using generalized additive models, adhering to the PAG structure.
This was done using the \texttt{simMixedDAG} R package \citep{lin2019simmixed},
with model coefficients randomly assigned, ensuring their absolute values exceeded 0.2.
This process yielded 300 potentially unfaithful mixed-data type datasets for each sample size.

The comparison of dcFCI with FCI and cFCI yielded results
consistent with those for Gaussian datasets, demonstrating dcFCI's
superior robustness, accuracy, data compatibility, and
adaptability across various data types and scenarios.

\begin{table}[h]
\centering
\caption{Percentage of valid PAGs from mixed-data type datasets.}
\begin{tabular}{c|c|c|c}
\hline
N & FCI & cFCI & dcFCI \\
\hline \hline
\multirow{1}{*}{1,000} & 97.67 & 52.00 & \bf 100.00 \\
\multirow{1}{*}{5,000} & 99.00 & 63.33 & \bf 100.00 \\
\multirow{1}{*}{10,000} & 97.67 & 69.33 & \bf 100.00 \\
\multirow{1}{*}{50,000} & 99.33 & 91.67 & \bf 100.00 \\
\hline \end{tabular}
\label{tab:mixed_valid_PAGs}
\end{table}

\begin{table}[h]
\centering
\caption{True PAG recovery rate in mixed-data type simulations.}
\begin{tabular}{c|c|c|c|c}
  \hline
N & FCI & cFCI & dcFCI ($k=1$) & dcFCI ($k=2$) \\
  \hline
  \hline
\multirow{1}{*}{1,000} & 5.67 & 0.33 & \bf 15.67 / 18 / 19.67 & \bf 17 / 19.67 / 35.67 \\
  \multirow{1}{*}{5,000} & 39.67 & 23.67 &\bf  55 / 56 / 56.33 & \bf 56.33 / 60.67 / 68 \\
  \multirow{1}{*}{10,000} & 58.33 & 43.33 & \bf 72.67 / 73.67 / - & \bf 73.67 / 75.33 / 79 \\
  \multirow{1}{*}{50,000} & 87.33 & 80 & \bf 94.67 / 95 / - & \bf 94.67 / 95 / 95.33 \\
   \hline
\end{tabular}
\label{tab:mixed_true_pag_rec_rate}
\end{table}

As shown in \Cref{tab:mixed_valid_PAGs}, dcFCI reliably recovers only valid PAGs,
whereas FCI and particularly cFCI produce invalid PAGs.
Moreover, as demonstrated in \Cref{tab:mixed_true_pag_rec_rate},
dcFCI achieves a substantially higher recovery rate of the true PAG, even with smaller
sample sizes.
Additionally Tables \ref{tab:mixed_all_sims_shd}, \ref{tab:mixed_all_sims_fdr} and
and \ref{tab:mixed_all_sims_for} (covering all simulations) and Tables
\ref{tab:mixed_valid_sims_shd}, \ref{tab:mixed_valid_sims_fdr}, and
\ref{tab:mixed_valid_sims_for} (focusing on simulations that yielded only valid PAGs),
show that dcFCI infers PAGs with significally lower (better) SHD, FDR, and FOR
values compared to both FCI and cFCI.
Finally, \Cref{tab:mixed_valid_sims_pag_score} shows that dcFCI consistently
generates PAGs with significantly higher overall data-PAG compatibility scores
than those produced by FCI and cFCI, further emphasizing its robustness and
superior ability to infer PAGs that better align with the data.

\begin{table}[h]
\centering
\setlength{\tabcolsep}{3pt}
\caption{Comparison in terms of Structural Hamming Distance (SHD) in simulations with mixed data types.
Lower SHD indicates more accurate edges.
Green downward arrows indicate significant improvement of dcFCI over SOTA algorithms.}
\begin{tabular}{l|c|c|c|c|c|c|c|c|c}
\hline
\multirow{2}{*}{Algorithm} & \multirow{2}{*}{N} & \multicolumn{6}{c|}{SHD}  & \multicolumn{2}{c}{Comparison}  \\
& & Min & Q1 & Med & Mean & Q3 & Max & p-value & Diff\\
\hline
\hline
FCI & \multirow{3}{*}{$1,000$} &   0 &   3 &   5 & 5.3 &   7 &  13 & 1e-05 & $\color{PineGreen}{\boldsymbol\downarrow}$ \\
  cFCI &  &   0 &   5 &   7 & 7.1 & 9.2 &  13 & 2.7e-24 & $\color{PineGreen}{\boldsymbol\downarrow}$ \\
  dcFCI &  &   0 &   2 &   4 & 4.4 &   6 &  15 &   &  \\
\hline
  FCI & \multirow{3}{*}{$5,000$} &   0 &   0 &   2 & 2.8 &   4 &  12 & 0.012 & $\color{PineGreen}{\boldsymbol\downarrow}$ \\
  cFCI &  &   0 &   1 &   4 &   4 &   7 &  13 & 2.3e-12 & $\color{PineGreen}{\boldsymbol\downarrow}$ \\
  dcFCI &  &   0 &   0 &   0 & 2.2 &   4 &  14 &   &  \\
\hline
  FCI & \multirow{3}{*}{$10,000$} &   0 &   0 &   0 & 1.9 &   3 &  12 & 0.00085 & $\color{PineGreen}{\boldsymbol\downarrow}$ \\
  cFCI &  &   0 &   0 &   2 & 2.6 &   4 &  13 & 1.5e-14 & $\color{PineGreen}{\boldsymbol\downarrow}$ \\
  dcFCI &  &   0 &   0 &   0 & 1.2 &   2 &  11 &   &  \\
\hline
  FCI & \multirow{3}{*}{$50,000$} &   0 &   0 &   0 & 0.59 &   0 &  11 & 0.006 & $\color{PineGreen}{\boldsymbol\downarrow}$ \\
  cFCI &  &   0 &   0 &   0 & 0.79 &   0 &  11 & 8.1e-08 & $\color{PineGreen}{\boldsymbol\downarrow}$ \\
  dcFCI &  &   0 &   0 &   0 & 0.23 &   0 &  11 &   &  \\
 \hline
\end{tabular}
\label{tab:mixed_all_sims_shd}
\end{table}

\section{Real-World Application}

We applied dcFCI to the
Diabetes Health Indicators Dataset (DHID)
from Kaggle, a curated subset of the
2015 Behavioral Risk Factor Surveillance System (BRFSS), which
includes responses from 253,680 individuals, focusing on lifestyle
factors and health indicators related to chronic conditions and
acute events (see \Cref{sec:appe_application}).
We used 16 mixed-type variables, with
Body Mass Index (BMI)
as the only continuous variable,
and the others categorical (binary or multinomial).
To satisfy the conditional Gaussianity assumption for certain
independence tests, we transformed BMI using a rank-based inverse normal
transformation from the \texttt{RNOmni} R package \citep{mccaw2023rnomni}.

The simulation results in Section \ref{sec:simulations}
suggest that 50,000 samples are sufficient to reliably
infer five-variable PAGs. Based on this, we
began by applying dcFCI, FCI, and cFCI algorithms to all possible
five-variable subsets and evaluating the consistency of the inferred
relationships across PAGs.

\begin{figure}[h]
    \begin{minipage}[b]{0.48\linewidth}
      \subfloat[]{
        \includegraphics[width=\linewidth]{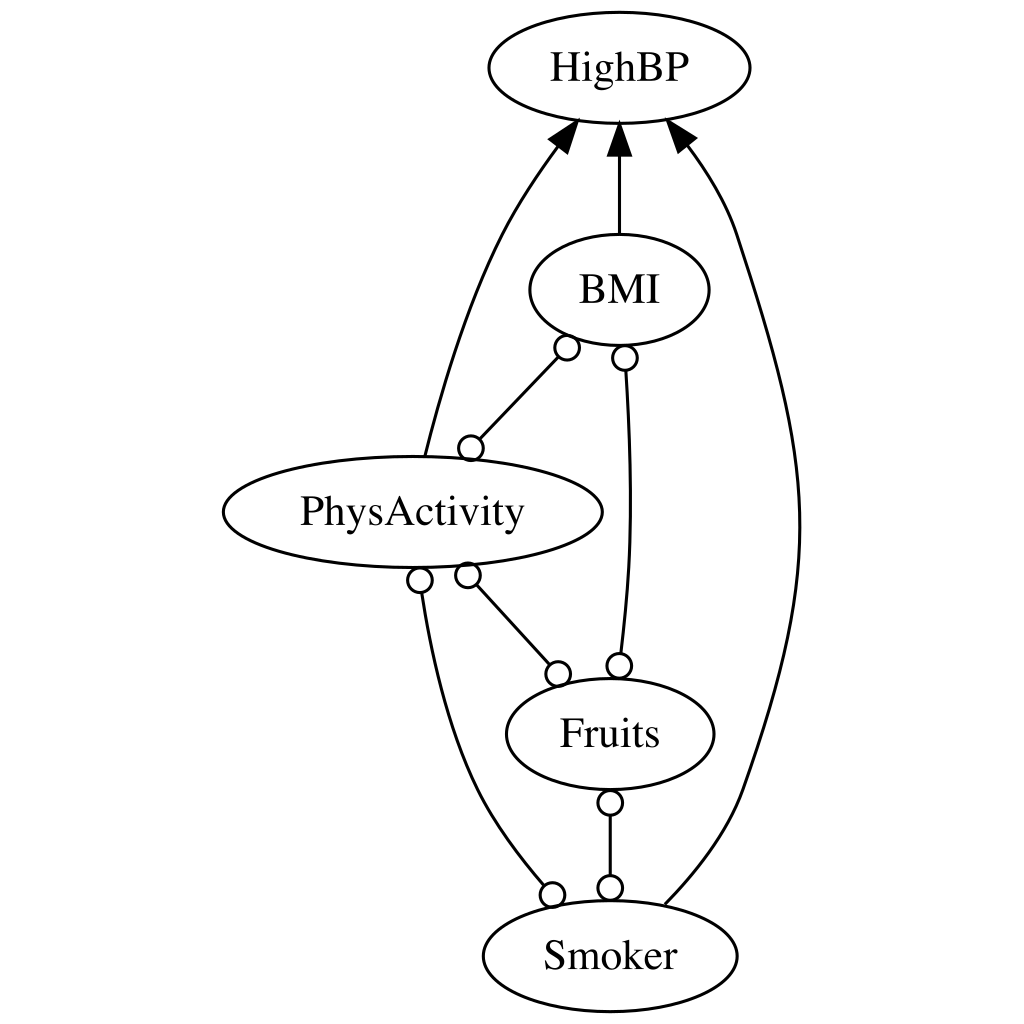}
    }
    \end{minipage}
    \begin{minipage}[b]{0.48\linewidth}
    \subfloat[]{
        \includegraphics[width=\linewidth]{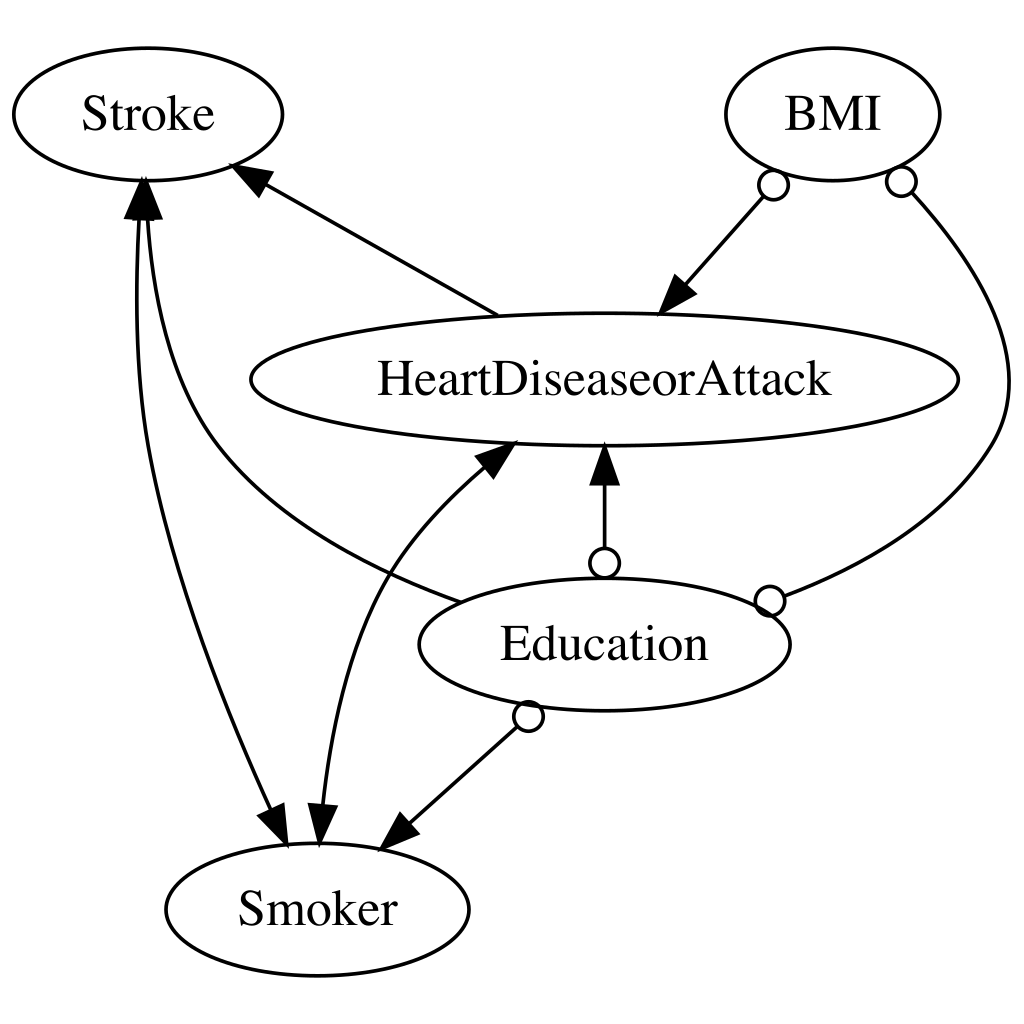}
    }
    \end{minipage}
    \caption{dcFCI, FCI, and cFCI agree on the inferred PAGs (a) and (b).
    In (a), physical activity (PhysActivity), BMI, and
    smoking status (Smoker) are causal contributors to HighBP. In (b),
    heart disease or myocardial infarction (HeartDiseaseorAttack) and
    education causal contributors to stroke. }
    \label{fig:dhid_agreed_PAGs}
\end{figure}

The PAGs in \Cref{fig:dhid_agreed_PAGs} were inferred by all three algorithms.
PAG (a) suggests with moderate confidence (score bounds of $[0.252, 0.484]$)
that physical activity (PhysActivity), BMI, and smoking status (Smoker)
causally contribute to high blood pressure (HighBP).
PAG (b), with lower confidence (score bounds of $[0, 0.0463]$),
indicates that heart disease or myocardial infarction (HeartDiseaseorAttack) and
education causally influence the likelihood of having a stroke.


The orientations inferred by dcFCI show strong consistency across
all 5-variable PAGs. Additionally, dcFCI identifies more causes of HighBP,
including income, sex, and healthcare coverage (AnyHealthcare)
(Figures \ref{fig:highBP_extra}a,d,g), as well as additional causes of stroke,
such as income, healthcare coverage, physical activity, high cholesterol, and
heavy alcohol consumption (HvyAlcoholConsump) (Figures \ref{fig:stroke_extra}a,d,g).
In contrast, FCI produced several inconsistencies. For instance, smoking status and
physical activity, inferred as causes of HighBP in \Cref{fig:dhid_agreed_PAGs}a,
appear as definite non-causes in other
PAGs (Figures \ref{fig:highBP_extra}e,h).
Similarly, HeartDiseaseorAttack, inferred as cause of stroke in \Cref{fig:dhid_agreed_PAGs}b,
is identified as a non-cause in \Cref{fig:stroke_extra}h.
Additionally, income appears both as a cause (\Cref{fig:stroke_extra}b) and as
a non-cause of stroke (\Cref{fig:stroke_extra}e,h).
While cFCI did not contradict itself in these relationships, it was often overly
conservative, leaving many edges unoriented and occasionally producing invalid
PAGs (Figures \ref{fig:highBP_extra}c,f,i and \ref{fig:stroke_extra}c,f,i).

When applying dcFCI to generate PAGs over six variables,
reliability dropped significantly. The resulting PAGs had very low scores,
and some orientations observed in the 5-variable PAGs were contradicted,
violating what we define as marginal consistency \cite{roumpelaki2016marginal}
(see \Cref{fig:dcfci_inconsistencies}).
This highlights a key scalability challenge in causal discovery: the primary
limitation is not computational complexity but insufficient sample size.
In complex real-world applications, particularly in medicine, collecting
datasets large enough to support reliable causal inference is often impractical.
The compatibility score thus serves as a valuable indicator of the maximum
number of variables that can be reliably analyzed, helping to balance the scope
of causal discovery with the statistical power of the data.

\section{Conclusion, Limitations, and Future Work}

We introduced a nonparametric score to assess data-PAG compatibility,
based on the skeleton and colliders with order, which jointly define the MEC.
We then developed dcFCI, a hybrid causal discovery algorithm that integrates this
score into a greedy (Anytime-)FCI-guided search.
It recovers PAGs that best reflect the data
while seamlessly resolving conflicts
that often arise in empirically unfaithful settings.
Additionally, it ensures that inferred PAGs
are valid and reliable for downstream causal analyses,
such as effect identification \cite{jaber2022causal}.
Crucially, dcFCI supports likelihood-ratio conditional independence tests,
which are especially effective for heterogeneous real-world datasets (e.g.
electronic health records) with mixed variable types.
Despite its greedy nature, dcFCI with $k=1$
matches the performance of MAGSL,
which guarantees global optimality but is restricted to Gaussian data.
Moreover, dcFCI consistently outperforms SOTA algorithms across all evaluated
scenarios, including small sample sizes, achieving superior results in SHD, FDR,
FOR, true PAG recovery, MEC validity, and data compatibility, while
supporting mixed variable types.

Despite dcFCI's substantial improvements in robustness, accuracy, and flexibility,
its greedy search may overlook PAGs that, with
additional iterations, could evolve into more compatible structures.
For practical use, we recommend users
assess the stability of the output by varying \( k \). However, it is important
to note that larger values of \( k \) may diminish the score's discriminative
power, as they require evaluating a greater number of conditional independence hypotheses.
Future work should explore strategies for selecting the optimal \( k \),
balancing computational efficiency with accuracy.

Remarkably, dcFCI also demonstrated significantly faster performance than
existing search-based algorithms, such as DCD and MAGSL. While this highlights
its potential scalability, the primary challenge in causal discovery remains
capturing statistical uncertainty rather than computational complexity alone.
Future directions also involve extending dcFCI to start with a carefully
selected subset of variables and progressively expand the PAG, continuing
the process as long as the score indicates sufficient reliability.
Additionally, the Bayesian score could be further leveraged to incorporate prior
knowledge, by either adjusting (in)dependency priors or updating scores
based on known (non-)ancestral relations \cite{ribeiro2024anchorfci, silva2024human}.
These advancements could further improve the robustness and applicability
of dcFCI in real-world causal discovery tasks.

\bibliography{references}
\bibliographystyle{IEEEtran}

\vspace{-0.25em}
\section{Biography Section}
\vspace{-3.2em}
\begin{IEEEbiographynophoto}{Adèle Ribeiro}
is a researcher at the Institute of Medical Informatics,
University of Münster, since 2024.
Previously, she worked at the University of Marburg (2022-2024),
Columbia University (2019-2022), and the Heart Institute (InCor) at
University of São Paulo (USP).
She holds a Ph.D., M.Sc., and B.Sc. from
the Institute of Mathematics and Statistics at USP.
Her research focuses on enhancing AI and Machine Learning through
causal reasoning.
\end{IEEEbiographynophoto}
\vspace{-3em}
\begin{IEEEbiographynophoto}{Dominik Heider}
is the director of the Institute of Medical Informatics,
University of Münster, since 2024.
He was a professor at University of Düsseldorf (2023-2024), Marburg (2016-2023),
and TUM Campus Straubing (2014-2016), and has been a visiting professor at Harvard University
since 2023. He holds a diploma in computer science (2006) and a PhD (2008) from
Münster and a habilitation (2012) from the University of Duisburg-Essen.
His research focuses on developing
AI and Machine Learning tools to address critical biomedical challenges.
\end{IEEEbiographynophoto}

\clearpage
\input{appendix}

\end{document}

%% file: appendix.tex
{\appendices

\section{Basic Concepts and Definitions}
\label{ape:further_definitions}
This work builds on established causality literature, utilizing conceptual
frameworks and tools under the assumption that the underlying SCM
is free from non-recursive mechanisms, ensuring that the
corresponding causal diagram is acyclic \citep{pearl2000causality}.

\vspace{0.3em}
\noindent\textbf{Structural Causal Model (SCM):  \citep{pearl2000causality}}
An SCM $\mathcal{M}$ is
a 4-tuple $\langle \*U, \*V, \mathcal{F}, P(\*U)\rangle$, where $\*{U}$
represents the set of exogenous (unobserved) variables, and $\*{V}$ denotes
the set of endogenous (observed) variables.
The collection $\mathcal{F} = \{f_i\}_{i=1}^{|\*{V}|}$ consists of functions
where each function $f_i \in \mathcal{F}$ describes how each endogenous
variable $V_i \in \*{V}$ is determined by its direct causes, which include
both its exogenous causes $\*{U}_i \subseteq \*{U}$ and its endogenous
causes (or parents) $Pa(V_i) \subseteq \*{V} \setminus \{V_i\}$.
Additionally, $P(\*{U})$ represents the probability distribution over
the exogenous variables.
\vspace{0.3em}

The following procedure shows that an SCM is uniquely represented by a
graphical causal model, commonly known as a causal diagram. In this model,
directed edges indicate causal relationships, while dashed bidirected edges
denote latent confounding, meaning the connected variables share an
unobserved common cause.

\vspace{0.3em}
\noindent\textbf{Causal Diagram:  \citep{pearl2000causality}}
Each SCM $\mathcal{M}$ induces a directed
acyclic graph (DAG) with bidirected edges
-- or an acyclic directed mixed graph (ADMG) -- denoted $G(\*V, \*E)$,
which is known as a causal diagram. This diagram captures the structural
relationships among the variables in $\*{V} \cup \*{U}$. In this representation,
each endogenous variable $V_i \in \*V$ is depicted as a vertex,
with directed edges $(V_j \rightarrow V_i)$ connecting each
parent $V_j \in Pa(V_i)$ to $V_i$. Additionally, dashed bidirected edges
$(V_j \dashleftarrow \!\!\!\!\!\!\!\!\! \dashrightarrow V_i)$ are drawn between
pairs of endogenous variables $V_i$ and $V_j$ whenever they share a common
exogenous cause, i.e., $\*U_i \cap \*U_j \neq \emptyset$.
\vspace{0.3em}

Importantly, \( G \) is not merely a depiction of causal directions among \( \*V \).
Rather, it is a formal construct -- a Causal Bayesian Network (CBN) -- that precisely
encodes not only the (in)dependence constraints among \( \*V \), thereby accurately
representing \( P(\*V) \), but also the full set of interventional distributions
\( P(\*V | do(\*X = \*x)) \) for every \( \*X \subset \*V \) induced by
\( \mathcal{M} \) \cite{pearl2000causality}.
This property is essential to any causal analysis.

\vspace{0.3em}
\noindent\textbf{Maximal Ancestral Graph (MAG): \citep{richardson2002ancestral}}
A MAG is a mixed graph -- containing directed ($\rightarrow$) and bidirected
($\leftrightarrow$) edges -- that represents a set of causal diagrams with the
same set of observed variables, capturing identical conditional independence and
ancestral relationships. It is both \emph{ancestral} and \emph{maximal}.
A node $A$ is said to be an ancestor of another node $B$ if there is a
directed path from $A$ to $B$. In contrast, $A$ is said to be a \textit{spouse}
of $B$ if $A\leftrightarrow B$ is present.
A graph is \emph{ancestral} if it does not contain directed or almost directed cycles.
In this context, a directed cycle occurs when a node is an ancestor of itself
through directed paths, while an almost directed cycle occurs when a node is
both a spouse and an ancestor of another node.
A graph is \emph{maximal} if no inducing path exists between any two
non-adjacent nodes. An inducing path is a path where every non-endpoint node
is a collider and an ancestor of one of the path's endpoints.

\vspace{0.3em}
\noindent\textbf{Partial Ancestral Graph (PAG): \cite{zhang2008fci}}
A PAG is a mixed graph that includes directed (\(\rightarrow\)),
bidirected (\(\leftrightarrow\)), partially directed (\(\circ\!\!\rightarrow\)),
undirected (\(-\)), and unspecified (\(\circ\!\!-\!\!\circ\)) edges.
It represents an equivalence class of MAGs with the same observed variables,
known as a Markov Equivalence Class (MEC). All models within a MEC share the
same set of conditional (in)dependence relations.
A PAG has the same adjacencies as every MAG in the class and displays all and
only those edge marks that are shared across all models in the MEC.
Tails and arrowheads indicate ancestral (causal) and non-ancestral (non-causal)
relationships, respectively, common to all models within the MEC.
A circle (\(\circ\)) edge mark indicates a non-invariant relationship,
meaning there is at least one model in the MEC where the edge mark is a tail
and another where it is an arrowhead.

\vspace{0.3em}
\noindent \textbf{M-separation in MAGs \cite{zhang2008causal}:}
M-separation extends d-separation to MAGs, where d-separation in a causal
diagram corresponds to m-separation in its unique associated MAG over the observed variables,
and vice versa. In a MAG, a path \( p \) between \( X \) and \( Y \) is active
(or m-connecting) relative to \( \mathbf{Z} \) (with \( X, Y \notin \mathbf{Z} \)) if:
(1) Every non-collider on \( p \) is not in \( \mathbf{Z} \); and
(2) Every collider on \( p \) is an ancestor of some \( Z \in \mathbf{Z} \).
Two disjoint sets of variables $\*X$ and $\*Y$ are \emph{m-separated} by \( \mathbf{Z} \)
if no m-connecting path exists between any variable \( X \in \*Z \) and \( Y \in \*Y\)
relative to \( \mathbf{Z} \).

\vspace{0.3em}
\noindent \textbf{M-separation in PAGs \cite{jaber2022causal}:}
In a PAG, let $\langle A,B,C\rangle$ be a triple along a path $p$.
The node $B$ is a \emph{collider} on $p$ if both edges are into $B$
(i.e., $A *\!\!\rightarrow B\leftarrow\!\!* C$). Conversely,
$B$ is a \emph{definite non-collider} on $p$ if one of the edges is directed away from $B$
(i.e., $A\leftarrow B *\!\!\--\!\!*C$ or $A *\!\!\--\!\!*B\rightarrow C$),
or both edges have circle marks at $B$ and there is no edge between $A$ and $C$
(i.e., $A\circ\!\!\--\!\!\circ B\circ\!\!\--\!\!\circ C$ , where $A$ and $B$ are not adjacent).
A path is said to have \textit{definite status} if every non-endpoint node along it
is either a collider or a definite non-collider.
Further, a path $p$ between $X$ and $Y$ is a definite m-connecting path relative
to a (possibly empty) set $\mathbf{Z}$ ($X,Y\not\in\mathbf{Z}$) if (1) $p$ has definite status;
(2) every definite non-collider on $p$ is not in $\mathbf{Z}$; and (3) every collider on $p$
is an ancestor of some $Z \in \mathbf{Z}$.
Two disjoint sets of variables $\*X$ and $\*Y$ are \emph{m-separated} by \( \mathbf{Z} \)
if no definite m-connecting path exists between any variable \( X \in \*Z \) and
\( Y \in \*Y\) relative to $\mathbf{Z}$.

\section{Related Work}
\label{sec:related_work}

Over the years, numerous causal discovery algorithms have been proposed,
leveraging conditional independence constraints, goodness-of-fit scores, or
a combination of both \citep{glymour2019review}. However, most of these methods
are tailored to simplified, potentially unrealistic scenarios where causal
sufficiency -- the inclusion of all confounder variables in the observed
set -- or specific parametric and distributional constraints are imposed.
In more complex real-world scenarios involving latent confounding and mixed
types of variables, causal discovery becomes significantly more challenging.

\paragraph{Fast Causal Inference (FCI) Algorithm} The seminal FCI algorithm,
first introduced by \cite{spirtes2001causation} and later refined with a
complete set of orientation rules by \cite{zhang2008fci}, remains a cornerstone
in this area due to its solid theoretical foundations, relatively few assumptions
compared to alternative approaches, and flexibility in accepting any appropriate
statistical test for conditional independencies. Remarkably, it is
nonparametric (i.e., it does not require any functional or distributional
assumptions beyond those of the conditional independence tests), and it is both
sound and complete, even in the presence of latent confounding and selection
bias. This makes it particularly well-suited for analyzing real-world datasets.
However, in data-limited scenarios, where the empirical faithful assumption is likely to
be violated and statistical tests for conditional independence are prone to
errors, it often produces highly unreliable results,  particularly due to the
sensitivity to the order of independence tests and the propagation of
errors \citep{colombo2012learning}.

\paragraph{Extensions of FCI and Hybrid Approaches} Due to the FCI's lack of
robustness, several algorithms have been developed to improve the reliability of
causal discovery under latent confounding. Some approaches focus on mitigating
the impact of potentially unfaithful conditional independence relations.
Examples include Anytime FCI \citep{spirtes01anytime}, Really Fast Causal
Inference (RFCI) \citep{colombo2012learning}, and Conservative FCI (cFCI)
\citep{colombo2012learning, colombo2014order}.
Other methods leverage a logical characterization
of constraint-based causal discovery \citep{claassen2011logical}, translating
conditional independence constraints into logical statements about ancestral
relationships, and use Answer Set Programming (ASP) solvers to identify optimal
causal structures \citep{claassen2012bccd, hyttinen2014constraint, magliacane2016ancestral}.
Hybrid and score-based approaches offer greater robustness by integrating a
scoring function that (approximately)
measures the compatibility between a candidate model and the observed data.
Notable early hybrid approaches include Greedy FCI (GFCI) \citep{ogarrio2016hybrid}
and Bayesian Constraint-Based Causal Discovery (BCCD) \citep{claassen2012bccd},
though both use scoring functions that do not account for latent confounding.
GFCI first applies the Fast Greedy Search (FGS) algorithm
\citep{ramsey2015scaling} -- a score-based search algorithm designed for causally
sufficient settings -- to obtain an approximate structure and then applies FCI
to adjust for potential confounding, modifying the structure accordingly.
 In contrast, BCCD integrates a logical version of cFCI with Bayesian reliability
 scores for both pairwise conditional independence tests and logical causal
 statements, guiding both the adjacency and orientation phases.

\paragraph{Score-Based Methods} More recently, score-based approaches have been
developed to explicitly account for latent confounding by searching the space of MAGs.
As shown by \cite{richardson2002ancestral, tsirlis2018scoring},
the Bayesian Information Criterion (BIC) asymptotically approximates the posterior
probability of MAGs under a linear Gaussian parameterization. Importantly, BIC
is score-equivalent, meaning that Markov-equivalent MAGs receive the same
asymptotic score. Additionally, the maximum likelihood estimates (MLEs)
required for its computation can be efficiently obtained using the Residual
Iterative Conditional Fitting (RICF) algorithm \citep{drton2009ricf}.
Due to its theoretical properties and computational efficiency, the Gaussian
BIC has become widely adopted in score-based causal discovery. Several algorithms
have been proposed to identify a MAG that represents the optimal MEC and its
corresponding PAG according to this score. These include Greedy Search for
MAGs (GSMAG), MAG Max–Min Hill-Climbing (M3HC) \citep{tsirlis2018scoring},
MAG Structure Learning (MAGSL) \citep{rantanen2021maximal}, and Differentiable
Causal Discovery (DCD) \citep{bhattacharya2021differentiable}. Notably,
MAGSL is an exact search algorithm that guarantees optimality under the
Gaussian BIC, rather than a greedy search method. Furthermore, DCD defines
the space of ancestral graphs using differentiable algebraic constraints and
employs gradient-based optimization to identify the optimal structure, instead
of relying on traditional search approaches. Finally, Greedy PAG
Search (GPS) \citep{claassen2022greedy} proposes a greedy search
for the optimal PAG based on the Gaussian BIC score while distinguishing itself
by operating directly in the space of PAGs. These advancements underscore the
growing role of score-based and hybrid approaches in causal discovery,
particularly in settings with latent confounding.

\section{On the Characterization of the MECs of MAGs}
\label{ape:markov_equivalence}

To enhance practical applicability, \cite{zhang2007characterization}
presents a graphical characterization of MECs that depends exclusively on
adjacencies and specific triples and paths within the graph. Specifically, he
shows that two MAGs
belong to the same MEC if and only if they share the same adjacencies,
unshielded colliders, and colliders identified by discriminating paths.
This characterization was later refined by \cite{ali2009mec},
utilizing the concept of triples with order.
In their definition \cite[Definition 3.11]{ali2009mec},
triples of order 0 are equivalent to unshielded triples,
while triples of higher orders are defined by discriminating paths,
with the order being roughly related to the length of these paths.
They demonstrated that two MAGs,
$G_1$ and $G_2$, are Markov equivalent if and only if they share the same
adjacencies (skeleton) and the same colliders with order \cite[Theorem 3.7]{ali2009mec}.
Moreover, they established that for any two Markov equivalent MAGs
$G_1$ and $G_2$, a collider (or non-collider) of a specific order in
$G_1$ is also a collider (or non-collider) of the same order in $G_2$
\cite[Proposition 3.12]{ali2009mec}.

Despite the significant simplification offered by this characterization,
algorithmically detecting triples of order higher than 0 in a graph is not
straightforward. To address this, \cite{claassen2022greedy} recently proposed an
even simpler characterization of the MEC of a MAG that does not depend on identifying
discriminating paths. Their characterization is based on the following recursive
definition of triples with order:

\begin{definition}[Triple with Order as in \cite{claassen2022greedy}]
Let $\mathfrak{C}_i$ resp. $\mathfrak{D}_i$ ($i \geq 0$) be the set of collider
resp. non-collider triples with order $i$ in a MAG $G$, defined recursively as follows:
\begin{itemize}
    \item A triple $\langle a, b, c \rangle \in \mathfrak{C}_0$ (resp. $\mathfrak{D}_0$),
    if $a \ast\!\!{-}\!\!\ast b \ast\!\!{-}\!\!\ast c$ is an unshielded collider
    (resp. non-collider) in $G$.
    \item A triple $\langle a, b, c \rangle \in \mathfrak{C}_i$
    (resp. $\mathfrak{D}_i$), with $i \geq 1$, if
    $\langle a, b, c \rangle \notin \mathfrak{C}_{j < i}$ (resp. $\mathfrak{D}_{j < i}$), and
    \begin{enumerate}
        \item $a \ast\!\!{-}\!\!\ast b \ast\!\!{-}\!\!\ast c$ is a collider
        (resp. non-collider) in $G$,
        \item $\exists q : \langle q, a, b \rangle \in \mathfrak{C}_{j < i}$,
        and $\langle q, a, c \rangle \in \mathfrak{D}_{k < i}$ .
    \end{enumerate}
\end{itemize}
\end{definition}
They specifically demonstrate that the MEC of a MAG $G$ is fully characterized by
the triple $\langle \mathfrak{S}, \mathfrak{C}, \mathfrak{D} \rangle$,
where $\mathfrak{S}$ is the (undirected) skeleton of $G$, and $\mathfrak{C}$ and
$\mathfrak{D}$ are the corresponding lists of colliders and non-colliders with order,
respectively, as defined above -- see \cite[Corollary 3]{claassen2022greedy}.
\cite{claassen2022greedy} also introduced the MAG-to-MEC algorithm, which
efficiently identifies all collider and non-collider triples with order in a
given MAG. The output of the MAG-to-MEC algorithm is the skeleton, besides the
lists $\mathfrak{C}$ and $\mathfrak{D}$ of the MAG. This algorithm enables the
determination of Markov equivalence between MAGs with polynomial time complexity
$\mathcal{O}(pe^4)$, where $p$ is the number of vertices and $e$ is the
number of edges in the graph.

\section{On Likelihood-Ratio Tests for Conditional Independence Among Variables of Diverse Types}
\label{ape:likelihood_ratio_tests}

As shown by \cite{tsagris2018constraint},
conditional independence among variables
of diverse types, including continuous, binary, nominal, and others, can be
tested by fitting two nested models.

Specifically, to test the conditional independence of
\(X\) and \(Y\), given a (possibly empty) set \(\mathbf{Z}\), the procedure
involves constructing: (1) a full model (\(M_1\)) for \(Y\) that depends on both
\(X\) and \(\mathbf{Z}\), and (2) a reduced model (\(M_0\)), nested within \(M_1\),
that depends only on \(\mathbf{Z}\). Model \(M_0\) is considered nested within
\(M_1\) if \(M_1\) can be reduced to \(M_0\) by setting the contribution of \(X\)
to zero. A likelihood-ratio test is then used to compare the goodness-of-fit of
the two models. If the models fit the data equally well, \(X\) and \(Y\) are
deemed conditionally independent given \(\mathbf{Z}\), assuming the models are
correctly specified. In practical terms, this indicates that \(X\) provides no
additional predictive information for \(Y\) once \(\mathbf{Z}\) has been accounted for.

The test statistic for a likelihood-ratio test follows a \(\chi^2\)-distribution
with degrees of freedom equal to \(|\Theta_1| - |\Theta_0|\), where \(|\Theta_1|\)
and \(|\Theta_0|\) are the number of parameters in \(M_1\) and \(M_0\), respectively.
While the test is asymptotically symmetric (yielding the same p-value when flipping
\(X\) and \(Y\)), this symmetry may not hold in finite samples. To address this,
\cite{tsagris2018constraint} proposed a symmetric conditional independence test
that combines the p-values from both directions (\(p_1\) and \(p_2\)) using the formula:
$p = \min\{2 \min(p_1, p_2), \max(p_1, p_2)\}.$

\section{On Bayes Factor Functions}
\label{ape:bff}

As detailed in \cite{johnson2023bayes}, standard test statistics such as
\(z\), \(t\), \(\chi^2\), and \(F\) follow well-defined distributions under
the null hypothesis. Under alternative hypotheses, their asymptotic distributions
depend solely on scalar-valued noncentrality parameters. By leveraging prior
density families indexed by standardized effect sizes, \cite{johnson2023bayes}
define Bayes factors as functions of these effect sizes, referring to them
as Bayes Factor Functions (BFFs).
In particular, the authors leverage prior densities that are special cases of
nonlocal alternative prior (NAP) densities, such as normal-moment and gamma priors,
which vanish when the noncentrality parameter is zero.
This property accelerates evidence accumulation for both true null and alternative
hypotheses while enabling closed-form Bayes factor expressions,
reducing computational complexity and improving practical usability.

The authors in \cite{johnson2023bayes} also propose that, in the absence of
prior knowledge about effect size,
one can assess how the evidence for a hypothesis varies across a range of
plausible effect sizes. For instance, plotting \( BF_{10} \) against
standardized effect size allows for selecting the maximum Bayes factor within a
meaningful range (e.g., effect sizes exceeding 0.1).
thereby identifying the strongest support for the alternative hypothesis,
derived from the specified family of prior densities on the noncentrality parameter,
while ensuring that the respective effect size is practically meaningful.
In causal discovery, particularly in conditional independence testing,
this approach prioritizes evidence for conditional dependence, making it
particularly valuable in low-sample settings where statistical power is limited,
while also reducing the risk of drawing conclusions from weak or unreliable
associations.

\section{Proofs of Propositions and Theorems}
\label{ape:proofs}

\noindent \textbf{Proof of \Cref{prop:minsep_noncollider}}

\begin{proof}[Proof of \Cref{prop:minsep_noncollider}]

Given that \(\*Z\) is a minimal separator of \(X\) and \(Y\), for each
\(Z_i \in \*Z\), there exists at least one m-connecting path between
\(X\) and \(Y\) given \(\*Z \setminus Z_i\), and such a path includes
\(Z_i\). Let \(\pi^{*}\) be the shortest among these paths.
Then, \(\pi^{*}\) is the closest m-connecting path to
\(\*Z \setminus Z_i\) (see \cite[Definition 3.23]{ali2009mec}) and,
by \cite[Proposition 3.24]{ali2009mec}, it is also a minimal
m-connecting path between \(X\) and \(Y\) given \(\*Z \setminus Z_i\)
(see \cite[Definition 3.18]{ali2009mec}).

By \cite[Theorem 2]{tian1998finding}, $Z_i$ in $\pi^{*}$ 
must be an ancestor of either $X$ or $Y$, since otherwise,
$\*Z \setminus Z_i$ would also be a
separator of $X$ and $Y$, contradicting the minimality of $\*Z$.
This ensures that $Z_i$ is part of a non-collider triple in $\pi^{*}$.

Let $\langle A, Z_i, C \rangle$ be the triple containing $Z_i$ in $\pi^{*}$:

\begin{itemize}
\item If $\langle A, Z_i, C \rangle$ is unshielded, then it is a
non-collider triple of order 0.
Notably, $(A,C)$, which can be equal to $(X,Y)$, is a pair of non-adjancent
nodes and, since $Z_i$ is a non-collider, any minimal separator
$\*Z' \in \operatorname{MinSep}_{\mathcal{P}^{(r)}}(A,C)$ must contain $Z_i$.

\item If $\langle A, Z_i, C \rangle$ is shielded, then, since
$\pi^{*}$ is a minimal m-connecting path, it follows from
\cite[Lemma 3.20]{ali2009mec} that $\langle A, Z_i, C \rangle$ lies on
a unique section of $\pi^{*}$ forming a discriminating path for $Z_i$.
Thus, $\langle A, Z_i, C \rangle$ is a non-colider with order $> 0$.
Notably, such a discriminating path takes the form
$\langle X', \ldots, A, Z_i, C \rangle$, where $(X',C)$ can be $(X,Y)$.
Thus, $(X',C)$ is a pair of non-adjancent nodes and, since $Z_i$ is a
non-collider, any minimal separator
$\*Z' \in \operatorname{MinSep}_{\mathcal{P}^{(r)}}(X',C)$ must contain $Z_i$.
\end{itemize}

Thus, every $Z_i \in \*Z$ is part of a non-collider triple of order
corresponding to some pair $(X',Y')$.
\end{proof}

\noindent \textbf{Proof of \Cref{prop:minimality_triples_order}}

\begin{proof}[Proof of \Cref{prop:minimality_triples_order}]

By definition (\cite[Definition 3.11]{ali2009mec}),
every triple with order $\langle A, B, C \rangle$ in an
$r$-PAG $\mathcal{P}^{(r)}$ corresponds
to a pair of non-adjacent nodes $(X,Y)$, where $X$ and $Y$ are either
the endpoints of an unshielded triple (with $X = A$ and $C = Y$),
or the endpoints of a discriminating path for $B$.

\vspace{1em}
\noindent \textbf{(If:)}
Let $\langle A, B, C \rangle$ be a triple
corresponding to (X,Y) in $\mathcal{P}^{(r)}$
and $\*Z \in \operatorname{MinSep}_{\mathcal{P}^{(r)}}(X,Y)$.
We show that, if $B \in \*Z$, then
$\langle A, B, C \rangle$ is a non-collider triple with order.
Otherwise, if $B \not \in \*Z$, then
$\langle A, B, C \rangle$ is a collider triple with order:

\vspace{0.5em}
\noindent $\langle A, B, C \rangle$ is an unshielded triple (with $X = A$ and $C = Y$)]:

Since the triple is unshielded, it has order 0.
  \begin{itemize}
  \item If $B \in \*Z$, then $\langle A, B, C \rangle$ must be a non-collider triple,
  otherwise, if $B$ were a collider, the triple would be active, forming an m-connecting path
  between $X$ and $Y$, contradicting the fact that $\*Z$ m-separates $X$ and $Y$.
  \item If $B \not \in \*Z$, then
  $\langle A, B, C \rangle$ must be a collider triple,
  otherwise, if $B$ were a non-collider, the triple would again be active,
  forming an m-connecting path between $X$ and $Y$,
  contradicting the m-separation of $X$ and $Y$ by $\*Z$.
  \end{itemize}

\vspace{0.5em}
\noindent $\langle A, B, C \rangle$ is part of a
discriminating path $\pi$ between $X$ and $Y$ for $B$:

Since $\*Z$ m-separates $X$ and $Y$, it follows from \cite[Lemma 3.19]{ali2009mec}
that every nonendpoint node in $\pi$, except possibly $B$,
is a collider in $\*Z$. Consequently,
the classification of the triple $\langle A, B, C \rangle$
depends entirely on whether $B$ belongs to $\*Z$:
  \begin{itemize}
  \item If $B \in \*Z$, then $\langle A, B, C \rangle$ must be
  a non-collider triple. Otherwise, if $B$ were a collider, the triple would be active,
  rendering the entire path $\pi$ $m$-connecting and contradicting the
  separation of $X$ and $Y$ by $\*Z$.
  \item If $B \not \in \*Z$, then  $\langle A, B, C \rangle$ must
  be a collider triple. Otherwise, if $B$ were non-collider, the triple would be active,
  once again making $\pi$ $m$-connecting and contradicting the fact
  that $\*Z$ is a separator of $X$ and $Y$.
  \end{itemize}

\vspace{1em}
\noindent \textbf{(Only If:)}
Let $\langle A, B, C \rangle$ be a triple with order
corresponding to (X,Y) in $\mathcal{P}^{(r)}$.
We show that, for any $\*Z \in \operatorname{MinSep}_{\mathcal{P}^{(r)}}(X,Y)$,
if $B$ is a non-collider in $\langle A, B, C \rangle$,
then $B \in \*Z$.
Conversely, if $B$ is a collider in $\langle A, B, C \rangle$,
then $B \not \in \*Z$.

\vspace{0.5em}
\noindent $\langle A, B, C \rangle$ is an unshielded triple (with $X = A$ and $C = Y$)]:
  \begin{itemize}
  \item If $B$ is a non-collider in $\langle A, B, C \rangle$, then,
  for any $\*Z \in \operatorname{MinSep}_{\mathcal{P}^{(r)}}(X,Y)$, we must have $B \in \*Z$.
  Otherwise, if $B \not \in \*Z$, the triple would be active, forming an m-connecting path
  between $X$ and $Y$, contradicting the fact that $\*Z$ separates $X$ and $Y$.
  \item If $B$ is a collider in $\langle A, B, C \rangle$, then
  for any $\*Z \in \operatorname{MinSep}_{\mathcal{P}^{(r)}}(X,Y)$, we must have $B \not \in \*Z$.
  Otherwise, if $B \in \*Z$, the triple would again be active,
  forming an m-connecting path between $X$ and $Y$,
  contradicting the m-separation of $X$ and $Y$ by $\*Z$.
  \end{itemize}

\vspace{0.5em}
\noindent $\langle A, B, C \rangle$ is part of a
discriminating path $\pi$ between $X$ and $Y$ for $B$:

It follows from \cite[Lemma 3.19]{ali2009mec}
that every nonendpoint node in $\pi$, except possibly $B$,
is a collider in any m-separator (minimal or not) of $X$ and $Y$.

Since $\*Z \in \operatorname{MinSep}_{\mathcal{P}^{(r)}}(X,Y)$,
whether $B$ belongs to $\*Z$ depends
entirely on the classification of the triple $\langle A, B, C \rangle$:
  \begin{itemize}
  \item If $B$ is a non-collider in $\langle A, B, C \rangle$, then,
  we must have $B \in \*Z$.
  Otherwise, if $B \not \in \*Z$, the triple would be active,
  rendering $\pi$ $m$-connecting and contradicting the fact
  that  $\*Z \in \operatorname{MinSep}_{\mathcal{P}^{(r)}}(X,Y)$.
  \item If $B$ is a collider in $\langle A, B, C \rangle$, then
  we must have $B \not \in \*Z$.
  Otherwise, if $B \in \*Z$, the triple would be active,
  once again making $\pi$ $m$-connecting and contradicting the fact
  that  $\*Z \in \operatorname{MinSep}_{\mathcal{P}^{(r)}}(X,Y)$
  \end{itemize}

This completes the proof.
\end{proof}

\noindent \textbf{Proof of \Cref{thm:complete_hypotheses_uncertainty}}

\begin{proof}[Proof of \Cref{thm:complete_hypotheses_uncertainty}]
Let $\mathcal{P}^{(r)} = (\*V, \*E^{(r)})$ be the $r$-PAG under evaluation.

\vspace{0.5em}
\noindent \textbf{(Sufficiency). }
We show that the (in)dependence relations in
$\*H_{\mathfrak{S}(\mathcal{P}^{(r)}), P} \cup
\*H_{\mathfrak{C}(\mathcal{P}^{(r)}), P}$ suffice to fully characterize
the uncertainty in $\mathcal{P}^{(r)}$. To establish this, we show that these relations
precisely characterize the uncertainty in the components that
uniquely define the MEC of $\mathcal{P}^{(r)}$:
the skeleton (adjacencies) and colliders with order
in $\mathcal{P}^{(r)}$ \citep[Theorem 3.7]{ali2009mec}.

Identifying the skeleton requires:
(a) Ensuring that pairs of adjacent nodes are dependent
conditioned on any subset $\*Z \subseteq \*V \setminus \{X,Y\}$ such that
$|\*Z| \leq r$; and
(b) Ensuring that pairs of non-adjacent nodes are independent
given their (minimal) separators in $\operatorname{MinSep}_{\mathcal{P}^{(r)}}$.

Thus, the uncertainty in identifying the skeleton arises from that in (a) and (b),
which correspond precisely to the first and second sets of (in)dependence hypotheses
in $\*H_{\mathfrak{S}(\mathcal{P}^{(r)}), P}$, as defined
in \Cref{def:skel_hypotheses}, Items \ref{itm:skel_a} and \ref{itm:skel_b}.
Note that the minimality of the separators in $\operatorname{MinSep}_{\mathcal{P}^{(r)}}$
is not directly relevant to identifying the skeleton
$\mathfrak{S}(\mathcal{P}^{(r)})$.
However, as shown in \Cref{prop:minimality_triples_order},
it fully characterizes all triples with order in $\mathcal{P}^{(r)}$, including
both non-collider and collider triples.

By \Cref{prop:minimality_triples_order}, any
non-collider triple with order
$\langle A, B, C \rangle$ in $\mathcal{P}^{(r)}$
corresponds to a pair of non-adjacent nodes $(X, Y)$ such that
$B \in \operatorname{MinSep}_{\mathcal{P}^{(r)}}(X, Y)$.
Thus, the third set of hypotheses
in $\*H_{\mathfrak{S}(\mathcal{P}^{(r)}), P}$, as defined
in \Cref{def:skel_hypotheses}, \Cref{itm:skel_c},
corresponds to dependencies associated with the activation of
all non-collider triples with order in $\mathcal{P}^{(r)}$.

These hypotheses not only suffice to characterize all non-collider triples
with order,
but also partially distinguish them from their collider triple
counterparts.
As a result, the process of quantifying their uncertainties
is inherently linked to quantifying the uncertainty associated
with colliders with order.

\Cref{prop:minimality_triples_order} also shows that, any collider triple with order
corresponds to a pair of non-adjacent nodes $(X, Y)$ such that
$B \not \in \operatorname{MinSep}_{\mathcal{P}^{(r)}}(X, Y)$.
The set $\*H_{\mathfrak{C}(\mathcal{P}^{(r)}), P}$
then precisely corresponds to dependencies associated with the activation of
$\mathfrak{C}(\mathcal{P}^{(r)})$, thus completing the characterization of the
uncertainty surrounding the identification of the
collider triples with order.

Thus, the uncertainty in identifying both the skeleton and the colliders with
order is fully captured through their directly implied (in)dependencies. Since
these components are sufficient for determining the MEC, they are also adequate
to both identify \( \mathcal{P}^{(r)} \) and distinguish it from any distinct \( r \)-PAG.

\vspace{0.5em}
\noindent \textbf{(Necessity). }
To prove necessity, we argue by contrapositive: if any
(in)dependence relation is omitted from
$\*H_{\mathfrak{S}(\mathcal{P}^{(r)}), P}$ or
$\*H_{\mathfrak{C}({\mathcal{P}^{(r)}}), P}$,
then two distinct $r$-MECs could appear indistinguishable in their
accounted hypotheses, even though they would differ when the relation is included.

We divide the proof based on the types of sets of relations:
\begin{enumerate}
\item Sets \ref{itm:skel_a} and \ref{itm:skel_b} of
$\*H_{\mathfrak{S}(\mathcal{P}^{(r)}), P}$, as in \Cref{def:skel_hypotheses},
allowing the distiction of PAGs differing in the presence or absence of an edge;
\item Set \ref{itm:skel_c} of $\*H_{\mathfrak{S}(\mathcal{P}^{(r)}), P}$,
as in \Cref{def:skel_hypotheses}, capturing dependencies arising from
active non-collider triples with order;
\item Set $\*H_{\mathfrak{C}(\mathcal{P}^{(r)}), P}$, capturing dependencies
arising from active collider triples with order.
\end{enumerate}

This categorization ensures that the necessity of hypotheses related to edge
existence and triple classification with order is systematically addressed.

\begin{enumerate}
\item \textbf{Relations in the sets \ref{itm:skel_a} and \ref{itm:skel_b} of
$\*H_{\mathfrak{S}(\mathcal{P}^{(r)}), P}$ (presence / absence of an edge).}
We establish the necessity of hypotheses in
the set \ref{itm:skel_a} of $\*H_{\mathfrak{S}(\mathcal{P}^{(r)}), P}$, which
captures dependencies due to existence of edges, paired with
hypotheses in the set \ref{itm:skel_b} of $\*H_{\mathfrak{S}(\mathcal{P}^{(r)}), P}$,
which accounts for independencies due to their absence.

We show that omitting a relation
corresponding to a dependence hypothesis
in set \ref{itm:skel_a}
(and the corresponding independence hypothesis in set \ref{itm:skel_b}),
may render two $r$-PAGs, differing by the presence of a single edge,
indistinguishable, despite representing distinct $r$-MECs.

Let $\{X, Y\}$ be an edge in $\*E^{(r)}$.
By contradiction, suppose that,
for some set $\*Z \subseteq \*V \setminus \{X, Y\}$,
the conditional independence relation
$(X,Y|\*Z)$ is not necessary for characterizing the uncertainty
in identifying the \( r \)-MEC represented by \( \mathcal{P}^{(r)} \).
For simplicity, take $|\*Z| = r$.

Such a relation would be considered in set \ref{itm:skel_a} of
$\*H_{\mathfrak{S}(\mathcal{P}^{(r)}), P}$.
However, under the assumption, consider a modified set of hypotheses:
$$\*H'_{\mathfrak{S}(\mathcal{P}^{(r)}), P}
= \*H_{\mathfrak{S}(\mathcal{P}^{(r)}), P} \setminus H_{(X \nindep Y | \*Z)_{P}}.$$

Now, consider a PAG $\mathcal{P}^{'(r)} = (\*V, \*E^{'(r)})$
differing from $\mathcal{P}^{(r)}$ only in the absence of the edge $\{X, Y\}$
-- i.e.,  $\*E'^{(r)} = \*E^{(r)} \setminus \{X, Y\}$ --,
with $\*Z$ being a minimal separator of $X$ and $Y$, i.e.,
$\*Z \in \operatorname{MinSep}_{\mathcal{P}^{'(r)}}(X,Y)$.

The relation $(X,Y|\*Z)$ would be considered in set \ref{itm:skel_b}
of $\*H_{\mathfrak{S}(\mathcal{P}^{'(r)}), P}$.
However, under the assumption that such a relation is not necessary,
consider the corresponding set of hypotheses
$$\*H'_{\mathfrak{S}(\mathcal{P}^{'(r)}), P}
= \*H_{\mathfrak{S}(\mathcal{P}^{'(r)}), P} \setminus H_{(X \indep Y | \*Z)_{P}}.$$

Since the two $r$-PAGs differ only by a single edge in their skeletons,
their corresponding modified set of hypotheses satisfy:
$$\*H'_{\mathfrak{S}(\mathcal{P}^{(r)}), P} = \*H'_{\mathfrak{S}(\mathcal{P}^{'(r)}), P}.$$

In cases where these $r$-PAGs share the same
colliders with order, trivially when no colliders are present, we have:
$$\*H'_{\mathfrak{S}(\mathcal{P}^{(r)}), P} \cup \*H'_{\mathfrak{C}(\mathcal{P}^{(r)}), P} =
\*H'_{\mathfrak{S}(\mathcal{P}^{'(r)}), P} \cup \*H'_{\mathfrak{C}(\mathcal{P}^{'(r)}), P}.$$
This equality implies that, although the two $r$-PAGs represent distinct MECs,
their uncertainties are assessed by the same set of hypotheses.

However, if the omitted relation were included and
$P( H_{(X \nindep Y | \*Z)_{P}} | \mathcal{D}) \neq
P( H_{(X \indep Y | \*Z)_{P}} | \mathcal{D})$,
the two $r$-PAGs could exhibit distinct uncertainties, allowing
the most plausible one to be identified.
This contradicts the assumption that a relation in
the set \ref{itm:skel_a} of
$\*H_{\mathfrak{S}(\mathcal{P}^{(r)}), P}$
is unnecessary, thereby establishing its necessity.

The proof of necessaty for the relations in
the set \ref{itm:skel_b} of
$\*H_{\mathfrak{S}(\mathcal{P}^{(r)}), P}$
is analogous, with the roles of the $r$-PAGs
$\mathcal{P}^{(r)}$ and $\mathcal{P}^{'(r)}$ being
swapped in the argument.

\item \textbf{Relations in the set \ref{itm:skel_c}
of $\*H_{\mathfrak{S}(\mathcal{P}^{(r)}), P}$ (active non-collider with order).}
Now, we establish the necessity of hypotheses in the set
\ref{itm:skel_c} of $\*H_{\mathfrak{S}(\mathcal{P}^{(r)}), P}$, with
each hypothesis, as shown in \Cref{prop:minsep_noncollider},
capturing a dependence arising from an active non-collider triplet with order.

We show that omitting a relation from the set \ref{itm:skel_c} makes
$\mathcal{P}^{(r)}$ to be equally distant from two distinct $r$-PAGs
$\mathcal{P}^{'(r)}$ and $\mathcal{P}^{''(r)}$,
contradicting the assumption that the hypothesis is irrelevant
for distinguishing PAGs that represent distinct MECs.

Let $\{X, Y\}$ be a pair of non-adjacent nodes in $\mathcal{P}^{(r)}$,
with $\*Z \in \operatorname{MinSep}_{\mathcal{P}^{(r)}}(X,Y)$ and $\*Z \neq \emptyset$.
By \Cref{prop:minsep_noncollider}, each $Z_i \in \*Z$ lies on a path $\pi$
between $X$ and $Y$ as a non-collider with order.
Thus, given $\*Z \setminus Z_i$, this non-collider is active, rendering
$X$ and $Y$ m-connecting.
By contradiction, suppose that, for some $Z_i \in \*Z$, the relation
$(X,Y|\*Z \setminus Z_i)$ is not necessary for characterizing the uncertainty
in identifying the \( r \)-MEC represented by \( \mathcal{P}^{(r)} \).
Such a relation would be considered in set (c) of
$\*H_{\mathfrak{S}(\mathcal{P}^{(r)}), P}$.
However, under the assumption, consider a modified set of hypotheses:
$$\*H'_{\mathfrak{S}(\mathcal{P}^{(r)}), P}
= \*H_{\mathfrak{S}(\mathcal{P}^{(r)}), P} \setminus
H_{(X \nindep Y | \*Z \setminus Z_i)_{P}}.$$

Now, consider a PAG $\mathcal{P}^{'(r)} = (\*V, \*E^{'(r)})$
that differs from $\mathcal{P}^{(r)}$ solely in that the set $\*Z' = \*Z \setminus Z_i$
is a minimal separator of $X$ and $Y$ in $\mathcal{P}^{'(r)}$ -- i.e.,
$\*Z' \setminus Z_i \in \operatorname{MinSep}_{\mathcal{P}^{'(r)}}(X,Y)$.
Structurally, PAG $\mathcal{P}^{'(r)}$ differs in that, the corresponding
path $\pi$ has now $Z_i$ as part of a collider triple
with order (see \Cref{prop:minimality_triples_order}).
The relation $(X,Y|\*Z \setminus Z_i)$ would be considered
in the set (b) of $\*H_{\mathfrak{S}(\mathcal{P}^{'(r)}), P}$.
However, under our assumption that such a relation is not necessary,
consider the corresponding set of hypotheses
$$\*H'_{\mathfrak{S}(\mathcal{P}^{'(r)}), P}
= \*H_{\mathfrak{S}(\mathcal{P}^{'(r)}), P} \setminus
H_{(X \indep Y | \*Z \setminus Z_i)_{P}}.$$

Then, consider a PAG $\mathcal{P}^{''(r)} = (\*V, \*E^{''(r)})$
differig from $\mathcal{P}^{(r)}$ solely from that
$X$ and $Y$ are adjacent, i.e., $\{X,Y\} \in \*E^{''(r)}$.
Structurally, PAG $\mathcal{P}^{''(r)}$ differs in that $Z_i$
is no longer part of a triple with order corresponding to $\{X,Y\}$.
The relation $(X,Y|\*Z \setminus Z_i)$ would be considered
in the set (a) of $\*H_{\mathfrak{S}(\mathcal{P}^{''(r)}), P}$.
However, under our assumption that such a relation is not necessary,
consider the corresponding set of hypotheses
$$\*H'_{\mathfrak{S}(\mathcal{P}^{''(r)}), P}
= \*H_{\mathfrak{S}(\mathcal{P}^{''(r)}), P} \setminus
H_{(X \nindep Y | \*Z \setminus Z_i)_{P}}.$$

An an example, let $\mathcal{P}^{(r)}$ consist of a
single unshielded triple
$X \circ\!\!-\!\!\circ Z_i \circ\!\!-\!\!\circ Y$, where
$\*Z = \{Z_i\}$ and $X \indep Y | \*Z$.
Further, let the PAG $\mathcal{P}^{'(r)}$ be
$X \circ\!\!\!\rightarrow Z_i \leftarrow\!\!\!\circ Y$, such that
$(X \indep Y)$.
Finally, let the PAG $\mathcal{P}^{''(r)}$ be
be a fully connected PAG between $X$, $Y$, and $Z_i$.

Under the assumption, the sets of modified hypotheses for \( \mathcal{P}^{'(r)} \)
and \( \mathcal{P}^{''(r)} \)
account for \( (X, Y | Z_i) \) -- both as ($X \nindep Y | Z_i$) --
but not for \( X \indep Y \) -- where they would differ.
As a result, the following equality holds:
\[
\*H'_{\mathfrak{S}(\mathcal{P}^{'(r)}), P} \cup \*H'_{\mathfrak{C}(\mathcal{P}^{'(r)}), P}
= \*H'_{\mathfrak{S}(\mathcal{P}^{''(r)}), P} \cup \*H'_{\mathfrak{C}(\mathcal{P}^{''(r)}), P}.
\]
This implies that \( \mathcal{P}^{'(r)} \) and \( \mathcal{P}^{''(r)} \)
are statistically indistinguishable, even asymptotically,
despite representing distinct MECs.
Furthermore, this results in the PAG \( \mathcal{P}^{(r)} \)
being equidistant from both \( \mathcal{P}^{'(r)} \) and \( \mathcal{P}^{''(r)} \),
contradicting the premise that the given set of hypotheses
is sufficient to distinguish between distinct MECs.

However, if the omitted relation were included and
\( P( H_{(X \nindep Y)_{P}} | \mathcal{D}) \neq
P( H_{(X \indep Y)_{P}} | \mathcal{D}) \),
then the two \( r \)-PAGs, \( \mathcal{P}^{'(r)} \) and \( \mathcal{P}^{''(r)} \),
could exhibit distinct levels of uncertainty.
This would enable the proper distinction
between \( \mathcal{P}^{(r)} \) and either of these two \( r \)-PAGs,
ensuring that their statistical differences are meaningfully captured.
This establishes the necessity of the omitted relation in
identifying and distinguishing between distinct MECs.

\item \textbf{Relations in $\*H_{\mathfrak{C}(\mathcal{P}^{(r)}), P}$
(active collider with order).}
We now establish the necessity of relations in
$\*H_{\mathfrak{C}(\mathcal{P}^{(r)}), P}$, which captures
dependencies arising from activation of collider triples with order.

Similarly from the previous case, we show that omitting a relation from the
set $\*H_{\mathfrak{C}(\mathcal{P}^{(r)}), P}$ makes
$\mathcal{P}^{(r)}$ to be equally distant from two distinct $r$-PAGs
$\mathcal{P}^{'(r)}$ and $\mathcal{P}^{''(r)}$,
contradicting the assumption that the hypothesis is irrelevant
for distinguishing PAGs that represent distinct MECs.

Let $\{X, Y\}$ be a pair of non-adjacent nodes in $\mathcal{P}^{(r)}$,
with $\*Z$ being a minimal separator of $X$ and $Y$,
such that there exists a
collider triple $\langle A, B, C\rangle$ corresponding to $(X,Y)$.
Notably, by \Cref{prop:minimality_triples_order}, $B \not \in \*Z$ and
$\langle A, B, C\rangle$ forms a collider triple with order.
Thus, $\*Z \cup \{B\}$ m-connects $X$ and $Y$.

By contradiction, suppose that the relation
$(X,Y|\*Z \cup  \{B\})$ is not necessary for characterizing the uncertainty
in identifying the \( r \)-MEC represented by \( \mathcal{P}^{(r)} \).
Such a relation would be considered in
$\*H_{\mathfrak{C}(\mathcal{P}^{(r)}), P}$.
However, under the assumption, consider a modified set of hypotheses:
$$\*H'_{\mathfrak{C}(\mathcal{P}^{(r)}), P}
= \*H_{\mathfrak{C}(\mathcal{P}^{(r)}), P} \setminus
H_{(X \nindep Y | \*Z \cup \{B\})_{P}}.$$

An an example, let $\mathcal{P}^{(r)}$ be
$X \circ\!\!\!\rightarrow Z_i \leftarrow\!\!\!\circ Y$, such that
$(X \indep Y)$.
Further, let the PAG $\mathcal{P}^{'(r)}$
consist of a single unshielded triple
$X \circ\!\!-\!\!\circ Z_i \circ\!\!-\!\!\circ Y$, where
$\*Z = \{Z_i\}$ and $X \indep Y | \*Z$.
Finally, let the PAG $\mathcal{P}^{''(r)}$ be
be a fully connected PAG between $X$, $Y$, and $Z_i$.

Under the assumption, the sets of modified hypotheses for
\( \mathcal{P}^{'(r)} \) and \( \mathcal{P}^{''(r)} \)
account for \( (X, Y | \emptyset) \) -- both as ($X \nindep Y$) --
but not for \( X \indep Y | Z_i\) -- where they would differ.
As a result, the following equality holds:
\[
\*H'_{\mathfrak{S}(\mathcal{P}^{'(r)}), P} \cup \*H'_{\mathfrak{C}(\mathcal{P}^{'(r)}), P}
= \*H'_{\mathfrak{S}(\mathcal{P}^{''(r)}), P} \cup \*H'_{\mathfrak{C}(\mathcal{P}^{''(r)}), P}.
\]
This implies that \( \mathcal{P}^{'(r)} \) and \( \mathcal{P}^{''(r)} \)
are statistically indistinguishable, even asymptotically,
despite representing distinct MECs.
Furthermore, this results in the PAG \( \mathcal{P}^{(r)} \)
being equidistant from both \( \mathcal{P}^{'(r)} \) and \( \mathcal{P}^{''(r)} \),
contradicting the premise that the given set of hypotheses
is sufficient to distinguish between distinct MECs.

However, if the omitted relation were included and
\( P( H_{(X \nindep Y | Z_i)_{P}} | \mathcal{D}) \neq
P( H_{(X \indep Y | Z_i)_{P}} | \mathcal{D}) \),
then the two \( r \)-PAGs, \( \mathcal{P}^{'(r)} \) and \( \mathcal{P}^{''(r)} \),
could exhibit distinct levels of uncertainty.
This would allow for a clear distinction between
\( \mathcal{P}^{(r)} \) and either of these two \( r \)-PAGs,
ensuring that their statistical differences are properly captured.
Thus, this underscores the necessity of including the omitted relation to
accurately capture the uncertainty in
identifying and distinguishing between distinct MECs.
\end{enumerate}

This concludes the proof.
\end{proof}


\noindent \textbf{Proof of \Cref{cor:score_equivalent}}

\begin{proof}
Due to their Markov equivalence, every MAG in an
\( r \)-PAG \( \mathcal{P}^{(r)} \) shares the same skeleton and colliders
with order. Consequently, they encode exactly the same set of conditional
(in)dependencies.

Tools for evaluating \( m \)-separation, identifying minimal separators,
and determining collider with order -- originally designed for MAGs --
can be directly extended to PAGs by using the notion of \( m \)-separation
for PAGs, which has been shown to be complete by \citep{jaber2022causal}.

From the proof of \Cref{thm:complete_hypotheses_uncertainty}, the score
$
S_{\mathcal{P}^{(r)}, P} = P(\*H_{\mathcal{P}^{(r)}, P} | \mathcal{D})
$
accounts for only (in)dependencies associated with the skeleton
and colliders with order. Since these properties are invariant across all MAGs
in a given \( r \)-PAG \( \mathcal{P}^{(r)} \), the corresponding
(in)dependency hypotheses are also shared among all models in the PAG.
This suffices to show that Markov-equivalent MAGs receive the same score,
which, asymptotically, will always be equal.

Conversely, distinct PAGs, and thus MAGs belonging to different PAGs,
necessarily differ in either at least one edge in the skeleton or in the
classification of at least one triple with order
(i.e., whether it is a collider with order or not).
By \Cref{thm:complete_hypotheses_uncertainty}, the score
$
S_{\mathcal{P}^{(r)}, P} = P(\*H_{\mathcal{P}^{(r)}, P} | \mathcal{D})
$
captures all (in)dependencies that define the skeleton and colliders with order,
thereby fully capturing the differences that distinguish PAGs or
MAGs that are not Markov-equivalent. Consequently,
these distinct structures will receive a score that, asymptotically,
are different.
\end{proof}

\noindent \textbf{Proof of \Cref{thm:dcfci_soundness}}

\begin{proof}[Proof of \Cref{thm:dcfci_soundness}]
We show that under the given assumptions, dcFCI correctly
identifies the true $r_{\text{max}}$-PAG $\mathcal{P}^{(r_{\text{max}})}$.

We leverage the soundness and completeness of FCI \cite{zhang2008fci}:
Given an accurate (faithful) set of minimal separators, FCI is sound and complete.
That is, if the correct minimal separators are identified, FCI outputs the true PAG.

If the significance level $\alpha$ is sufficiently small and an appropriate
conditional independence test is used, then all true (conditional)
independencies are observed in the dataset $\mathcal{D}$.
Consequently, the true set of minimal separators is a subset of the
observed independencies in $\mathcal{D}$.

We prove this by induction on \( r \), showing that for each iteration
\( r = 0, \dots, r_{\text{max}} \), the true \( r \)-PAG \( \mathcal{P}^{(r)} \)
is included in the candidate list \( \mathcal{L}^{(r)} \) and, under the assumption
that it ranks among the \( k \) most data-compatible candidates
(according to \( S_{\mathcal{P}^{(r)}, \mathcal{D}, \mathcal{L}^{(r)}} \)),
it is retained for the next iteration. There, it is certainly refined to
yield \( \mathcal{P}^{(r+1)} \). By induction, \( \mathcal{P}^{(r_{\text{max}})} \)
is guaranteed to be in \( \mathcal{L}^{(r_{\text{max}})} \).

At initialization (line 7), dcFCI starts with the complete graph
(implying an empty set of independencies), which is trivially sound when no
independencies are observed. This represents the most conservative starting point.

\begin{itemize}
\item \textbf{Base Case (\( r = 0 \)):}
dcFCI constructs all $0$-PAGs from every possible
subset of the observed marginal independencies.
Since the true set of marginal independencies is among theses subsets,
it is considered, yielding the true $\mathcal{P}^{(0)}$.
Hence, $\mathcal{P}^{(0)}$ is included in the candidate $\mathcal{L}^{(0)}$.
By assumption, its score $S_{\mathcal{P}^{(0)}, \mathcal{D}, \mathcal{L}^{(0)}}$
ranks among the $k$ highest in \(\mathcal{L}^{(0)}\), ensuring its
selection for iteration \(r = 1\).

\item \textbf{Inductive Step (\( r \rightarrow r+1 \)):}
Assume that at iteration \( r \), the true \( \mathcal{P}^{(r)} \) is
among the \( k \) highest-scoring \( r \)-PAGs
in \( \mathcal{L}^{(r)} \).
We show that the true \( \mathcal{P}^{(r+1)} \) will also be among
the \( k \) highest-scoring \( (r+1) \)-PAGs in \( \mathcal{L}^{(r+1)} \).

The dcFCI algorithm generates all possible \((r+1)\)-PAGs by expanding the set of minimal
separators of the current $r$-PAGs in $\mathcal{L}^{(r)}$, considering all
subsets of observed independencies conditioned on sets of size \(r+1\).
Since all true conditional independencies of size \( r+1 \) are identified,
the set of minimal separators for $\mathcal{P}^{(r)}$ is guaranteed to be
expanded to include them, thus generating
the set of true minimal separators for \( \mathcal{P}^{(r+1)} \)
and ensuring \( \mathcal{P}^{(r+1)} \) is among the
$r+1$-PAG candidates.

By assumption, the true \( \mathcal{P}^{(r+1)} \) ranks among the \( k \)
most data-compatible \( (r+1) \)-PAGs --
i.e., its score $S_{\mathcal{P}^{(r+1)}, \mathcal{D}, \mathcal{L}^{(r+1)}}$ is among
the $k$ highest. Thus, it is retained in
\( \mathcal{L}^{(r+1)} \) for the next iteration.

\item \textbf{Termination (\( r = r_{\text{max}} \))}:
At \( r_{\text{max}} \), the true
\( \mathcal{P}^{(r_{\text{max}})} \) is guaranteed to be in the output list,
as all preceding $\mathcal{P}^{(0)}, \ldots, \mathcal{P}^{(r_{\text{max}} -1)}$
were selected in previous iterations. Since \( \mathcal{P}^{(r_{\text{max}})} \)
is among the top $k$ highest-scoring $r_{\text{max}}$-PAGs, it is selected for
output in \( \mathcal{L}^{(r_{\text{max}})} \).
\end{itemize}

Thus, by induction, we demonstrate that dcFCI outputs the true
\( \mathcal{P}^{(r_{\text{max}})} \) under the given assumptions.
\end{proof}

\section{On the computational complexity of dcFCI}
\label{ape:dcfci_complexity}

\subsection{Complexity Analysis}
At each iteration \( r \), dcFCI identifies all conditional independencies given
subsets of size \( r \) for all pairs of adjacent nodes.
For \( \binom{p}{2} = O(p^2) \) pairs of variables, the number of
conditional independence tests
given subsets of size \( r \) is \( \binom{p}{2} \times \binom{p-2}{r} \).
This number grows combinatorially with \( p \) but can be
approximated as \( O(p^r) \) for moderate \( r \), assuming \( r \ll p \).
Therefore, the total number of tests is \( O(p^{r+2}) \).

Let \( n_r \) be the number of identified conditional independencies
given subsets of size \( r \).
Then, dcFCI generates an \( r \)-PAG for each of the possible subsets of
these \( n_r \) independencies (i.e., the powerset),
leading to a total of \( 2^{n_r} \) possible \( r \)-PAGs.

To generate each of these \( r \)-PAGs, dcFCI applies the FCI orientation rules
and computes their scores, which involves identifying minimal separators and
colliders with order. These tasks are done in polynomial time, with
the most demanding operation being \( O(p^5) \), even in the worst case:
\begin{itemize}
\item Finding minimal separators can be performed in at most \( O(p^2) \),
as nodes can be removed one by one \citep{tian1998finding}.
\item Applying FCI’s orientation rules takes \( O(p^5) \):
for each of the \( O(p^2) \) edge marks, FCI checks for specific
paths in \( \mathcal{P} \), a process that runs in \( O(pm) \), where \( m \)
is the number of edges. In dense (complete) graphs, where \( m = O(p^2) \),
this results in an overall \( O(p^5) \) complexity \citep{zhang2008fci}.
\item Identifying colliders via the MAG-to-MEC algorithm has complexity
\( O(p^4) \) in general \citep{claassen2022greedy}.
\end{itemize}
Therefore, each iteration of dcFCI takes \( O(2^{n_r} p^5) \).

In practice, the number $n_r$ of observed independencies
is typically much smaller than the worst-case bound of \( O(p^{r+2}) \),
making dcFCI computationally feasible in most cases.
Consequently, the complexity of each iteration $r$ of dcFCI when $n_r$ is moderate,
is $O(2^{n_r} p^5)$, which is significantly
smaller than the worst-case exponential growth \( O(2^{p^{r+2}}) \), 
which can be even super-exponential when \( r = p-2 \).

When considering that the algorithm runs for \( r_{\text{max}} \) iterations
(with \( r = 0, 1, \dots, r_{\text{max}} \)), the overall complexity depends
on the complexity of each iteration.
The total complexity of the \( r_{\text{max}} \) iterations is therefore the
sum of the complexities for each individual iteration:
\[
O\left( \sum_{r=0}^{r_{\text{max}}} 2^{n_r} p^5 \right).
\]
Therefore, for a moderate number \( n_r \) of identified independencies and
moderate values of \( r_{\text{max}} \), the growth remains manageable.

\noindent \textit{Worst-case performance:} In the worst case,
for each \( r \), \( n_r \) could be of order \( O(p^{r+2}) \).
Thus, the overall complexity across all iterations would be:
\[
O\left( \sum_{r=0}^{r_{\text{max}}} 2^{O(p^{r+2})} p^5 \right).
\]
This grows very quickly, and for large values of \( r_{\text{max}} \),
this sum becomes dominated by the highest values of \( r \).

For comparison, the adjacency search phase is the most computationally
demanding part of the FCI algorithm, with a complexity of
\( O(p^{r_{\text{max}} + 2}) \). This arises from the algorithm's search for
separating sets of up to \( r_{\text{max}} \) nodes from a set of \( p-2 \)
variables, while evaluating \( O(p^2) \) node pairs \citep{spirtes2001causation}.
small values of \( r_{\text{max}} \), the complexity remains polynomial,
making the search computationally feasible. However, when considering large
conditioning sets, particularly in the worst case where \( r_{\text{max}} = p-2 \),
the complexity also grows exponentially.

\subsection{Computational Optimizations}

Notably, dcFCI may not be efficient for when $n_r$ is not moderated.
One potential solution is to apply a threshold to the probabilities
of the conditional independencies, classifying those with high probability as
certain, and focusing only on the less probable independencies in the powerset.

Despite its higher computational cost, dcFCI implements several optimizations
to reduce runtime and make the algorithm more practical for larger datasets.
One such optimization is caching the results of conditional independence tests.
These tests are performed once and stored, so that when the same test is needed
in subsequent iterations, it can be retrieved rather than recomputed.
This significantly reduces redundant computations and helps mitigate
the exponential growth.

Additionally, dcFCI leverages parallelization to construct multiple
$r$-PAGs simultaneously. Instead of running FCI sequentially for each subset of
independencies, dcFCI utilizes parallel processing to handle different runs
concurrently, thereby accelerating the computation. This parallelization is a
significant advantage, allowing dcFCI to better handle the combinatorial growth
of possible subsets.

\section{Example Using the Straightforward PAG Score}

\begin{exmp}
\label{ex:straigthf_score_baselines}
Consider the example from \Cref{fig:dcfci_steps}, over
$p = 4$ variables, namely $A$, $B$, $X$, and $Y$. Following the formula shown
in Eq. \eqref{eq:pairwise_ci}, the total number $T$ of pairwise conditional
independencies relations when $r_{\text{max}}$, the maximum size of the
conditioning set, is set to 2 is:
$$
T = \binom{4}{2} \times \sum_{r=0}^{2} \binom{4-2}{r} = 24.
$$
Table \ref{tab:ex1_citests} presents the p-values and posterior probabilities
for testing hypotheses of independence and dependence for each pairwise conditional
independence relation among \(A\), \(B\), \(X\), and \(Y\).
The ``Truth'' column indicates the conditional dependencies and independencies
implied by the true PAG, while the ``FCI'', ``cFCI \& GPS'', ``BCCD'', and ``DCD''
columns show whether the PAGs inferred by these algorithms correctly reflect
the true (in)dependence relationships.

Using the information from this table, we can compute the Fréchet bounds for the
scores associated with the true PAG \(\mathcal{P}^*\) and the PAGs inferred by
the SOTA algorithms. The bounds for the true PAG, which was accurately recovered
by MAGSL, are \(S_{\mathcal{P}^*, \mathcal{D}} \in (0, 0.612)\).
In comparison, the bounds for the PAG inferred by FCI are
\(S_{\mathcal{P}_{\text{FCI}}, \mathcal{D}} \in (0, 0.0136)\),
while the bounds for the PAGs inferred by cFCI and GPS are
\(S_{\mathcal{P}_{\text{cFCI \& GPS}}, \mathcal{D}} \in (0, 0.0366)\).
For the BCCD PAG, as it is invalid, inferring conditional (in)dependence
relations via m-separation is not appropriate. However, the bounds for
the closest valid PAG (with a circle instead of an arrowhead into \(Y\))
are \(S_{\mathcal{P'}_{\text{BCCD}}, \mathcal{D}} \in (0, 0.1848)\).
Finally, the bounds for the DCD PAG are
\(S_{\mathcal{P}_{\text{DCD}}, \mathcal{D}} \in (0, 0.0366)\).

While the Fréchet bounds for all PAGs overlap, making it difficult to identify
statistically significant differences, a closer look at the upper bounds reveals
that the true (and MAGSL) PAG aligns more closely with the data than those inferred
by other algorithms. The Fréchet upper bound, determined by the minimum of the
individual probabilities, indicates that the true PAG's score is supported by
the strongest evidence in the data, with at least 61.2\% probability for its implied
(in)dependencies. In contrast, other PAGs represent hypotheses with much lower
probabilities, such as \( (X \nindep Y \mid B) \) (FCI PAG) at 1.36\%,
\( (A \nindep Y) \) (all but BCCD PAG) at 3.66\%, and \( (B \nindep X \mid A) \)
(cFCI, GPS, BCCD PAGs) at 18.46\%. This highlights the need for further improvement
in causal discovery algorithms, particularly in accounting for uncertainty in
conditional independence tests and achieving a closer alignment with the data.

\end{exmp}

\section{Example Using the MEC-targeted PAG Score}
\begin{exmp}
\label{ex:mec_score_FCI_vs_cFCI}
Let us revisit the example in Figure \ref{fig:ex_1} and focus on comparing the
PAGs inferred by FCI and cFCI, which differ only in the orientation of the triple
\(\langle A, B, Y \rangle\).

For illustration, suppose we narrowed the list down to only include
the FCI and cFCI PAGs, allowing us to apply the proposed MEC-targeted compatibility
score. Additionally, we set \( r = 2 \), which, enables us to evaluate the PAGs
after considering independence relations conditional on every possible
conditioning set.

Both PAGs share the same skeleton, with
\(\mathbf{E}^{(r)} = \{\{X,A\}, \{A,B\}, \{B,Y\}\}\). As a result, both
\(\*H_{\mathfrak{S}\left(\mathcal{P_{\text{FCI}}}^{(r)}\right), P}\) and
\(\*H_{\mathfrak{S}\left(\mathcal{P_{\text{cFCI}}}^{(r)}\right), P}\) include
all dependence relationships between adjacent pairs of variables, conditioned on
every set of up to \(r=2\) variables.

The FCI PAG implies that the empty set is the minimal separator
for the pairs \(\{A, Y\}\), \(\{B, X\}\), and \(\{X, Y\}\). Consequently,
\(\*H_{\mathfrak{S}\left(\mathcal{P_{\text{FCI}}}^{(r)}\right), P}\) also
includes the marginal independencies \( (A \indep Y), (B \indep X)\), and
\((X \indep Y) \). The separation of \(A\) and \(Y\) led to the formation
of the collider triple \(\langle A, B, Y \rangle\), meaning that
\(\*H_{\mathfrak{C}\left(\mathcal{P_{\text{FCI}}}^{(r)}\right), P}\) includes
the hypothesis \( (A \nindep Y | B) \). Additionally, the separation of \(B\) and
\(X\) led to the formation of the collider triple \(\langle B, A, X \rangle\),
so \(\*H_{\mathfrak{C}(\mathcal{P}^{(r)}), P}\) also includes the hypothesis
\((B \nindep X | A)\).

In contrast, the cFCI PAG implies that the empty set is a minimal
separator for the pairs \(\{A, Y\}\) and \(\{X, Y\}\), but not for \(\{B, X\}\).
This arises from the uncertainty regarding its minimal separator, which could
instead be \(\{A\}\). If we consider this possibility,
\(\*H_{\mathfrak{S}\left(\mathcal{P_{\text{cFCI}}}^{(r)}\right), P}\) includes
the independencies \((A \indep Y)\) and \((X \indep Y)\), as well as
\((B \indep X | A)\) and the dependence hypothesis \((B \nindep X)\), implied by
the minimality of the separator.
Notably, this separation led to the formation of the non-collider triple
\(\langle B, A, X \rangle\), so \(\*H_{\mathfrak{D}(\mathcal{P}^{(r)}), P}\)
also includes the hypothesis \((B \nindep X)\), which is already accounted for.

To compare both PAGs, we focus on independence relations
that are relevant for either model but differ between them. These are
exactly \(( B, X |\emptyset)\) and \(( B, X | A)\).
The Fréchet bounds of the score for the FCI PAG are
\( P( \{(X \indep B), (X \nindep B| A)\} )  = (0.486, 0.671) \),
while the Fréchet bounds of the score for the cFCI PAG are
\( P(\{ (X \nindep B), (X \indep B | A)\}) = (0, 0.185) \). Based on these scores,
a statistically meaningful decision can be made, favoring the FCI PAG as the best of the two.
\end{exmp}

\section{Simulations: Technical Details}
\label{sec:simulation_details}

The \textbf{FCI} and \textbf{cFCI} algorithms were run using their
implementations in the \texttt{pcalg} R package \cite{pcalg2012}.

For the other SOTA algorithms, we used the lastest version of the code provided
by their respective authors:

\begin{itemize}
    \item \textbf{BCCD}: We used the latest R implementation available on GitLab
    at \url{https://gitlab.science.ru.nl/gbucur/RUcausal/}.
    \item \textbf{DCD}: The implementation is available on GitLab at
    \url{https://gitlab.com/rbhatta8/dcd}. To learn MAGs, we set
    \texttt{admg\_class = "ancestral"} and kept all other hyperparameters at their
    default values, including \texttt{num\_restarts = 5}, which runs the algorithm five
    times and selects the best result.
    \item \textbf{MAGSL}: Implemented in C++ and available at
    \url{https://www.cs.helsinki.fi/group/coreo/magsl/}. Following the
    example in their README file, we set
    \texttt{max-vars = 5} (maximum number of variables),
    \texttt{max-comp = 4} (c-component size limit), and \texttt{max-pars = 6}
    (maximum number of parents), ensuring comprehensive search coverage for
    learning 5-variable MAGs.
    \item \textbf{GPS}: Implemented in MATLAB and available on GitHub at
    \url{https://github.com/tomc-ghub/gps_uai2022}. We used the hybrid version
    (\texttt{GPS\_RunType = 2}), which runs the baseline search by default and
    switches to the extended version only when it gets stuck.
\end{itemize}

The performance metrics are briefly described as follows:
\begin{description}
\item[Structural Hamming Distance (SHD):] Quantifies the total number of edgemark
(circle, tail, or arrowhead) modifications needed to convert the inferred PAG
into the true PAG, providing an overall measure of structural accuracy.
\item[False Discovery Rate (FDR):]  Represents the proportion of incorrectly
inferred definite relationships -- ancestralities (adjacencies with an arrowhead),
non-ancestralities (adjacencies with a tail), or non-adjacencies
(missing edges) -- among all inferred definite relationships. It quantifies the
fraction of definite inferences that are incorrect, providing insight into the
algorithm's reliability in identifying the invariances
(independencies and non-circles) of the MEC.
\item[False Omission Rate (FOR):]  Represents the proportion of incorrectly
inferred non-definite relationships -- adjacencies with a circle -- among all
inferred non-definite relationships. It quantifies the fraction of undetermined
relationships that are incorrect, providing insight into the algorithm's
reliability in identifying the non-invariances (circles) of the MEC.
\end{description}

\vspace{0.5em}
\noindent \textbf{Selecting a valid MAG in a PAG:}
A valid MAG within the MEC represented by a PAG can be obtained by applying
the arrowhead augmentation method of \cite{zhang2008causal} and then orienting
each chordal component to ensure the resulting graph is a valid MAG.
The \texttt{pag2magAM} function in the \texttt{pcalg} R package \citep{pcalg2012}
facilitates this process.

\vspace{0.5em}
\noindent \textbf{Verifying the validity of a PAG:}
A PAG is considered valid if it accurately represents the MEC of a valid MAG.
To verify this, we first select a valid MAG in the given PAG and then check whether
the PAG reconstructed from this MAG matches the original.
If so, the given PAG correctly encodes an MEC.

\vspace{0.5em}
\noindent \textbf{Computing the BIC for a PAG:}
To compute the BIC of a PAG, we first select a valid MAG in the given
PAG and then estimate the BIC of a Gaussian linear SEM
that aligns with the MAG structure and best fits the observed data
using the \texttt{SEMgraph} R package \citep{grassi2022semgraph}.

\section{Simulation Results with Gaussian Data}
\label{sec:app_sim_gaussian}

Due to uncertainty in the causal orientations, dcFCI can output a list of
multiple PAGs identified as equally compatible with the data.
To better understand the extent of this uncertainty, we analyze the number of
PAGs generated by dcFCI. In the following, we present the distribution of PAGs
in the output lists across 300 simulations for each sample size (\(N\)) considered.
When we set \(k = 1\), the following were the numbers of PAGs generated by dcFCI
across all simulations:

\begin{itemize}
    \item For \(N = 1,000\): In 288 simulations, the output list contained
    a single PAG, while the remaining 12 simulations returned lists with 2 PAGs.

    \item For \(N = 5,000\): Similarly, 288 simulations produced a single PAG
    in the output list. The remaining 12 simulations were distributed as follows:
    10 simulations returned lists with 2 PAGs, 1 simulation returned a list with
    3 PAGs, and 1 run returned a list with 5 PAGs.

    \item For \(N = 10,000\): Almost all simulations (298) resulted in a single
    PAG in the output list, with only 2 simulations producing lists with 2 PAGs.

    \item For \(N = 50,000\): All 300 simulations consistently produced output
    lists with a single PAG.
\end{itemize}

In contrast, for $k = 2$, the following were the numbers of PAGs generated by
dcFCI across all simulations:

\begin{itemize}
    \item For \(N = 1,000\), 283 simulations yielded a single PAG. Among the
    remaining 17, 12 produced 2 PAGs, 3 produced 3 PAGs, and 2 produced 4 PAGs.

    \item For \(N = 5,000\), 257 simulations returned a single PAG, while 24
    produced 2 PAGs, 8 produced 3 PAGs, and 11 generated lists ranging from 4 to 14 PAGs.

    \item For \(N = 10,000\), 291 simulations resulted in a single PAG.
    The remaining 9 were distributed as follows: 4 produced 2 PAGs,
    2 produced 3 PAGs, 2 produced 4 PAGs, and 1 produced 14 PAGs.

    \item For \(N = 50,000\), all 300 simulations consistently produced a single PAG.
\end{itemize}

For larger \( k \), more PAGs are evaluated, increasing the likelihood of
selecting an intermediate PAG that is not among the top \( k' \)-scored ones
for \( k' < k \) but may still lead, in subsequent iterations, to a more refined
PAG that fits the data equally well or better and potentially aligns with the true PAG.

Note, however, that selecting more PAGs increases their divergence in hypotheses,
reducing the number of shared assumptions treated as certain and raising
the likelihood of multiple PAGs being considered equally compatible with the data.
Nevertheless, simulation results indicate that as the sample size (\(N\)) grows,
dcFCI more frequently returns output lists containing a single PAG, demonstrating
that larger datasets enhance certainty about the plausible PAGs.

\section{Simulation Results with Mixed Data Types}

As in the analysis of the Gaussian datasets, we illustrate how dcFCI's
certainty about the plausible PAGs increases with the sample size. Below, we
present the distribution of the number of output PAGs generated by dcFCI across
all 300 simulations conducted for each sample size considered.
\begin{itemize}
    \item For \(N = 1,000\): In 254 simulations, the output list contained a
    single PAG. The remaining 34 simulations were distributed as follows: 20
    simulations returned lists with 2 PAGs, 8 simulations returned lists with 3
    PAGs, and 4 run returned a list with 4 PAGs.
    \item For \(N = 5,000\): 290 simulations produced a single PAG in the output
    list, while 9 simulations returned lists with 2 PAGs, and 1
    simulation returned a list with 3 PAGs.
    \item For \(N = 10,000\): 291 simulations resulted in a single PAG in the output
    list, with 9 simulations producing lists with 2 PAGs.
    \item For \(N = 50,000\): All 300 simulations consistently produced output lists
    with a single PAG.
\end{itemize}

In contrast, for $k = 2$, the following were the numbers of PAGs generated by
dcFCI across all simulations:

\begin{itemize}
    \item For \(N = 1,000\), 235 simulations yielded a single PAG, while the
    remaining 65 were distributed as follows: 41 produced 2 PAGs, 8 produced 3 PAGs,
    and 16 generated lists ranging from 4 to 33 PAGs.

    \item For \(N = 5,000\), 270 simulations resulted in a single PAG,
    22 produced 2 PAGs, 5 produced 3 PAGs, and 3 produced 4 PAGs.

    \item For \(N = 10,000\), 288 simulations returned a single PAG,
    while 10 produced 2 PAGs, 1 produced 7 PAGs, and 1 produced 20 PAGs.

    \item For \(N = 50,000\), 298 simulations generated a single PAG, and 2 produced 2 PAGs.
\end{itemize}

This analysis clearly demonstrates that, with larger sample sizes, the algorithm
more consistently converges toward a single, most probable PAG.

\section{Further details on the application to the Diabetes Health Indicators Dataset}
\label{sec:appe_application}

We used the Diabetes Health Indicators Dataset (DHID),
available on Kaggle at
\url{https://www.kaggle.com/datasets/alexteboul/diabetes-health-indicators-dataset},
It is a cleaned and curated version of the Behavioral Risk Factor
Surveillance System (BRFSS) dataset, also available on
Kaggle at \url{https://www.kaggle.com/datasets/cdc/behavioral-risk-factor-surveillance-system},
under a Public Domain license.
The BRFSS dataset is one of the largest ongoing
health surveys, conducted annually by the Centers for Disease Control and
Prevention (CDC) via telephone in the United States. The DHID is a
cross-sectional subset of the 2015 BRFSS data, including
complete responses from $253,680$ individuals.

Below, we describe the 16 out of the
21 variables we selected for our application, as they are all objectively
quantifiable and clinically relevant, offering direct insight into
the underlying health conditions.

\begin{itemize}
\item \textit{Diabetes}: 0 (no diabetes), 1 (prediabetes), 2 (diabetes)
\item \textit{Sex}: 0 (female), 1 (male)
\item \textit{Age}: A 13-level age category, with 1 (18-24), 9 (60-64), and 13 (80 or older)
\item \textit{Body Mass Index (BMI)}: A continuous variable, ranging from 12 to 98 (mean $\pm$ sd: 28.4 $\pm$ 6.61)
\item \textit{High Blood Pressure (HighBP)}: 0 (no), 1 (yes)
\item \textit{High Cholesterol (HighChol)}: 0 (no), 1 (yes)
\item \textit{History of Stroke (Stroke)}: 0 (no), 1 (yes)
\item \textit{History of Coronary Heart Disease or Myocardial Infarction (HeartDiseaseorAttack)}: 0 (no), 1 (yes)
\item \textit{Physical Activity in the past 30 days, excluding job-related activity (PhysActivity)}: 0 (no), 1 (yes)
\item \textit{Smoking Status (Smoker)}: Whether the individual has smoked at least 100 cigarettes in their lifetime (5 packs)
\item \textit{Heavy Alcohol Consumption (HvyAlcoholConsump)}: 0 (no), 1 (yes) --  for adult men, more than 14 drinks per week, for adult women, more than 7 drinks per week
\item \textit{Consumption of a fruit per day (Fruits)}: 0 (no), 1 (yes)
\item \textit{Consumption of a vegetable per day (Veggies)}: 0 (no), 1 (yes)
\item \textit{Income per year (Income)}: From 1 to 8, where 1 ($<$ \$10,000), 5 ($<$ \$35,000), and 8 ($\geq$ \$75,000)
\item \textit{Health Care Coverage (AnyHealthcare)}: 0 (no),1 (yes)
\item \textit{Education Level (Education)}: From 1 to 6, where 1 (never attended school or only kindergarten),
2 (grades 1–8, elementary), 3 (grades 9–11, some high school), 4 (grade 12 or GED, high school graduate),
5 (some college or technical school, 1–3 years), and 6 (college graduate, 4 years or more)
\end{itemize}

\begin{figure}[H]
\begin{minipage}{\linewidth}
    \begin{minipage}{0.32\linewidth}
     \begin{titlepanel}[width=\linewidth]
        {\small \bf dcFCI}
     \end{titlepanel}
    \end{minipage}
    \begin{minipage}{0.32\linewidth}
     \begin{titlepanel}[width=\linewidth]
        {\small \bf FCI}
     \end{titlepanel}
    \end{minipage}
    \begin{minipage}{0.32\linewidth}
     \begin{titlepanel}[width=\linewidth]
        {\small \bf cFCI}
     \end{titlepanel}
    \end{minipage}
\end{minipage}

\vspace{1em}
\begin{minipage}{\linewidth}
    \begin{minipage}{0.32\linewidth}
    \subfloat[]{
        \centering
        \includegraphics[width=1.2\linewidth]{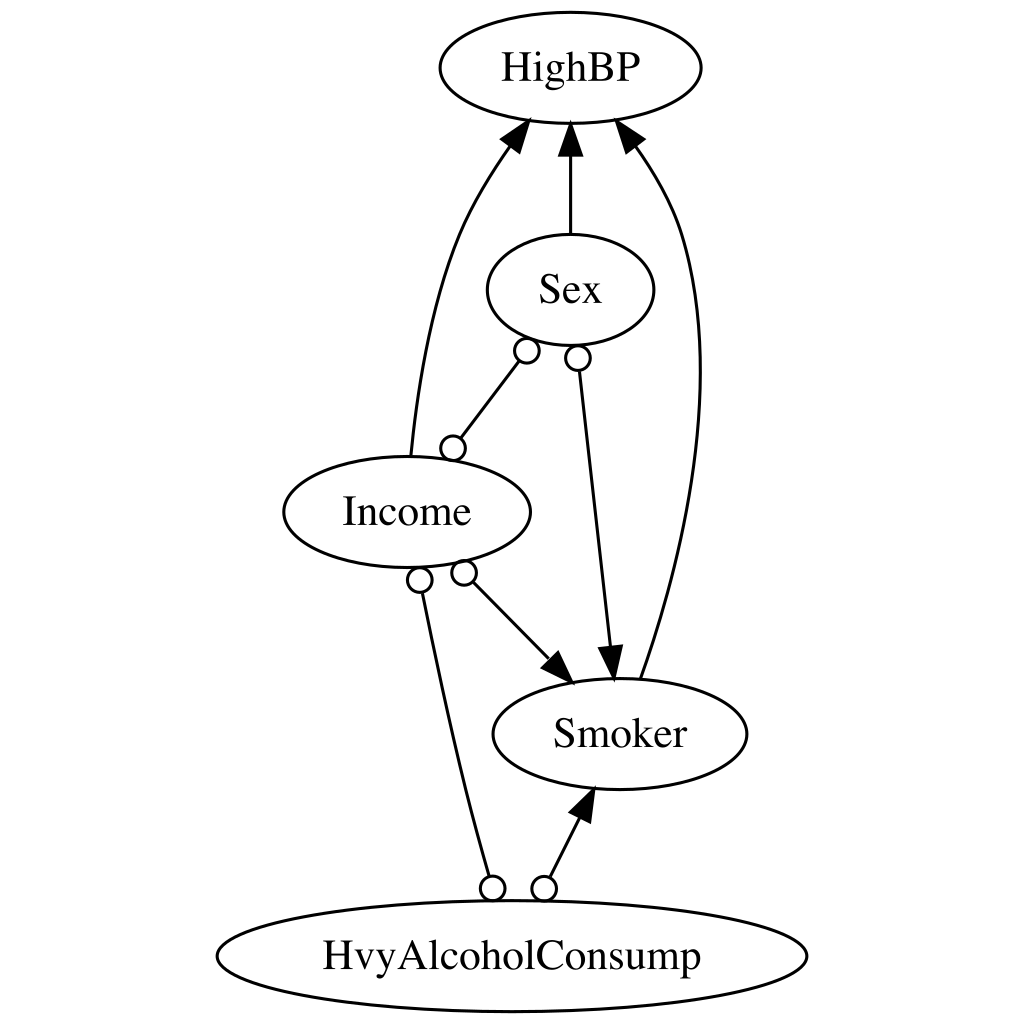}
    }
    \end{minipage}
   \begin{minipage}{0.32\linewidth}
    \subfloat[]{
        \centering
        \includegraphics[width=1.2\linewidth]{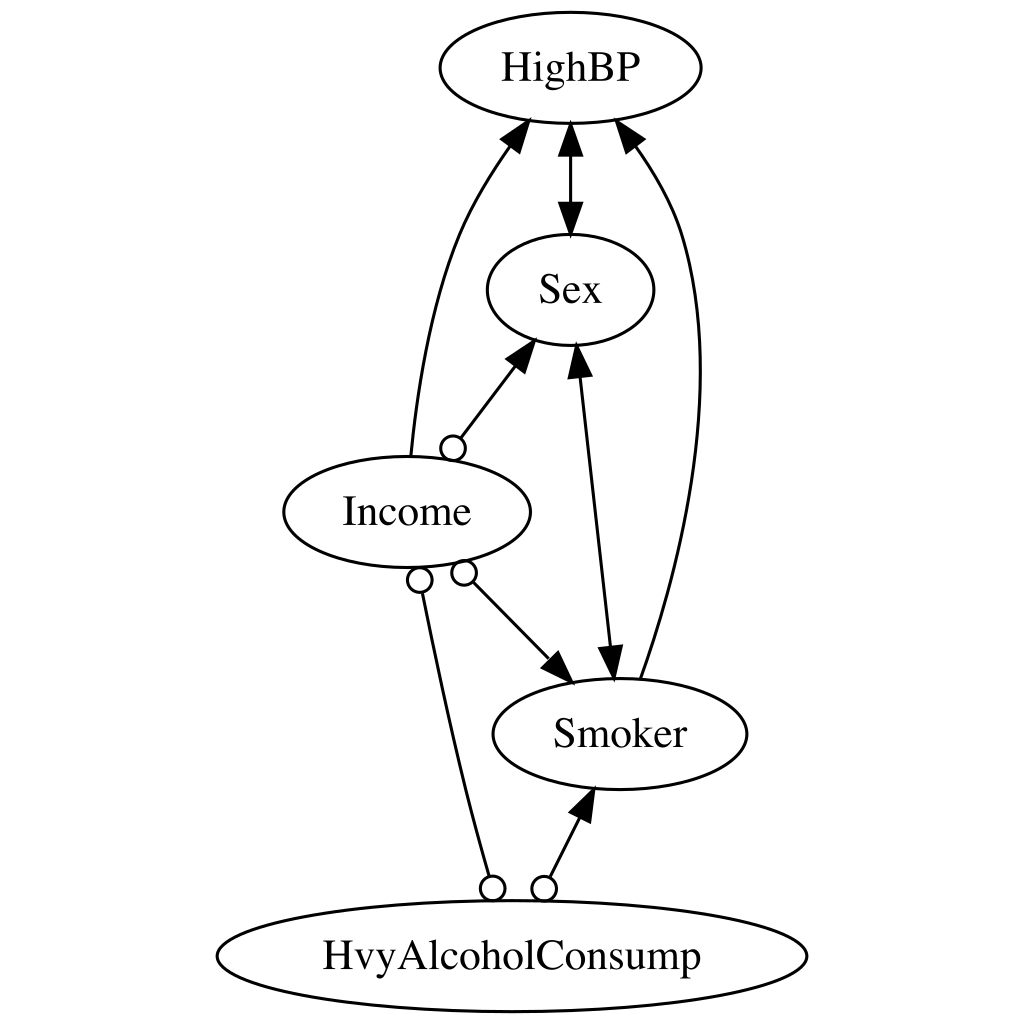}
    }
    \end{minipage}
   \begin{minipage}{0.32\linewidth}
        \centering
    \subfloat[]{
        \includegraphics[width=1.2\linewidth]
        {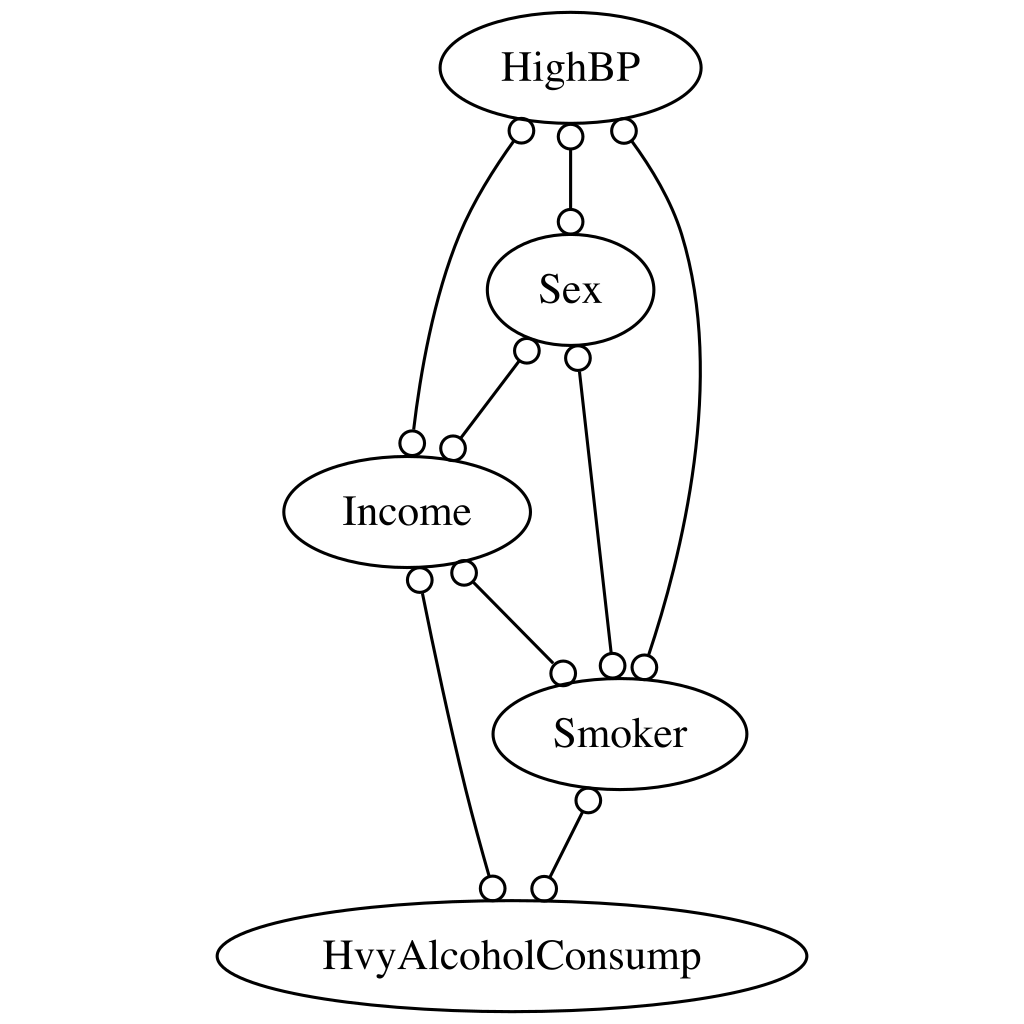}
    }
    \end{minipage}
\end{minipage}

\begin{minipage}{\linewidth}
    \begin{minipage}{0.32\linewidth}
    \subfloat[]{
        \centering
        \includegraphics[width=1.2\linewidth]{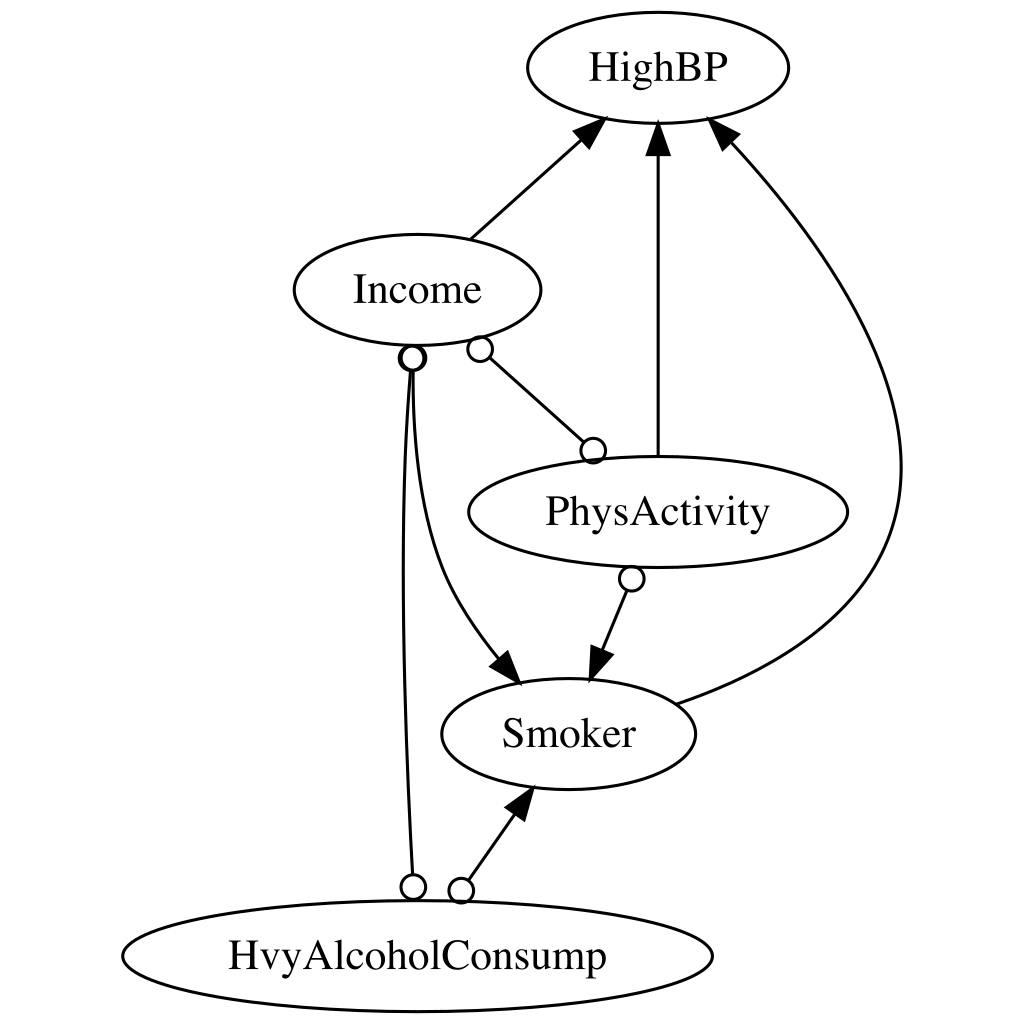}
        }
    \end{minipage}
   \begin{minipage}{0.32\linewidth}
   \subfloat[]{
        \centering
        \includegraphics[width=1.2\linewidth]{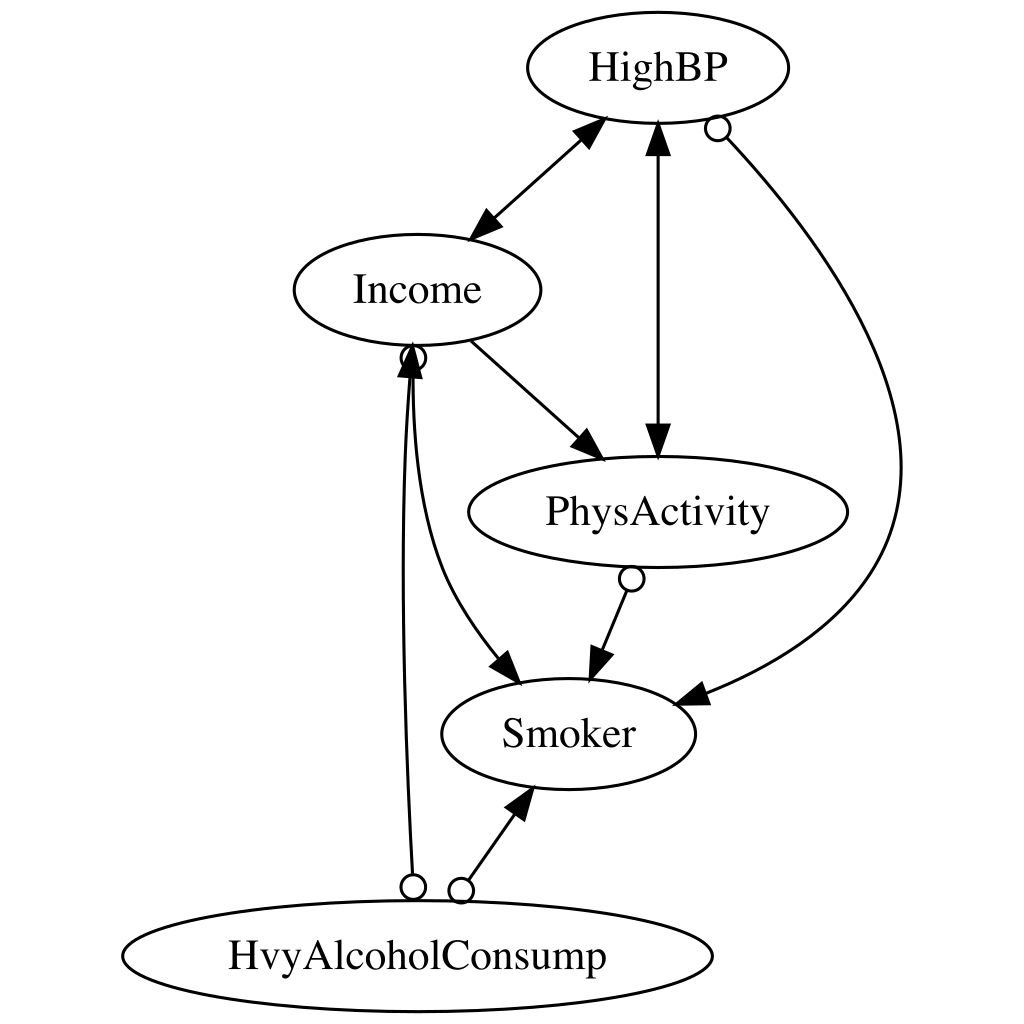}
        }
    \end{minipage}
 \begin{minipage}{0.32\linewidth}
 \subfloat[]{
        \centering
        \includegraphics[width=1.2\linewidth]
        {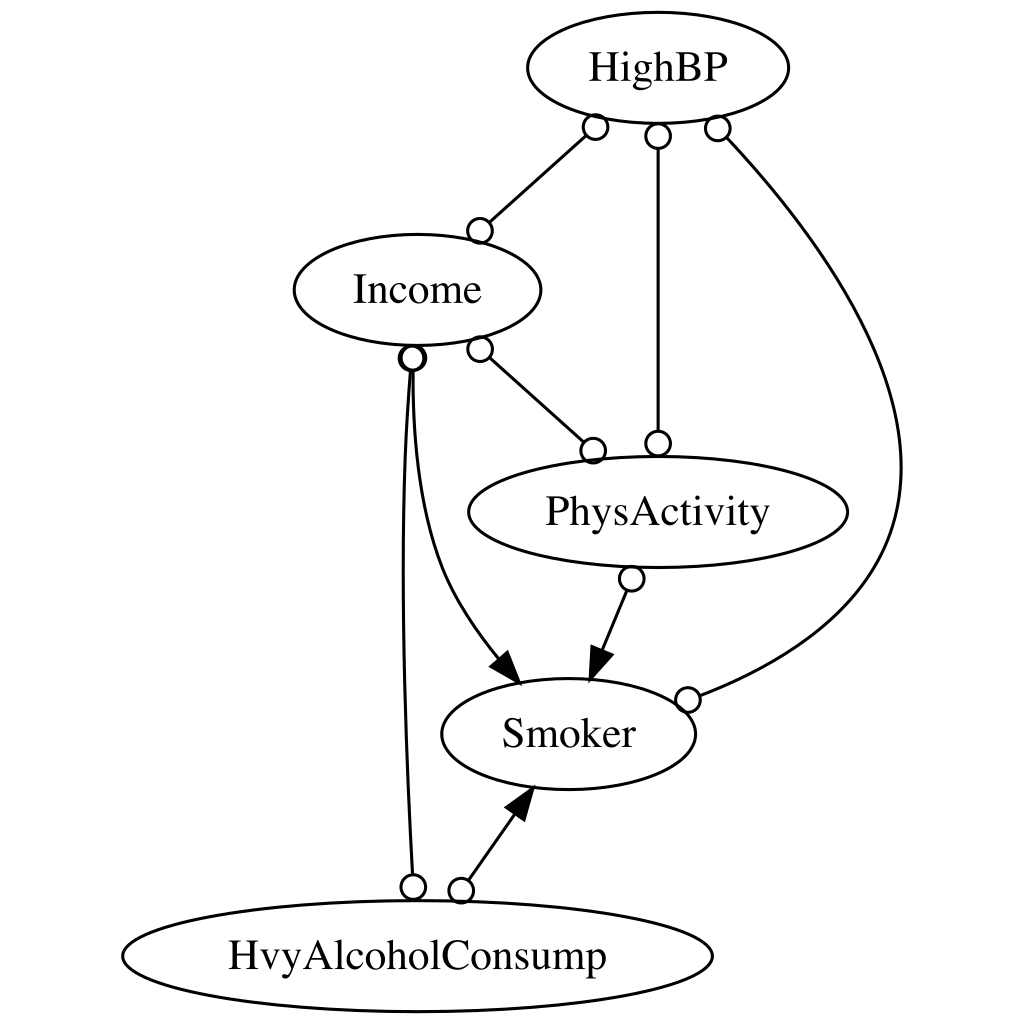}
    }
    \end{minipage}
\end{minipage}

\begin{minipage}{\linewidth}
    \begin{minipage}{0.32\linewidth}
    \subfloat[]{
    \centering
        \includegraphics[width=1.2\linewidth]{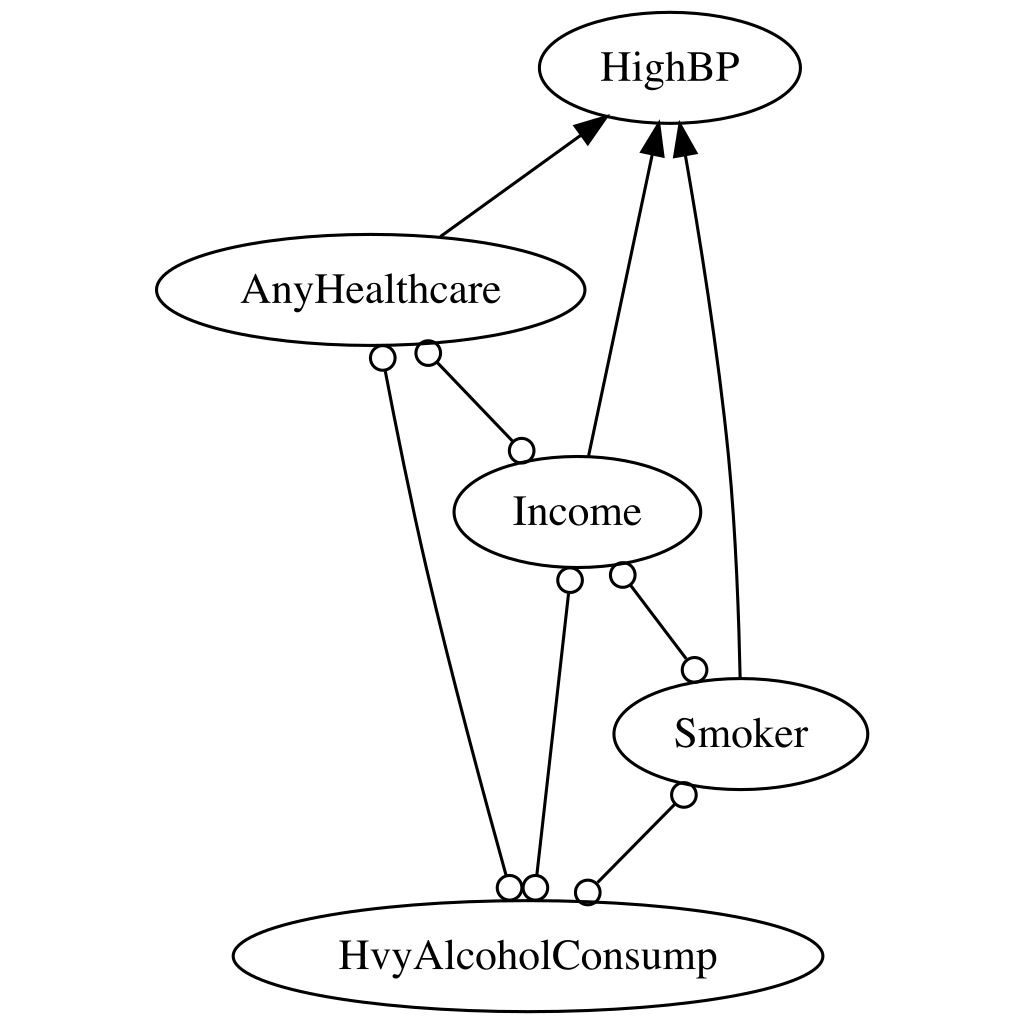}
    }
    \end{minipage}
   \begin{minipage}{0.32\linewidth}
    \subfloat[]{
    \centering
        \includegraphics[width=1.2\linewidth]{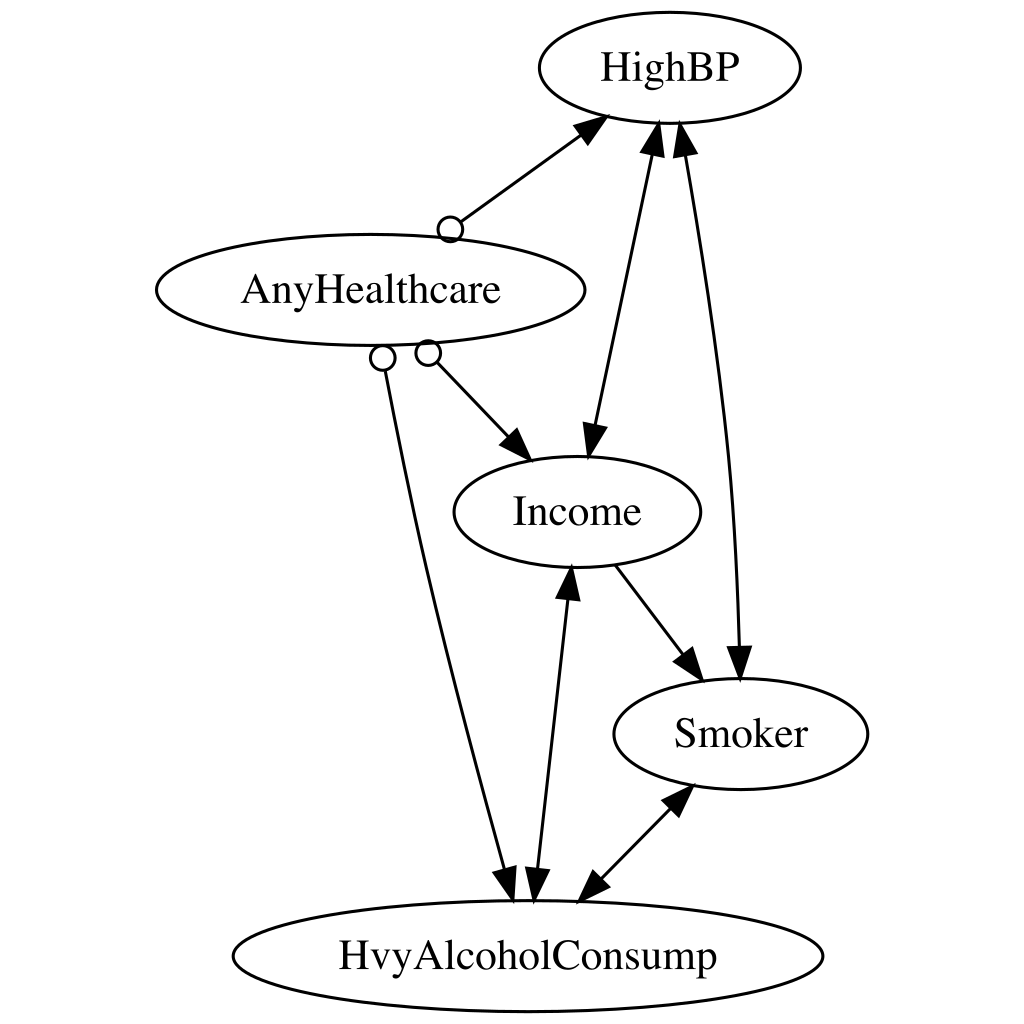}
    }
    \end{minipage}
   \begin{minipage}{0.32\linewidth}
    \subfloat[]{
    \centering
        \includegraphics[width=1.2\linewidth]
        {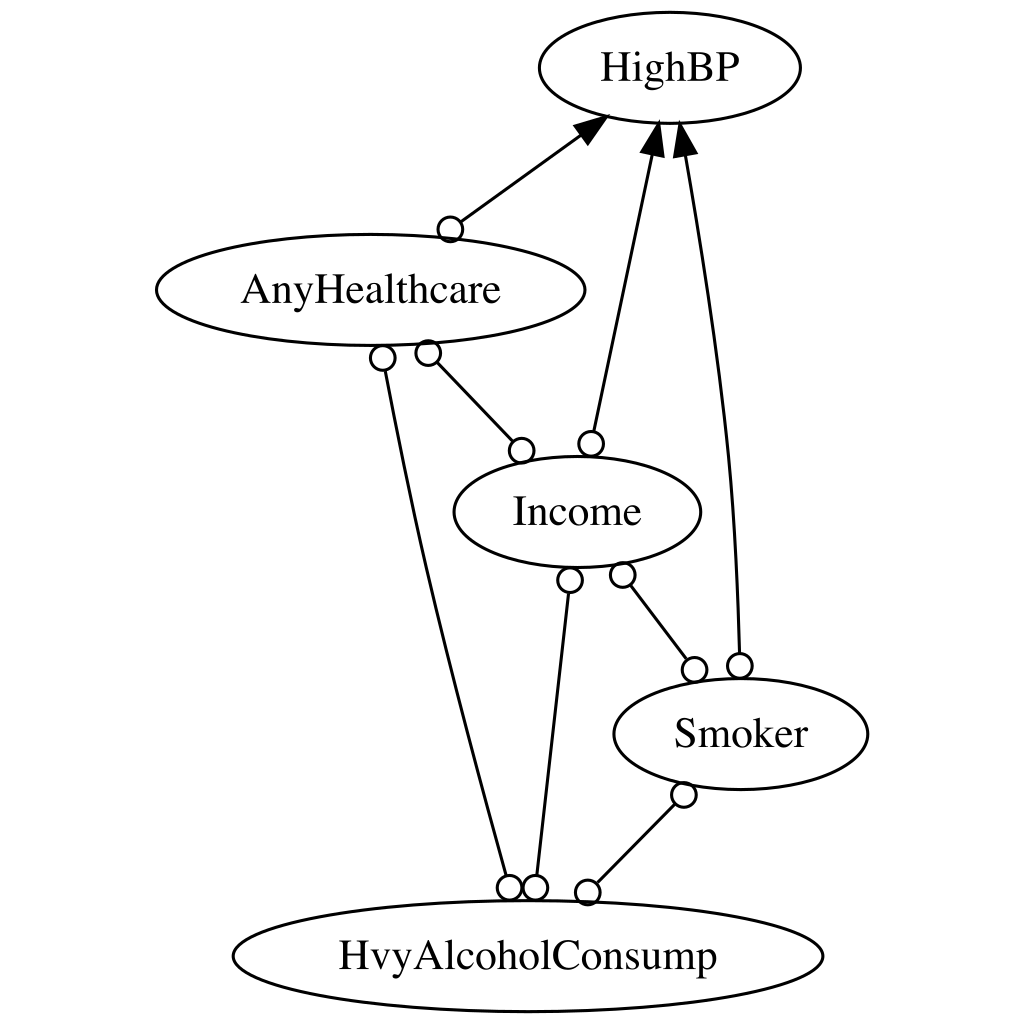}
    }
    \end{minipage}
\end{minipage}
\caption{Comparison of PAGs learned by dcFCI (left), FCI (center), and cFCI (right)
for five variables:
\textit{HighBP}, \textit{Sex}, \textit{Income}, \textit{Smoker}, and
\textit{HvyAlcoholComsump}.
dcFCI inferred PAGs with higher compatibility scores
(bounds: (a) [0, 0.0209], (d) [0, 0.0155], (g) [0, 0.0249])
compared to FCI ((b) [0, 0.0119], (e) [0, 0.0134], (h) [0, 0.00892]).
cFCI yields invalid PAGs in two cases ((f) and (i)), with one valid result (c) $[0, 2.82 × 10^{-10}]$.}
\label{fig:highBP_extra}
\end{figure}

\begin{figure}[ht]
\begin{minipage}{\linewidth}
    \begin{minipage}{0.32\linewidth}
     \begin{titlepanel}[width=\linewidth]
        {\small \bf dcFCI}
     \end{titlepanel}
    \end{minipage}
    \begin{minipage}{0.32\linewidth}
     \begin{titlepanel}[width=\linewidth]
        {\small \bf FCI}
     \end{titlepanel}
    \end{minipage}
    \begin{minipage}{0.32\linewidth}
     \begin{titlepanel}[width=\linewidth]
        {\small \bf cFCI}
     \end{titlepanel}
    \end{minipage}
\end{minipage}

\vspace{1em}
\begin{minipage}{\linewidth}
    \begin{minipage}{0.32\linewidth}
    \subfloat[]{
        \hspace{-1.7em}
        \includegraphics[width=1.2\linewidth, left]{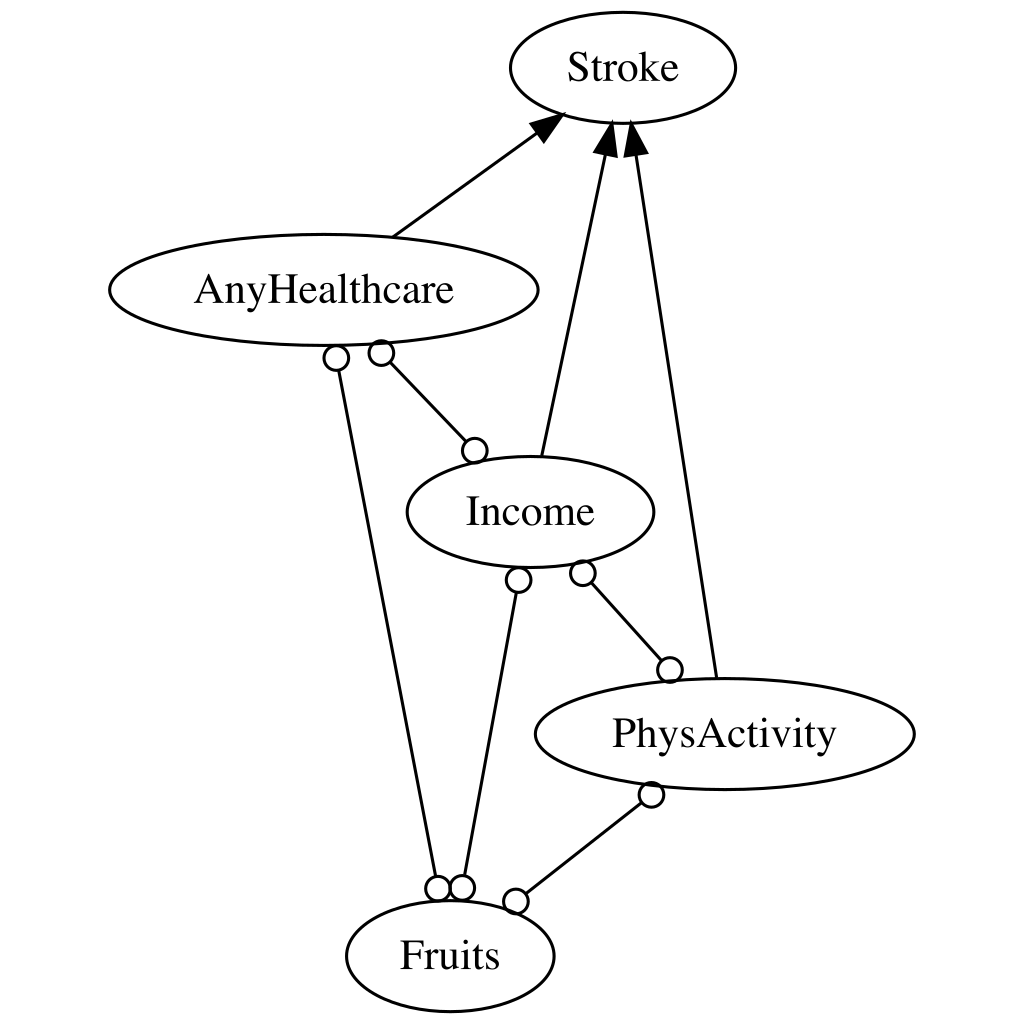}
    }
    \end{minipage}
   \begin{minipage}{0.32\linewidth}
    \subfloat[]{
        \hspace{-1.7em}
        \includegraphics[width=1.2\linewidth, left]{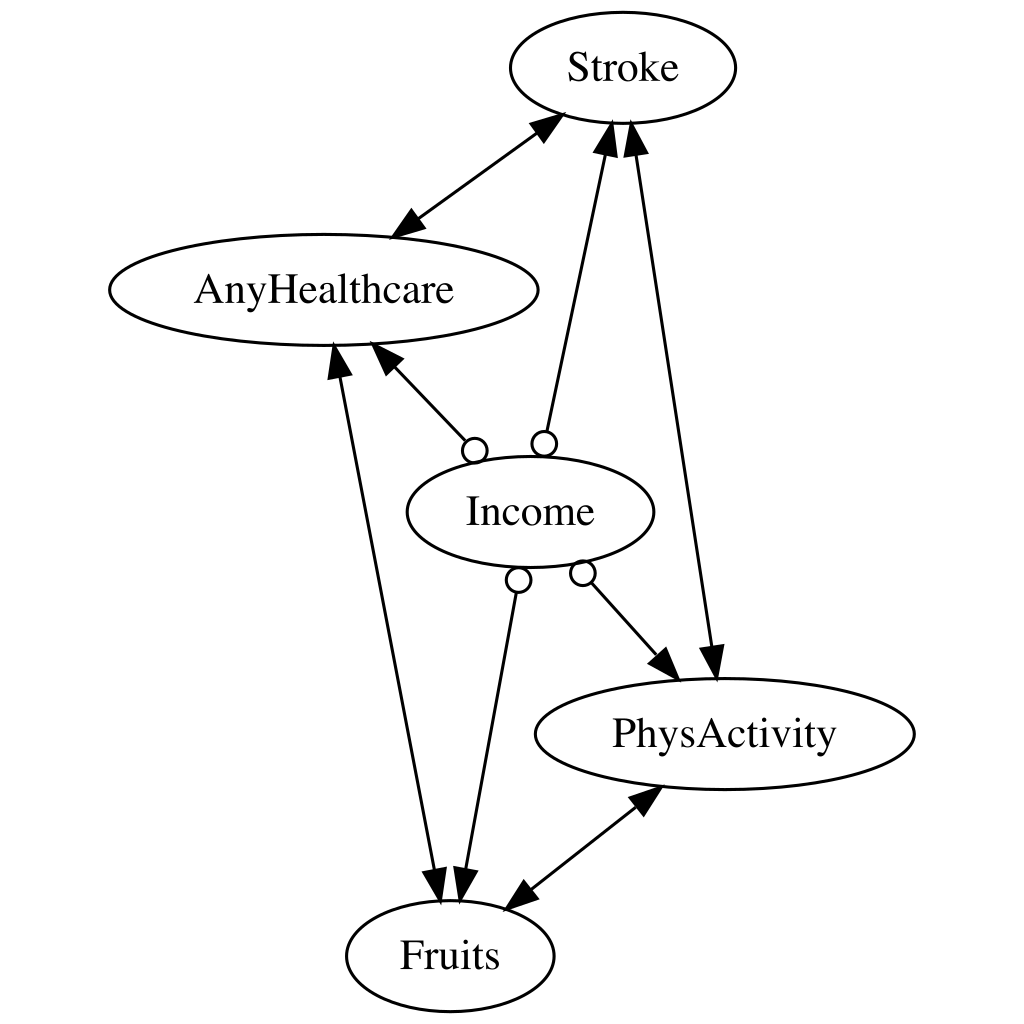}
    }
    \end{minipage}
 \begin{minipage}{0.32\linewidth}
    \subfloat[]{
        \hspace{-1.7em}
        \includegraphics[width=1.2\linewidth, left]
       {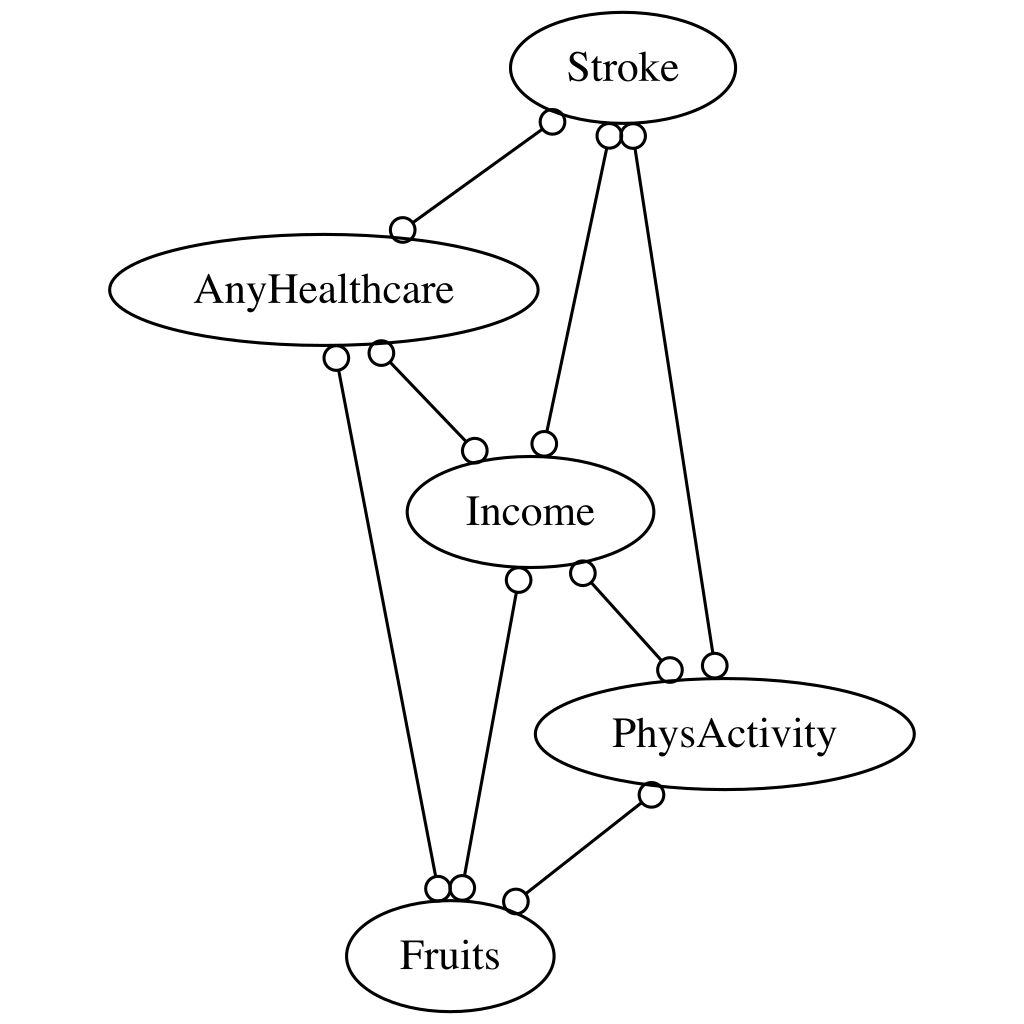}
    }
    \end{minipage}
\end{minipage}

\begin{minipage}{\linewidth}
    \begin{minipage}{0.32\linewidth}
    \subfloat[]{
        \hspace{-1.7em}
        \includegraphics[width=\linewidth, left]{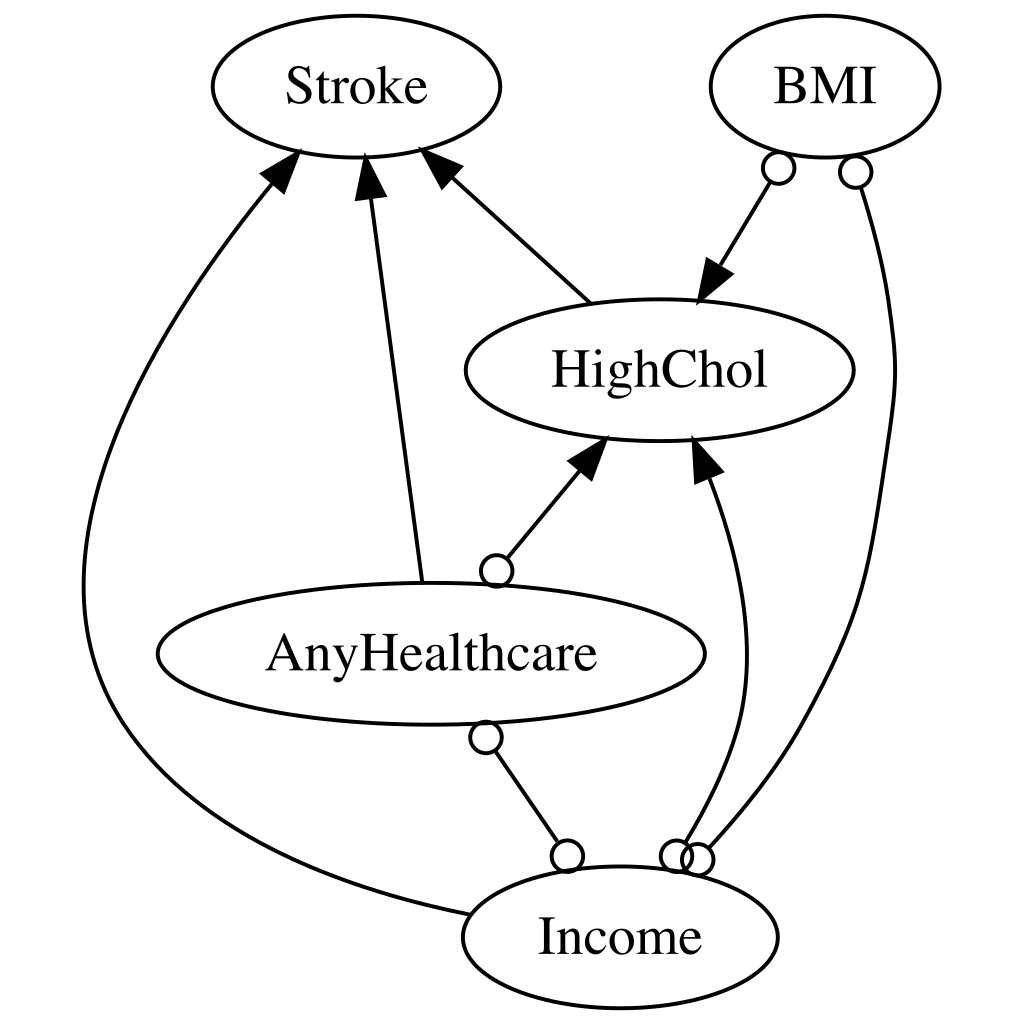}
    }
    \end{minipage}
   \begin{minipage}{0.32\linewidth}
    \subfloat[]{
        \hspace{-1.7em}
        \includegraphics[width=\linewidth, left]{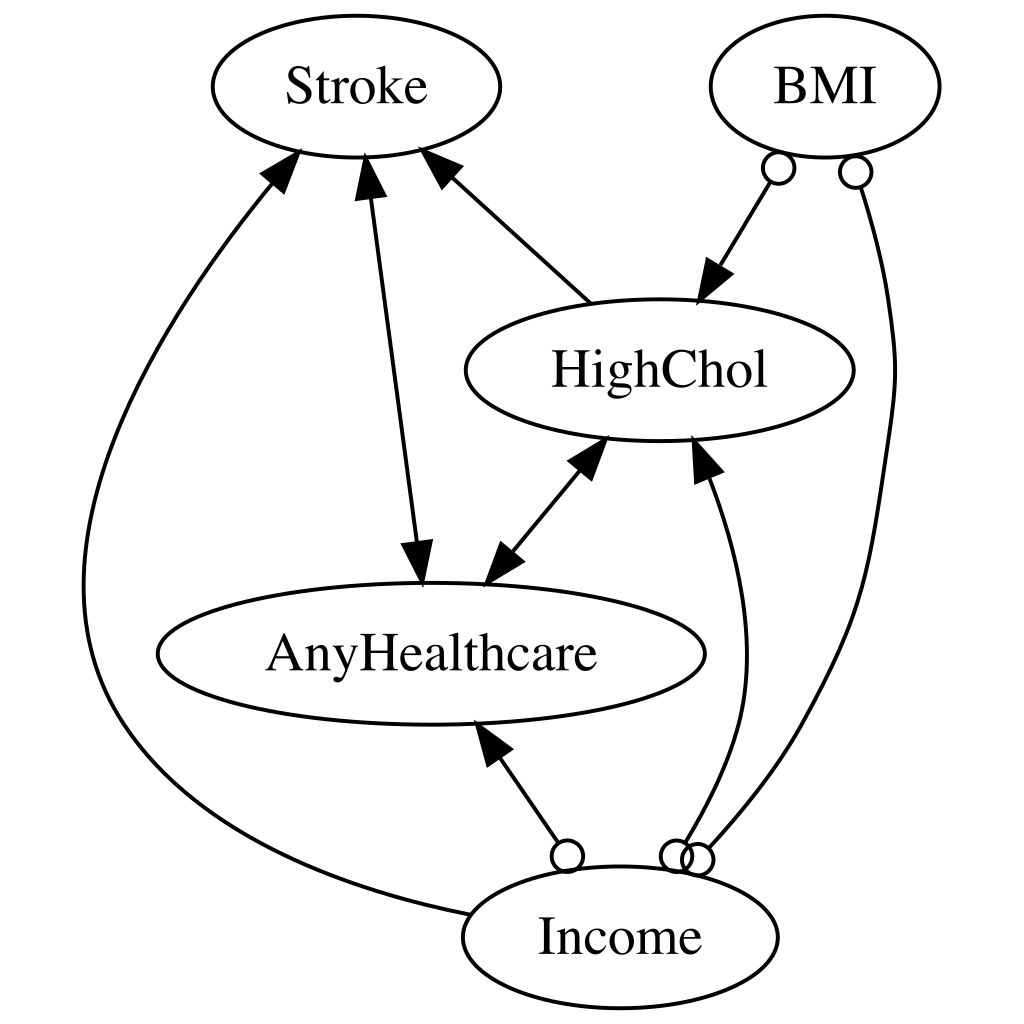}
    }
    \end{minipage}
 \begin{minipage}{0.32\linewidth}
    \subfloat[]{
        \hspace{-1.7em}
        \includegraphics[width=\linewidth, left]
       {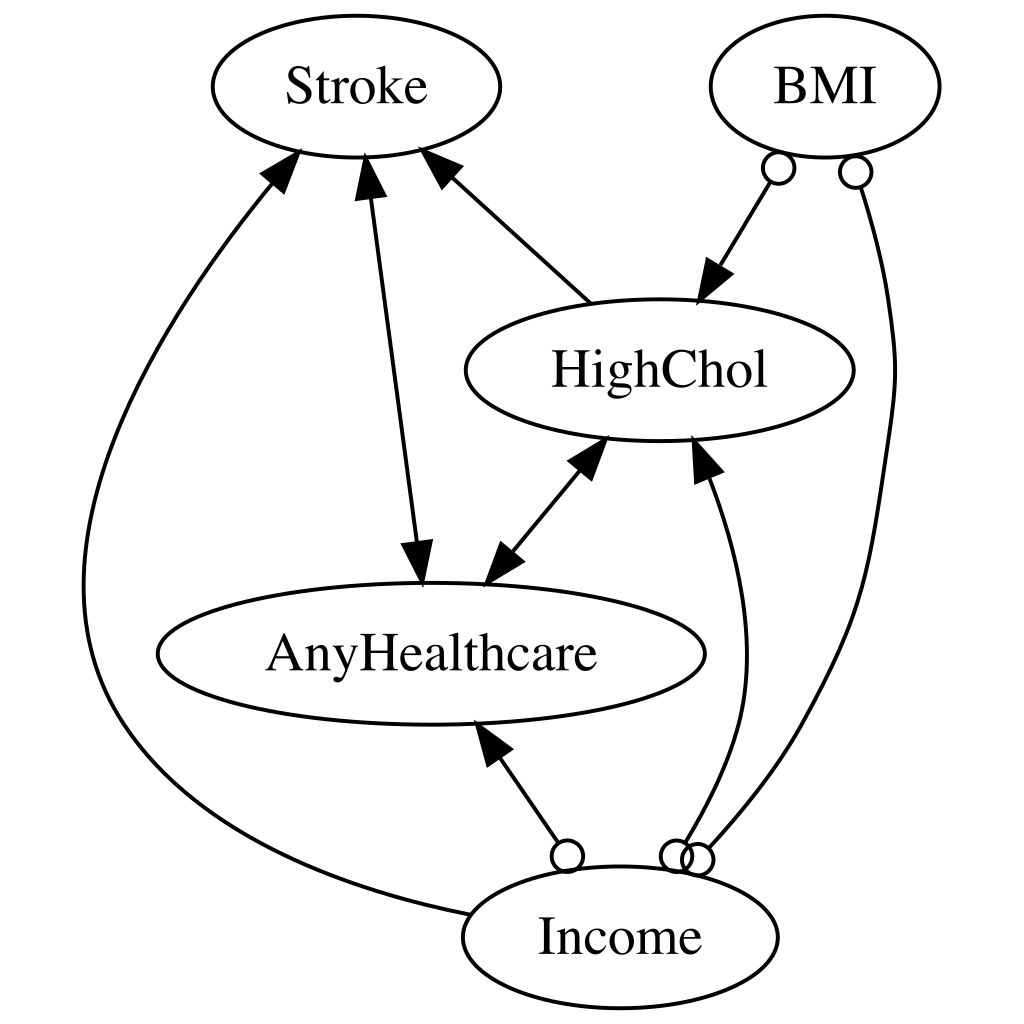}
    }
    \end{minipage}
\end{minipage}

\begin{minipage}{\linewidth}
    \begin{minipage}{0.32\linewidth}
    \subfloat[]{
        \hspace{-1.7em}
        \includegraphics[width=\linewidth, left]{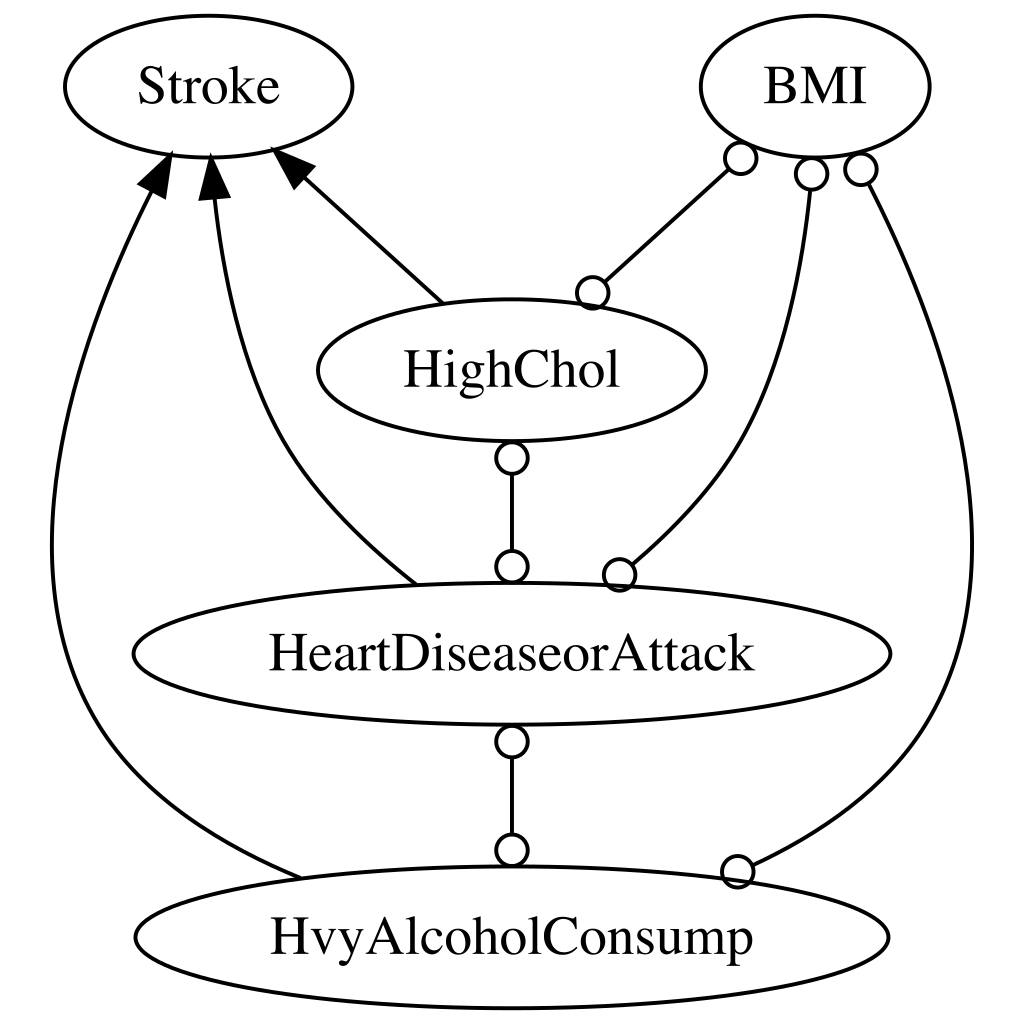}
    }
    \end{minipage}
   \begin{minipage}{0.32\linewidth}
    \subfloat[]{
        \hspace{-1.7em}
        \includegraphics[width=\linewidth, left]{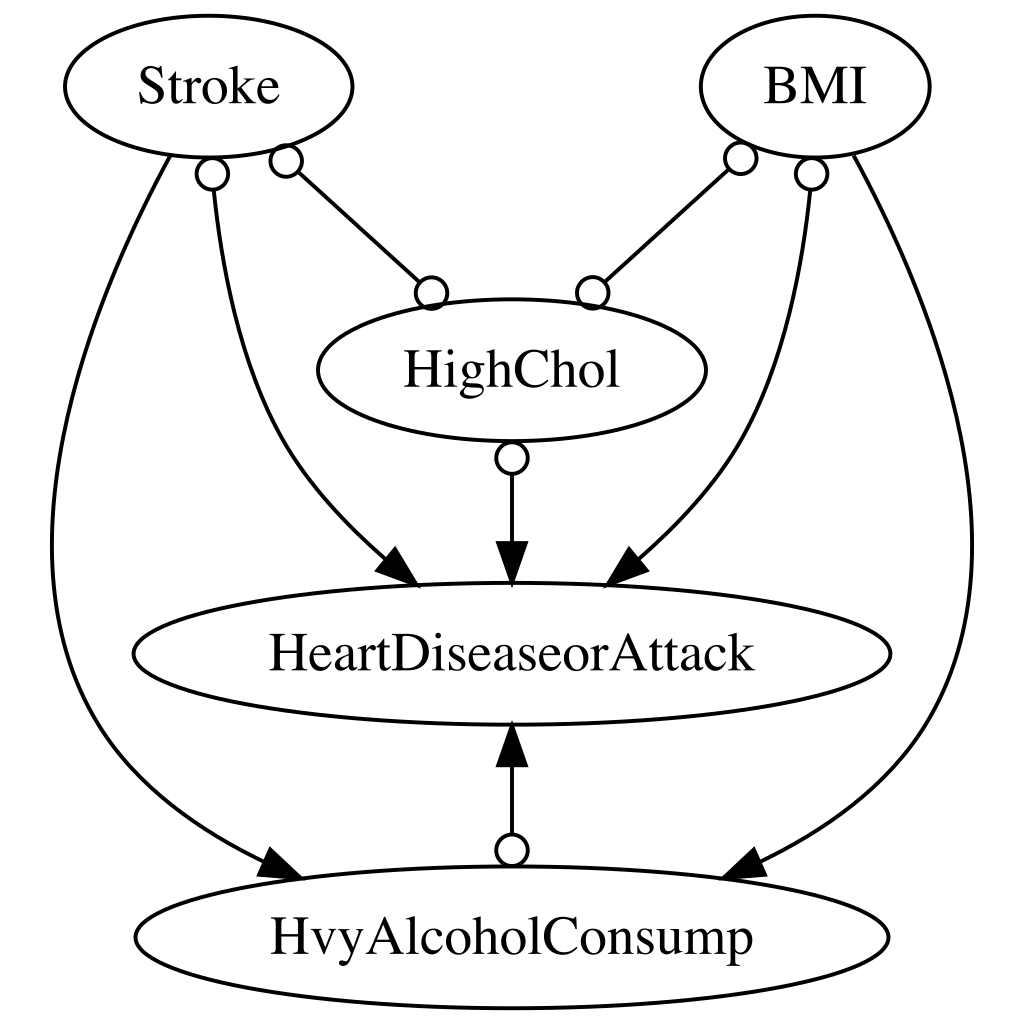}
    }
    \end{minipage}
 \begin{minipage}{0.32\linewidth}
    \subfloat[]{
        \hspace{-1.7em}
        \includegraphics[width=\linewidth, left]
       {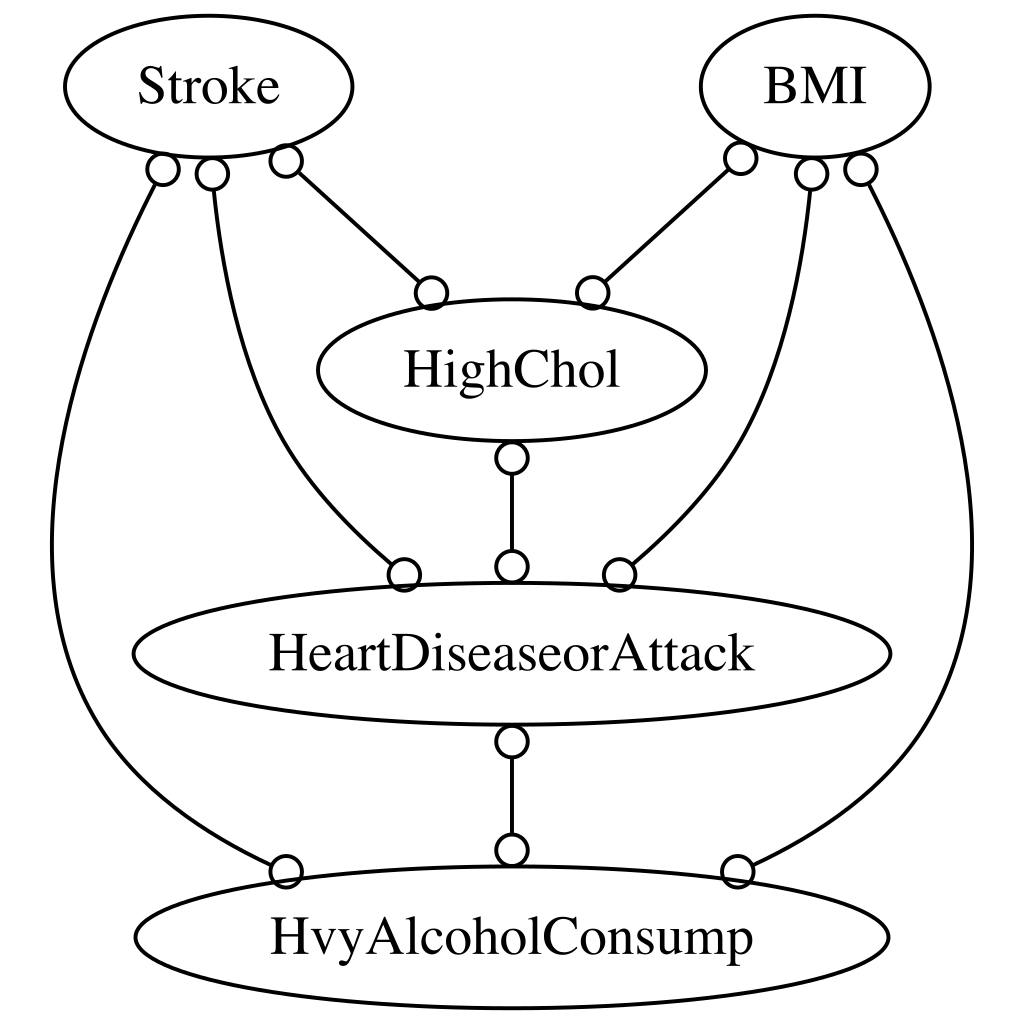}
    }
    \end{minipage}
\end{minipage}
\caption{Comparison of PAGs learned by dcFCI (left), FCI (center), and cFCI (right)
for five variables:
\textit{Stroke}, \textit{BMI}, \textit{HighChol}, \textit{HeartDiseaseorAttack}, and
\textit{HvyAlcoholComsump}.
dcFCI inferred PAGs with higher compatibility scores
(bounds: (a) [0, 0.0302], (d) [0, 0.00976], (g) [0, 0.00795])
compared to FCI ((b) [0, 0.0099], (e) [0, 0.00955], (h) [0, 0.00672]).
cFCI yields invalid PAGs in two cases ((c) and (i)), with one valid result (f) [0, 0.00955].}
\label{fig:stroke_extra}
\end{figure}


\begin{figure}[H]
    \centering
    \begin{minipage}{0.32\linewidth}
    \subfloat[]{
    \centering
        \includegraphics[width=0.85\linewidth]
        {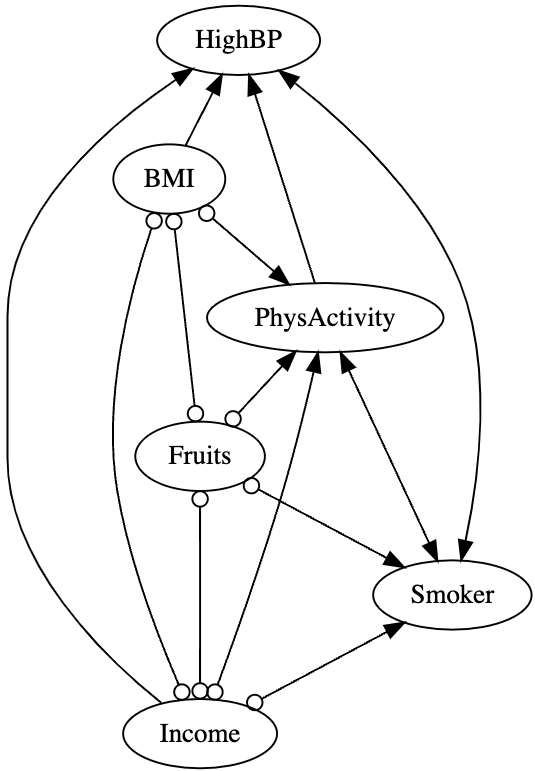}
    }
    \end{minipage}
   \begin{minipage}{0.3\linewidth}
    \subfloat[]{
        \hspace{-1.7em}
        \includegraphics[width=1.2\linewidth]{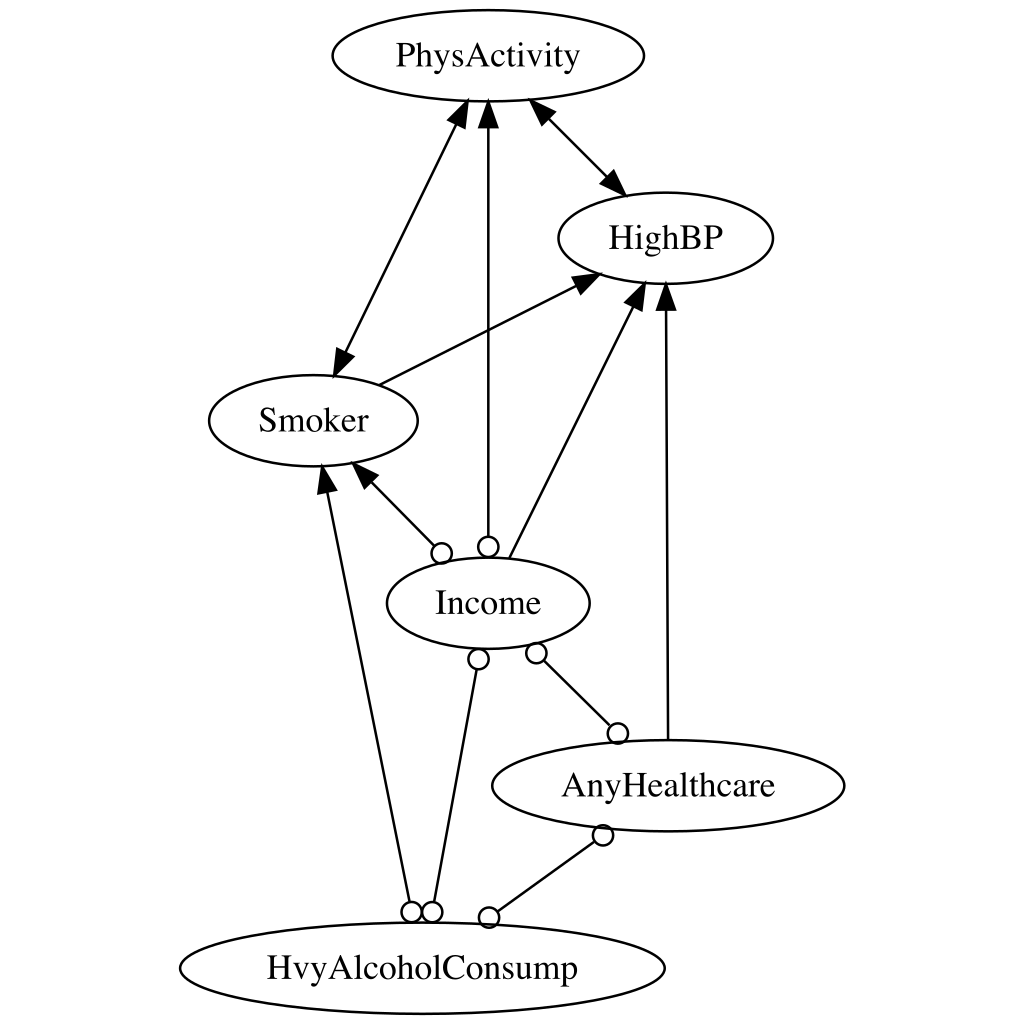}
    }
    \end{minipage}
 \begin{minipage}{0.32\linewidth}
    \subfloat[]{
        \hspace{-1.7em}
        \includegraphics[width=1.2\linewidth]{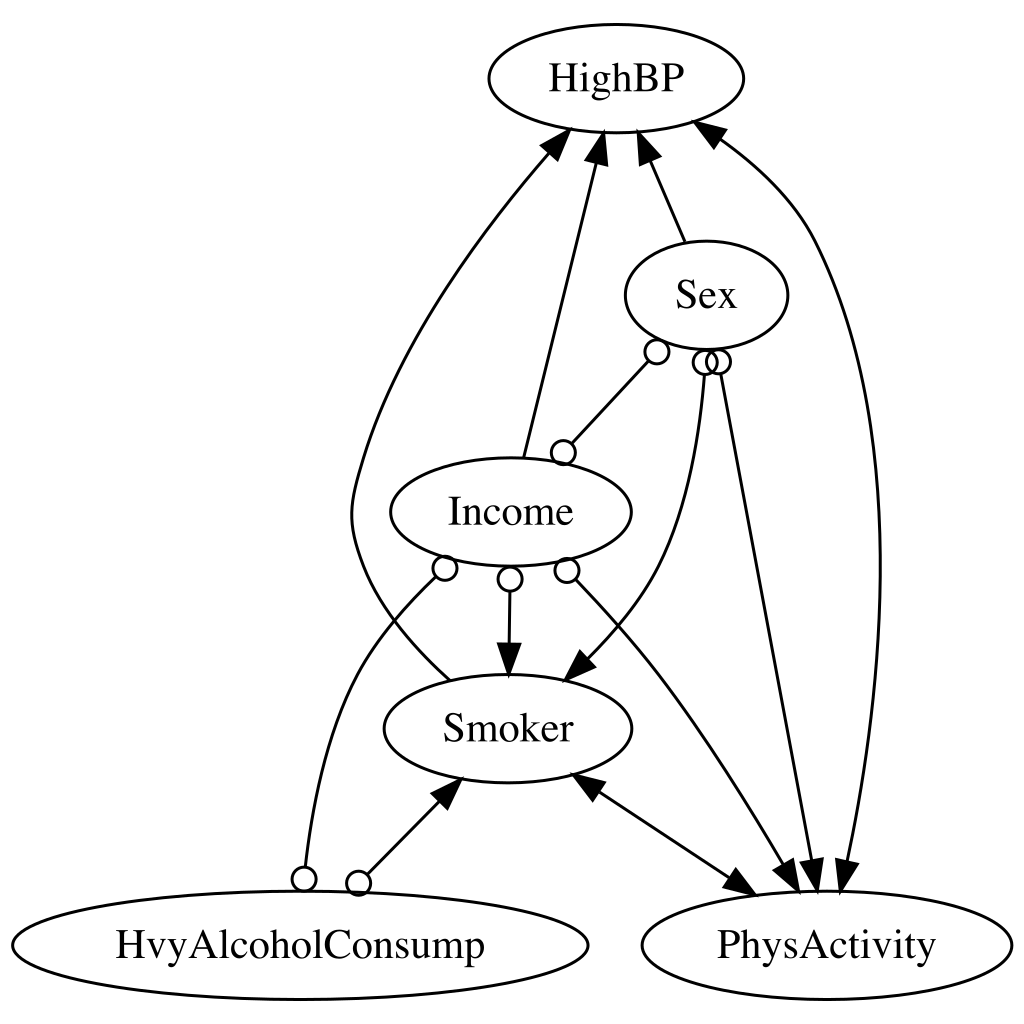}
    }
    \end{minipage}
    \caption{6-variable PAGs inferred by dcFCI, with score bounds of
    (a) [0, 0.0109], (b) [0, 0.0134], and (c) [0, 0.0134].
    The identification of Smoker in (a) and PhysActivity in (b,c) as
    non-causes of HighBP contradicts \Cref{fig:dhid_agreed_PAGs}-a.
 }
 \label{fig:dcfci_inconsistencies}
\end{figure}

\clearpage
\begin{table*}[ht]
\centering
\setlength{\tabcolsep}{1pt}
\caption{Bold rows highlight falsely identified independencies, where p-values
for the conditional independence tests exceed the significance
threshold of 0.01.}
\begin{tabular}{c|c|c|c|c|c|c|c|c|c|c|c}
  \hline
\multirow{2}{*}{$r$} & \multirow{2}{*}{$X$} & \multirow{2}{*}{$Y$} & \multirow{2}{*}{$\*Z$} &
\multirow{2}{*}{pvalue} & \multirow{2}{*}{$P(H_{X \indep Y | \*Z}|\mathcal{D})$} &
\multirow{2}{*}{$P(H_{X \nindep Y | \*Z}|\mathcal{D})$} & \multirow{2}{*}{Truth} &
\multirow{2}{*}{FCI} & \multirow{1}{*}{cFCI \&} & \multirow{2}{*}{BCCD} & \multirow{2}{*}{DCD} \\
& & & & & & & & & GPS & &  \\
  \hline \hline
   0 & A & B &  & $2.64 \times 10^{-14}$ & $1.78 \times 10^{-12}$ &  1 & dep & \cmark & \cmark & \cmark & \cmark\\
     0 & A & X &  & $4.13 \times 10^{-311}$ &    0 &  1 & dep & \cmark & \cmark & \cmark & \cmark\\
     0 & A & Y &  & 0.00276 & 0.0366 & 0.963 & dep & \xmark & \xmark & \cmark & \xmark\\
     0 & B & X &  & 0.496 & 0.672 & 0.328 & indep & \cmark & \xmark & \xmark & \xmark \\
     0 & B & Y &  &    0 &    0 &    1 & dep & \cmark & \cmark  & \cmark & \cmark \\
     0 & X & Y &  & 0.00937 & 0.105 & 0.895 & dep  & \xmark & \xmark & \cmark & \xmark \\
\hline
\end{tabular}
\label{tab:ex1_citests}
\end{table*}

\begin{table*}[h!]
\centering
\setlength{\tabcolsep}{2pt}
\caption{Number of perfectly recovered PAG across 300 simulations per sample size,
using Gaussian datasets under varying faithfulness degrees.
``Total'' indicated the number of datasets generated with full faithfulness
(defined by all and only dependencies with a p-value $\leq$ 0.05 or $P(H_1|\mathcal{D}) \geq 0.5$)
or r-MEC faithfulness.}
\begin{tabular}{l|c|c|c|c|c|c|c|c|c}
  \hline
\multirow{3}{*}{Algorithm} & \multirow{3}{*}{N} & \multicolumn{2}{x{2cm}|}{Full Faithful (p-value)} & \multicolumn{2}{x{2cm}|}{Full Faithful (Posterior Prob.)} &
\multicolumn{2}{c|}{\multirow{2}{*}{$r$-MEC Faithfull}} & \multicolumn{2}{c}{\multirow{2}{*}{Unfaithful}} \\
& & Total & Recoved & Total & Recoved & Total & Recoved & Total & Recoved \\
  \hline
 \hline
FCI & \multirow{8}{*}{1,000} & \multirow{8}{*}{0} &  0 &  \multirow{8}{*}{0} &  0 &  \multirow{8}{*}{0} &    0 &  \multirow{8}{*}{300} &   10 \\
  cFCI          &        &      &    0 &     &    0 &     &    0 &    &    1 \\
  BCCD          &        &      &    0 &     &    0 &     &    0 &    &    0 \\
  DCD           &        &      &    0 &     &    0 &     &    0 &    &    6 \\
  MAGSL         &        &      &    0 &     &    0 &     &    0 &    &   31 \\
  GPS           &        &      &    0 &     &    0 &     &    0 &    &   16 \\
\hline
\end{tabular}
\label{tab:gaussian_faith_pag_rec_rate}
\end{table*}

\begin{table*}[h!]
\centering
\setlength{\tabcolsep}{3pt}
\caption{Comparison in terms of False Discovery Rate (FDR) in simulations with Gaussian data.
For each sample size, all 300 simulations are considered.
Lower FDR indicates more accurate oriented edges.
Green downward arrows indicate significant improvement of dcFCI over SOTA algorithms.}
\begin{tabular}{l|c|c|c|c|c|c|c|c|c}
  \hline
  \multirow{2}{*}{Algorithm} & \multirow{2}{*}{N} & \multicolumn{6}{c|}{FDR}  & \multicolumn{2}{c}{Comparison}  \\
 & & Min & Q1 & Med & Mean & Q3 & Max & p-value & Diff\\
  \hline
 \hline
FCI & \multirow{7}{*}{$1,000$}  &   0 & 0.12 & 0.2 & 0.2 & 0.27 & 0.6 & 0.81 & -- \\
  cFCI &  &   0 & 0.091 & 0.14 & 0.17 & 0.25 & 0.6 & 0.0048 & $\uparrow$ \\
  BCCD &  &   0 & 0.091 & 0.14 & 0.17 & 0.25 & 0.62 & 0.078 & -- \\
  DCD &  &   0 & 0.18 & 0.29 & 0.29 & 0.4 & 0.75 & 1.6e-16 & $\color{PineGreen}{\boldsymbol\downarrow}$ \\
  MAGSL &  &   0 & 0.083 & 0.14 & 0.18 & 0.25 & 0.67 & 0.019 & $\uparrow$ \\
  GPS &  &   0 & 0.12 & 0.25 & 0.26 & 0.36 & 0.69 & 6.7e-06 & $\color{PineGreen}{\boldsymbol\downarrow}$ \\
\hline
\end{tabular}
\label{tab:gaussian_all_sims_fdr}
\end{table*}

\begin{table*}[h]
\centering
\setlength{\tabcolsep}{3pt}
\caption{Comparison in terms of False Omission Rate (FOR) in simulations with Gaussian data.
For each sample size, all 300 simulations are considered.
Lower FOR indicates more accurate non-oriented edges.
Green downward arrows indicate significant improvement of dcFCI over SOTA algorithms.}
\begin{tabular}{l|c|c|c|c|c|c|c|c|c}
  \hline
  \multirow{2}{*}{Algorithm} & \multirow{2}{*}{N} & \multicolumn{6}{c|}{FOR}  & \multicolumn{2}{c}{Comparison}  \\
 & & Min & Q1 & Med & Mean & Q3 & Max & p-value & Diff\\
 \hline
  \hline
FCI & \multirow{7}{*}{$1,000$} &   0 &   0 & 0.5 & 0.45 & 0.75 &   1 & 0.4 & -- \\
  cFCI &  &   0 & 0.57 & 0.75 & 0.68 & 0.88 &   1 & 7.9e-23 & $\color{PineGreen}{\boldsymbol\downarrow}$ \\
  BCCD &  &   0 & 0.5 & 0.67 & 0.62 & 0.83 &   1 & 3.5e-14 & $\color{PineGreen}{\boldsymbol\downarrow}$ \\
  DCD &  &   0 & 0.5 & 0.67 & 0.63 & 0.88 &   1 & 2.2e-13 & $\color{PineGreen}{\boldsymbol\downarrow}$ \\
  MAGSL &  &   0 &   0 & 0.5 & 0.45 & 0.75 &   1 & 0.32 & -- \\
  GPS &  &   0 & 0.24 & 0.5 & 0.51 & 0.88 &   1 & 0.015 & $\color{PineGreen}{\boldsymbol\downarrow}$ \\
\hline
\end{tabular}
\label{tab:gaussian_all_sims_for}
\end{table*}

\begin{table*}[h]
\centering
\setlength{\tabcolsep}{3pt}
\caption{Time comparison in simulations with Gaussian data.
Green downward arrows indicate significant improvement of dcFCI over SOTA algorithms.}
\begin{tabular}{l|c|c|c|c|c|c|c|c|c}
  \hline
  \multirow{3}{*}{Algorithm} & \multicolumn{6}{c|}{Time Taken (sec)}  & \multicolumn{3}{c}{Comparison}  \\
 & Min & Q1 & Med & Mean & Q3 & Max & p-value & p-value & Diff\\
 & &  &  &  &  &  & $k=1$ & $k=2$ & \\
 \hline
  \hline
  FCI  & 0.0476 & 0.0562 & 0.0587 & 0.0602 & 0.0605 & 0.2988 & 0 & 0 & $\uparrow$ \\
  cFCI  & 0.0518 & 0.0646 & 0.0681 & 0.0711 & 0.0729 & 0.4688 & 0 & 0 & $\uparrow$ \\
  BCCD  & 0.1453 & 0.1542 & 0.1610 & 0.1657 & 0.1717 & 0.4265 & 0 & 0 & $\uparrow$ \\
  DCD  & 48.75 & 320.92 & 779.43 & 716.38 & 985.49 & 5322.20 & 0 & 0 & $\color{PineGreen}{\boldsymbol\downarrow}$ \\
  MAGSL  & 3.3531 & 4.4828 & 4.9220 & 5.2544 & 5.4161 & 37.5230 & 1.3e-62 & 6.4e-257 & $\color{PineGreen}{\boldsymbol\downarrow}$ \\
  GPS &   0.0481 & 0.1537 & 0.1889 & 0.1973 & 0.2309 & 0.5609 & 0 & 0 & $\uparrow$ \\
  dcFCI ($k=1$) & 1.6590 & 2.3843 & 2.5932 & 2.9106 & 2.9275 & 25.7740 &  & &  \\
  dcFCI ($k=2$) & 1.2230 & 2.5976 & 3.3894 & 4.6335 & 5.0313 & 47.7950 &  & &   \\
   \hline
\end{tabular}
\label{tab:gaussian_all_sims_time_taken}
\end{table*}

\begin{table*}[h!]
\centering
\setlength{\tabcolsep}{3pt}
\caption{Comparison in terms of Structural Hamming Distance (SHD) in simulations with Gaussian data.
For each comparison, only simulations with valid inferred PAGs are considered.
Lower SHD indicates more accurate edges.
Green downward arrows indicate significant improvement of dcFCI over SOTA algorithms.}
\begin{tabular}{l|c|c|c|c|c|c|c|c|c|c}
  \hline
  \multirow{2}{*}{Algorithm} & \multirow{2}{*}{N} & \multirow{2}{*}{\# Sims} & \multicolumn{6}{c|}{SHD}  & \multicolumn{2}{c}{Comparison}  \\
 & & & Min & Q1 & Med & Mean & Q3 & Max & p-value & Diff\\
  \hline
 \hline
FCI & \multirow{7}{*}{$1,000$} & 293 &   0 &   4 &   5 & 5.3 &   7 &  12 & 0.15 & -- \\
  cFCI &  & 140 &   0 &   6 &   7 & 7.4 &  10 &  12 & 0.0046 & $\color{PineGreen}{\boldsymbol\downarrow}$ \\
  BCCD &  & 206 &   2 &   5 & 6.5 & 6.8 &   9 &  13 & 0.00037 & $\color{PineGreen}{\boldsymbol\downarrow}$ \\
  DCD &  & 300 &   0 &   6 &   8 &   8 &  10 &  16 & 4.9e-19 & $\color{PineGreen}{\boldsymbol\downarrow}$ \\
  MAGSL &  & 300 &   0 &   3 &   5 &   5 &   7 &  14 & 0.125 & -- \\
  GPS &  & 290 &   0 &   5 &   7 & 6.5 &   9 &  15 & 3.4e-05 & $\color{PineGreen}{\boldsymbol\downarrow}$ \\
\hline
\end{tabular}
\label{tab:gaussian_valid_sims_shd}
\end{table*}

\begin{table*}[h]
\centering
\setlength{\tabcolsep}{3pt}
\caption{Comparison in terms of False Discovery Rate (FOR) in simulations with Gaussian data.
For each comparison, only simulations with valid inferred PAGs are considered.
Lower FDR indicates more accurate oriented edges.
Green downward arrows indicate significant improvement of dcFCI over SOTA algorithms.}
\begin{tabular}{l|c|c|c|c|c|c|c|c|c|c}
  \hline
  \multirow{2}{*}{Algorithm} & \multirow{2}{*}{N} & \multirow{2}{*}{\# Sims} & \multicolumn{6}{c|}{FDR}  & \multicolumn{2}{c}{Comparison}  \\
 & & & Min & Q1 & Med & Mean & Q3 & Max & p-value & Diff\\
 \hline
  \hline
FCI & \multirow{7}{*}{$1,000$} & 293 &   0 & 0.12 & 0.2 & 0.2 & 0.27 & 0.6 & 0.87 & -- \\
  cFCI &  & 140 &   0 & 0.12 & 0.2 & 0.21 & 0.3 & 0.6 & 0.85 & -- \\
  BCCD &  & 206 &   0 & 0.1 & 0.14 & 0.19 & 0.25 & 0.62 & 1 & -- \\
  DCD &  & 300 &   0 & 0.18 & 0.29 & 0.29 & 0.4 & 0.75 & 1.6e-16 & $\color{PineGreen}{\boldsymbol\downarrow}$ \\
  MAGSL &  & 300 &   0 & 0.083 & 0.14 & 0.18 & 0.25 & 0.67 & 0.019 & $\uparrow$ \\
  GPS &  & 290 &   0 & 0.12 & 0.25 & 0.25 & 0.36 & 0.69 & 4.4e-05 & $\color{PineGreen}{\boldsymbol\downarrow}$ \\
\hline
\end{tabular}
\label{tab:gaussian_valid_sims_fdr}
\end{table*}

\begin{table*}[h]
\centering
\setlength{\tabcolsep}{3pt}
\caption{Comparison in terms of False Omission Rate (FOR) in simulations with Gaussian data.
For each comparison, only simulations with valid inferred PAGs are considered.
Lower FDR indicates more accurate non-oriented edges.
Green downward arrows indicate significant improvement of dcFCI over SOTA algorithms.}
\begin{tabular}{l|c|c|c|c|c|c|c|c|c|c}
  \hline
  \multirow{2}{*}{Algorithm} & \multirow{2}{*}{N} & \multirow{2}{*}{\# Sims} & \multicolumn{6}{c|}{FOR}  & \multicolumn{2}{c}{Comparison}  \\
 & & & Min & Q1 & Med & Mean & Q3 & Max & p-value & Diff\\
 \hline
  \hline
FCI & \multirow{7}{*}{$1,000$} & 293 &   0 &   0 & 0.5 & 0.45 & 0.75 &   1 & 0.39 & -- \\
  cFCI &  & 140 &   0 & 0.5 & 0.69 & 0.67 & 0.9 &   1 & 0.00013 & $\color{PineGreen}{\boldsymbol\downarrow}$ \\
  BCCD &  & 206 &   0 & 0.5 & 0.59 & 0.61 & 0.83 &   1 & 7.5e-07 & $\color{PineGreen}{\boldsymbol\downarrow}$ \\
  DCD &  & 300 &   0 & 0.5 & 0.67 & 0.63 & 0.88 &   1 & 2.2e-13 & $\color{PineGreen}{\boldsymbol\downarrow}$ \\
  MAGSL &  & 300 &   0 &   0 & 0.5 & 0.45 & 0.75 &   1 & 0.32 & -- \\
  GPS &  & 290 &   0 & 0.25 & 0.5 & 0.53 & 0.88 &   1 & 0.005 & $\color{PineGreen}{\boldsymbol\downarrow}$ \\
\hline
   \end{tabular}
\label{tab:gaussian_valid_sims_for}
\end{table*}

\begin{table*}[h]
\centering
\setlength{\tabcolsep}{1pt}
\caption{Comparison in terms of (straighforward) PAG Scores in simulations with Gaussian data.
For each comparison, only simulations with valid inferred PAGs are considered.
Higher scores indicate stronger support from the data.
Green upward arrows indicate significant improvement of dcFCI over SOTA algorithms.}
\begin{tabular}{l|c|c|c|c|c|c|c|c|c|c}
  \hline
  \multirow{2}{*}{Algorithm} & \multirow{2}{*}{N} & \multirow{2}{*}{\# Sims} & \multicolumn{6}{c|}{PAG Score}  & \multicolumn{2}{c}{Comparison}  \\
 & & & Min & Q1 & Med & Mean & Q3 & Max & p-value & Diff\\
  \hline
 \hline
FCI & \multirow{7}{*}{$1,000$} & 293 &   0 & 0.018 & 0.021 & 0.027 & 0.023 & 0.71 & 5e-04 & $\color{PineGreen}{\boldsymbol\uparrow}$ \\
  cFCI &  & 140 &   0 & 0.021 & 0.021 & 0.035 & 0.026 & 0.71 & 0.22 & -- \\
  BCCD &  & 206 &   0 & 0.021 & 0.022 & 0.041 & 0.03 & 0.71 & 1.5e-07 & $\downarrow$ \\
  DCD &  & 300 & $1.3 \times 10^{-9}$ & 0.0005 & 0.013 & 0.017 & 0.021 & 0.39 & 2.2e-22 & $\color{PineGreen}{\boldsymbol\uparrow}$ \\
  MAGSL &  & 300 & 0.021 & 0.021 & 0.022 & 0.036 & 0.026 & 0.71 & 1.1e-15 & $\downarrow$ \\
  GPS &  & 290 &   0 & 0.021 & 0.021 & 0.03 & 0.024 & 0.71 & 0.21 & -- \\
\hline
\end{tabular}
\label{tab:gaussian_valid_sims_pag_score}
\end{table*}

\begin{table*}[h]
\centering
\setlength{\tabcolsep}{3pt}
\caption{Comparison in terms of BIC distance in simulations with Gaussian data.
For each comparison, only simulations with valid inferred PAGs are considered.
Lower distances indicate more accurate PAGs.
Green upward arrows indicate significant improvement of dcFCI over SOTA algorithms.}
\begin{tabular}{l|c|c|c|c|c|c|c|c|c|c}
  \hline
  \multirow{2}{*}{Algorithm} & \multirow{2}{*}{N} & \multirow{2}{*}{\# Sims} & \multicolumn{6}{c|}{BIC Distance}  & \multicolumn{2}{c}{Comparison}  \\
 & & & Min & Q1 & Med & Mean & Q3 & Max & p-value & Diff\\
 \hline
  \hline
  FCI & \multirow{7}{*}{$1,000$} & 293 & $9.1 \times 10^{-13}$ & 5.3 & $2.8 \times 10^{3}$ & $1.8 \times 10^{3}$ & $2.8 \times 10^{3}$ & $8.4 \times 10^{3}$ & 3.8e-07 & $\color{PineGreen}{\boldsymbol\downarrow}$ \\
  cFCI &  & 140 & $4.5 \times 10^{-12}$ & 3.4 & 7.7 & $1.3 \times 10^{3}$ & $2.8 \times 10^{3}$ & $8.4 \times 10^{3}$ & 0.57 & -- \\
  BCCD &  & 206 & 0.01 &   3 &  14 & $1.5 \times 10^{3}$ & $2.8 \times 10^{3}$ & $5.7 \times 10^{3}$ & 0.059 & -- \\
  DCD &  & 300 & $2.7 \times 10^{-12}$ &  10 & $2.8 \times 10^{3}$ & $2.3 \times 10^{3}$ & $2.8 \times 10^{3}$ & $5.7 \times 10^{3}$ & 0.00015 & $\color{PineGreen}{\boldsymbol\downarrow}$ \\
  MAGSL &  & 300 & $9.1 \times 10^{-13}$ & 2.6 & 7.1 & $1.2 \times 10^{3}$ & $2.8 \times 10^{3}$ & $5.7 \times 10^{3}$ & 0.016 & $\uparrow$ \\
  GPS &  & 290 &   0 & 2.6 & 7.7 & $1.3 \times 10^{3}$ & $2.8 \times 10^{3}$ & $5.7 \times 10^{3}$ & 0.21 & -- \\
\hline
\end{tabular}
\label{tab:gaussian_valid_sims_bic_distance}
\end{table*}

\begin{table*}[h]
\centering
\setlength{\tabcolsep}{3pt}
\caption{Comparison in terms of BIC values in simulations with Gaussian data.
For each comparison, only simulations with valid inferred PAGs are considered.
Lower scores indicate a better goodness-of-fit.}
\begin{tabular}{l|c|c|c|c|c|c|c|c|c|c}
  \hline
  \multirow{2}{*}{Algorithm} & \multirow{2}{*}{N} & \multirow{2}{*}{\# Sims} & \multicolumn{6}{c|}{BIC}  & \multicolumn{2}{c}{Comparison}  \\
 & & & Min & Q1 & Med & Mean & Q3 & Max & p-value & Diff\\
  \hline
  FCI & \multirow{7}{*}{$1,000$} & 293 & $2.2 \times 10^{3}$ & $7.5 \times 10^{3}$ & $7.9 \times 10^{3}$ & $8.6 \times 10^{3}$ & $1 \times 10^{4}$ & $1.4 \times 10^{4}$ & 0.00094 & $\uparrow$ \\
  cFCI &  & 140 & $2.2 \times 10^{3}$ & $7.5 \times 10^{3}$ & $7.8 \times 10^{3}$ & $8.5 \times 10^{3}$ & $1 \times 10^{4}$ & $1.3 \times 10^{4}$ & 0.057 & -- \\
  BCCD &  & 206 & $2.4 \times 10^{3}$ & $7.6 \times 10^{3}$ & $7.8 \times 10^{3}$ & $8.3 \times 10^{3}$ & $8.3 \times 10^{3}$ & $1.1 \times 10^{4}$ & 1.9e-13 & $\uparrow$ \\
  DCD &  & 300 & $3.7 \times 10^{3}$ & $7.3 \times 10^{3}$ & $7.8 \times 10^{3}$ & $7.9 \times 10^{3}$ & $8.1 \times 10^{3}$ & $1.4 \times 10^{4}$ & 0.025 & $\uparrow$ \\
  MAGSL &  & 300 & $4.8 \times 10^{3}$ & $7.7 \times 10^{3}$ & $8 \times 10^{3}$ & $8.8 \times 10^{3}$ & $1 \times 10^{4}$ & $1.1 \times 10^{4}$ & 1.5e-24 & $\uparrow$ \\
  GPS &  & 290 & $4.7 \times 10^{3}$ & $7.6 \times 10^{3}$ & $8.1 \times 10^{3}$ & $8.9 \times 10^{3}$ & $1.1 \times 10^{4}$ & $1.4 \times 10^{4}$ & 0.064 & -- \\
\hline
\end{tabular}
\label{tab:gaussian_valid_sims_bic}
\end{table*}

\begin{table*}[h]
\centering
\setlength{\tabcolsep}{3pt}
\caption{Comparison in terms of False Discovery Rate (FDR) in simulations with mixed data types.
For each sample size, all 300 simulations are considered.
Lower FDR indicates more accurate oriented edges.
Green downward arrows indicate significant improvement of dcFCI over SOTA algorithms.}
\begin{tabular}{l|c|c|c|c|c|c|c|c|c}
\hline
\multirow{2}{*}{Algorithm} & \multirow{2}{*}{N} & \multicolumn{6}{c|}{FDR}  & \multicolumn{2}{c}{Comparison}  \\
& & Min & Q1 & Med & Mean & Q3 & Max & p-value & Diff\\
\hline
\hline
FCI & \multirow{3}{*}{$1,000$} &   0 & 0.12 & 0.21 & 0.22 & 0.3 & 0.67 & 1.1e-09 & $\color{PineGreen}{\boldsymbol\downarrow}$ \\
  cFCI &  &   0 &   0 & 0.2 & 0.18 & 0.29 & 0.67 & 0.61 & -- \\
  dcFCI &  &   0 & 0.083 & 0.14 & 0.17 & 0.23 & 0.67 &   &  \\
\hline
  FCI & \multirow{3}{*}{$5,000$} &   0 &   0 & 0.091 & 0.11 & 0.19 & 0.57 & 7.3e-05 & $\color{PineGreen}{\boldsymbol\downarrow}$ \\
  cFCI &  &   0 &   0 &   0 & 0.096 & 0.19 & 0.62 & 0.67 & -- \\
  dcFCI &  &   0 &   0 &   0 & 0.084 & 0.12 & 0.59 &   &  \\
\hline
  FCI & \multirow{3}{*}{$10,000$} &   0 &   0 &   0 & 0.073 & 0.12 & 0.58 & 8.5e-06 & $\color{PineGreen}{\boldsymbol\downarrow}$ \\
  cFCI &  &   0 &   0 &   0 & 0.065 & 0.12 & 0.58 & 0.084 & -- \\
  dcFCI &  &   0 &   0 &   0 & 0.045 & 0.073 & 0.54 &   &  \\
\hline
  FCI & \multirow{3}{*}{$50,000$} &   0 &   0 &   0 & 0.018 &   0 & 0.5 & 0.0052 & $\color{PineGreen}{\boldsymbol\downarrow}$ \\
  cFCI &  &   0 &   0 &   0 & 0.021 &   0 & 0.5 & 0.00082 & $\color{PineGreen}{\boldsymbol\downarrow}$ \\
  dcFCI &  &   0 &   0 &   0 & 0.01 &   0 & 0.53 &   &  \\
   \hline
\end{tabular}
\label{tab:mixed_all_sims_fdr}
\end{table*}

\begin{table*}[h]
\centering
\setlength{\tabcolsep}{3pt}
\caption{Comparison in terms of False Omission Rate (FOR) in simulations with mixed data types.
For each sample size, all 300 simulations are considered.
Lower FOR indicates more accurate non-oriented edges.
Green downward arrows indicate significant improvement of dcFCI over SOTA algorithms.}
\begin{tabular}{l|c|c|c|c|c|c|c|c|c}
\hline
\multirow{2}{*}{Algorithm} & \multirow{2}{*}{N} & \multicolumn{6}{c|}{FOR}  & \multicolumn{2}{c}{Comparison}  \\
& & Min & Q1 & Med & Mean & Q3 & Max & p-value & Diff\\
\hline
\hline
FCI & \multirow{3}{*}{$1,000$} &   0 &   0 & 0.45 & 0.42 & 0.75 &   1 & 0.22 & -- \\
  cFCI &  &   0 & 0.5 & 0.71 & 0.65 & 0.84 &   1 & 1.7e-24 & $\color{PineGreen}{\boldsymbol\downarrow}$ \\
  dcFCI &  &   0 &   0 & 0.25 & 0.34 & 0.67 &   1 &   &  \\
\hline
  FCI & \multirow{3}{*}{$5,000$} &   0 &   0 &   0 & 0.21 & 0.4 &   1 & 0.069 & -- \\
  cFCI &  &   0 &   0 & 0.5 & 0.43 & 0.75 &   1 & 1.7e-23 & $\color{PineGreen}{\boldsymbol\downarrow}$ \\
  dcFCI &  &   0 &   0 &   0 & 0.16 & 0.14 &   1 &   &  \\
\hline
  FCI & \multirow{3}{*}{$10,000$} &   0 &   0 &   0 & 0.13 &   0 &   1 & 0.02 & $\color{PineGreen}{\boldsymbol\downarrow}$ \\
  cFCI &  &   0 &   0 &   0 & 0.27 & 0.57 &   1 & 3.6e-21 & $\color{PineGreen}{\boldsymbol\downarrow}$ \\
  dcFCI &  &   0 &   0 &   0 & 0.09 &   0 &   1 &   &  \\
\hline
  FCI & \multirow{3}{*}{$50,000$} &   0 &   0 &   0 & 0.048 &   0 &   1 & 3.6e-05 & $\color{PineGreen}{\boldsymbol\downarrow}$ \\
  cFCI &  &   0 &   0 &   0 & 0.081 &   0 &   1 & 2.6e-12 & $\color{PineGreen}{\boldsymbol\downarrow}$ \\
  dcFCI &  &   0 &   0 &   0 & 0.0094 &   0 &   1 &   &  \\
   \hline
\end{tabular}
\label{tab:mixed_all_sims_for}
\end{table*}

\begin{table*}[h]
\centering
\setlength{\tabcolsep}{3pt}
\caption{Time comparison in simulations with mixed data types.
Green downward arrows indicate significant improvement of dcFCI over SOTA algorithms.}
\begin{tabular}{l|c|c|c|c|c|c|c|c|c}
\hline
\multirow{2}{*}{Algorithm} & \multirow{2}{*}{N} & \multicolumn{6}{c|}{Time Taken} (seconds) & \multicolumn{2}{c}{Comparison}  \\
& & Min & Q1 & Med & Mean & Q3 & Max & p-value & Diff\\
\hline
\hline
FCI & \multirow{3}{*}{$1,000$} & 0.016 & 0.027 & 0.035 & 0.035 & 0.041 & 0.069 & 9.8e-91 & $\uparrow$  \\
  cFCI &  & 0.017 & 0.04 & 0.053 & 0.055 & 0.066 & 0.12 & 9.8e-91 & $\uparrow$  \\
  dcFCI &  & 1.3 & 5.9 & 7.2 & 9.2 & 9.7 &  56 &   &  \\
\hline
  FCI & \multirow{3}{*}{$5,000$} & 0.017 & 0.036 & 0.042 & 0.044 & 0.048 & 0.35 & 9.8e-91 & $\uparrow$  \\
  cFCI &  & 0.028 & 0.054 & 0.065 & 0.067 & 0.081 & 0.13 & 9.8e-91 & $\uparrow$  \\
  dcFCI &  & 4.2 & 5.6 &   6 & 6.4 & 6.8 &  30 &   &  \\
\hline
  FCI & \multirow{3}{*}{$10,000$} & 0.021 & 0.038 & 0.043 & 0.05 & 0.053 & 0.41 & 9.8e-91 & $\uparrow$  \\
  cFCI &  & 0.034 & 0.056 & 0.069 & 0.074 & 0.084 & 0.42 & 9.8e-91 & $\uparrow$  \\
  dcFCI &  & 4.1 & 5.8 & 6.1 & 6.3 & 6.4 &  36 &   &  \\
\hline
  FCI & \multirow{3}{*}{$50,000$} & 0.025 & 0.041 & 0.047 & 0.054 & 0.06 & 0.55 & 9.8e-91 & $\uparrow$  \\
  cFCI &  & 0.046 & 0.06 & 0.074 & 0.082 & 0.096 & 0.2 & 9.8e-91 & $\uparrow$  \\
  dcFCI &  & 4.3 & 5.7 & 6.1 &   6 & 6.3 & 8.8 &   &  \\
   \hline
\end{tabular}
\end{table*}

\begin{table*}[h]
\centering
\setlength{\tabcolsep}{3pt}
\caption{Comparison in terms of (straightforward) PAG score in simulations with mixed data types.
For each comparison, only simulations with valid inferred PAGs are considered.
Higher scores indicate stronger support from the data.
Green upward arrows indicate significant improvement of dcFCI over SOTA algorithms.}
\begin{tabular}{l|c|c|c|c|c|c|c|c|c|c}
  \hline
  \multirow{2}{*}{Algorithm} & \multirow{2}{*}{N} & \multirow{2}{*}{\# Sims} & \multicolumn{6}{c|}{PAG Score}  & \multicolumn{2}{c}{Comparison}  \\
 & & & Min & Q1 & Med & Mean & Q3 & Max & p-value & Diff\\
  \hline
  \hline
FCI & \multirow{3}{*}{$1,000$} & 293 &   0 & 0.01 & 0.021 & 0.022 & 0.023 & 0.25 & 3.2e-51 & $\color{PineGreen}{\boldsymbol\uparrow}$  \\
  cFCI &  & 156 &   0 & 0.0067 & 0.021 & 0.025 & 0.024 & 0.25 & 1.5e-23 & $\color{PineGreen}{\boldsymbol\uparrow}$  \\
  dcFCI &  & 300 & 0.021 & 0.024 & 0.38 & 0.26 & 0.38 & 0.54 &   &  \\
\hline
  FCI & \multirow{3}{*}{$5,000$} & 297 &   0 & 0.0021 & 0.012 & 0.028 & 0.015 & 0.93 & 6.5e-66 & $\color{PineGreen}{\boldsymbol\uparrow}$  \\
  cFCI &  & 190 &   0 & 0.0012 & 0.012 & 0.038 & 0.016 & 0.93 & 3.9e-38 & $\color{PineGreen}{\boldsymbol\uparrow}$  \\
  dcFCI &  & 300 & 0.0021 & 0.16 & 0.17 & 0.22 & 0.25 & 0.88 &   &  \\
\hline
  FCI & \multirow{3}{*}{$10,000$} & 293 &   0 & 0.0044 & 0.0052 & 0.052 & 0.02 & 0.99 & 3.6e-60 & $\color{PineGreen}{\boldsymbol\uparrow}$  \\
  cFCI &  & 208 &   0 & 0.0044 & 0.0075 & 0.071 & 0.03 & 0.99 & 9.4e-39 & $\color{PineGreen}{\boldsymbol\uparrow}$  \\
  dcFCI &  & 300 & 0.0013 & 0.084 & 0.11 & 0.25 & 0.47 & 0.87 &   &  \\
\hline
  FCI & \multirow{3}{*}{$50,000$} & 298 &   0 & 0.0019 & 0.012 & 0.21 & 0.21 &   1 & 6.1e-42 & $\color{PineGreen}{\boldsymbol\uparrow}$  \\
  cFCI &  & 275 &   0 & 0.0019 & 0.015 & 0.22 & 0.24 &   1 & 2.8e-38 & $\color{PineGreen}{\boldsymbol\uparrow}$  \\
  dcFCI &  & 300 & 0.00055 & 0.051 & 0.29 & 0.44 & 0.86 & 0.99 &   &  \\
   \hline
\end{tabular}
\label{tab:mixed_valid_sims_pag_score}
\end{table*}

\begin{table*}[h]
\centering
\setlength{\tabcolsep}{3pt}
\caption{Comparison in terms of Structural Hamming Distance (SHD) in simulations with mixed data types.
For each comparison, only simulations with valid inferred PAGs are considered.
Lower SHD indicates more accurate edges.
Green downward arrows indicate significant improvement of dcFCI over SOTA algorithms.}
\begin{tabular}{l|c|c|c|c|c|c|c|c|c|c}
  \hline
  \multirow{2}{*}{Algorithm} & \multirow{2}{*}{N} & \multirow{2}{*}{\# Sims} & \multicolumn{6}{c|}{SHD}  & \multicolumn{2}{c}{Comparison}  \\
 & & & Min & Q1 & Med & Mean & Q3 & Max & p-value & Diff\\
  \hline
  \hline
FCI & \multirow{3}{*}{$1,000$} & 293 &   0 &   3 &   5 & 5.3 &   7 &  13 & 1.1e-05 & $\color{PineGreen}{\boldsymbol\downarrow}$ \\
  cFCI &  & 156 &   0 &   5 &   8 & 7.4 &  10 &  13 & 4.4e-07 & $\color{PineGreen}{\boldsymbol\downarrow}$ \\
  dcFCI &  & 300 &   0 &   2 &   4 & 4.4 &   6 &  15 &   &  \\
\hline
  FCI & \multirow{3}{*}{$5,000$} & 297 &   0 &   0 &   2 & 2.8 &   4 &  12 & 0.014 & $\color{PineGreen}{\boldsymbol\downarrow}$ \\
  cFCI &  & 190 &   0 &   0 &   2 & 3.3 &   5 &  13 & 0.00022 & $\color{PineGreen}{\boldsymbol\downarrow}$ \\
  dcFCI &  & 300 &   0 &   0 &   0 & 2.2 &   4 &  14 &   & \\
\hline
  FCI & \multirow{3}{*}{$10,000$} & 293 &   0 &   0 &   0 & 1.8 &   3 &  12 & 0.0026 & $\color{PineGreen}{\boldsymbol\downarrow}$ \\
  cFCI &  & 208 &   0 &   0 &   0 & 1.9 & 2.2 &  13 & 1.1e-05 & $\color{PineGreen}{\boldsymbol\downarrow}$ \\
  dcFCI &  & 300 &   0 &   0 &   0 & 1.2 &   2 &  11 &   &  \\
\hline
  FCI & \multirow{3}{*}{$50,000$} & 298 &   0 &   0 &   0 & 0.54 &   0 &  11 & 0.0046 & $\color{PineGreen}{\boldsymbol\downarrow}$ \\
  cFCI &  & 275 &   0 &   0 &   0 & 0.54 &   0 &  11 & 0.00088 & $\color{PineGreen}{\boldsymbol\downarrow}$ \\
  dcFCI &  & 300 &   0 &   0 &   0 & 0.23 &   0 &  11 &   &  \\
   \hline
\end{tabular}
\label{tab:mixed_valid_sims_shd}
\end{table*}

\begin{table*}[h]
\centering
\setlength{\tabcolsep}{3pt}
\caption{Comparison in terms of False Discovery Rate (FDR) in simulations with mixed data types.
For each comparison, only simulations with valid inferred PAGs are considered.
Lower FDR indicates more accurate oriented edges.
Green downward arrows indicate significant improvement of dcFCI over SOTA algorithms.}
\begin{tabular}{l|c|c|c|c|c|c|c|c|c|c}
  \hline
  \multirow{2}{*}{Algorithm} & \multirow{2}{*}{N} & \multirow{2}{*}{\# Sims} & \multicolumn{6}{c|}{FDR}  & \multicolumn{2}{c}{Comparison}  \\
 & & & Min & Q1 & Med & Mean & Q3 & Max & p-value & Diff\\
  \hline
  \hline
FCI & \multirow{3}{*}{$1,000$} & 293 &   0 & 0.12 & 0.21 & 0.22 & 0.3 & 0.67 & 5.5e-09 & $\color{PineGreen}{\boldsymbol\downarrow}$ \\
  cFCI &  & 156 &   0 & 0.14 & 0.2 & 0.23 & 0.33 & 0.67 & 0.14 & -- \\
  dcFCI &  & 300 &   0 & 0.083 & 0.14 & 0.17 & 0.23 & 0.67 &   &  \\
\hline
  FCI & \multirow{3}{*}{$5,000$} & 297 &   0 &   0 & 0.091 & 0.11 & 0.19 & 0.57 & 0.00019 & $\color{PineGreen}{\boldsymbol\downarrow}$ \\
  cFCI &  & 190 &   0 &   0 & 0.083 & 0.1 & 0.2 & 0.57 & 0.26 & -- \\
  dcFCI &  & 300 &   0 &   0 &   0 & 0.084 & 0.12 & 0.59 &   &  \\
\hline
  FCI & \multirow{3}{*}{$10,000$} & 293 &   0 &   0 &   0 & 0.064 & 0.12 & 0.58 & 0.00013 & $\color{PineGreen}{\boldsymbol\downarrow}$ \\
  cFCI &  & 208 &   0 &   0 &   0 & 0.059 & 0.11 & 0.58 & 0.0032 & $\color{PineGreen}{\boldsymbol\downarrow}$ \\
  dcFCI &  & 300 &   0 &   0 &   0 & 0.045 & 0.073 & 0.54 &   &  \\
\hline
  FCI & \multirow{3}{*}{$50,000$} & 298 &   0 &   0 &   0 & 0.015 &   0 & 0.38 & 0.013 & $\color{PineGreen}{\boldsymbol\downarrow}$ \\
  cFCI &  & 275 &   0 &   0 &   0 & 0.016 &   0 & 0.38 & 0.0066 & $\color{PineGreen}{\boldsymbol\downarrow}$ \\
  dcFCI &  & 300 &   0 &   0 &   0 & 0.01 &   0 & 0.53 &   &  \\
   \hline
\end{tabular}
\label{tab:mixed_valid_sims_fdr}
\end{table*}

\begin{table*}[h]
\centering
\caption{Comparison in terms of False Omission Rate (FOR) in simulations with mixed data types.
For each comparison, only simulations with valid inferred PAGs are considered.
Lower FOR indicates more accurate non-oriented edges.
Green downward arrows indicate significant improvement of dcFCI over SOTA algorithms.}
\setlength{\tabcolsep}{3pt}
\begin{tabular}{l|c|c|c|c|c|c|c|c|c|c}
  \hline
  \multirow{2}{*}{Algorithm} & \multirow{2}{*}{N} & \multirow{2}{*}{\# Sims} & \multicolumn{6}{c|}{FOR}  & \multicolumn{2}{c}{Comparison}  \\
 & & & Min & Q1 & Med & Mean & Q3 & Max & p-value & Diff\\
  \hline
  \hline
FCI & \multirow{3}{*}{$1,000$} & 293 &   0 &   0 & 0.5 & 0.43 & 0.75 &   1 & 0.16 & -- \\
  cFCI &  & 156 &   0 & 0.33 & 0.75 & 0.62 & 0.9 &   1 & 6.3e-05 & $\color{PineGreen}{\boldsymbol\downarrow}$ \\
  dcFCI &  & 300 &   0 &   0 & 0.25 & 0.34 & 0.67 &   1 &   &  \\
\hline
  FCI & \multirow{3}{*}{$5,000$} & 297 &   0 &   0 &   0 & 0.22 & 0.4 &   1 & 0.069 & -- \\
  cFCI &  & 190 &   0 &   0 &   0 & 0.31 & 0.67 &   1 & 8.6e-06 & $\color{PineGreen}{\boldsymbol\downarrow}$ \\
  dcFCI &  & 300 &   0 &   0 &   0 & 0.16 & 0.14 &   1 &   &  \\
\hline
  FCI & \multirow{3}{*}{$10,000$} & 293 &   0 &   0 &   0 & 0.14 &   0 &   1 & 0.02 & $\color{PineGreen}{\boldsymbol\downarrow}$ \\
  cFCI &  & 208 &   0 &   0 &   0 & 0.15 &   0 &   1 & 6.6e-05 & $\color{PineGreen}{\boldsymbol\downarrow}$ \\
  dcFCI &  & 300 &   0 &   0 &   0 & 0.09 &   0 &   1 &   &  \\
\hline
  FCI & \multirow{3}{*}{$50,000$} & 298 &   0 &   0 &   0 & 0.048 &   0 &   1 & 3.6e-05 & $\color{PineGreen}{\boldsymbol\downarrow}$ \\
  cFCI &  & 275 &   0 &   0 &   0 & 0.053 &   0 &   1 & 5.7e-06 & $\color{PineGreen}{\boldsymbol\downarrow}$ \\
  dcFCI &  & 300 &   0 &   0 &   0 & 0.0094 &   0 &   1 &   &  \\
   \hline
\end{tabular}
\label{tab:mixed_valid_sims_for}
\end{table*}
}